\documentclass[]{dukedissertation}
\usepackage[draft]{hyperref}

\usepackage{amsmath, amssymb, amsfonts, amsthm}
\usepackage{graphicx}
\usepackage{natbib}
\usepackage{color}
\usepackage{bm}
\usepackage{subfigure}
\usepackage{graphicx}
\usepackage[graphicx]{realboxes}
\usepackage{mathabx}
\usepackage{multirow,tabularx,adjustbox}
\usepackage{longtable}
\usepackage{rotating}
\usepackage{setspace}
\usepackage{pdflscape}
\usepackage{afterpage}
\usepackage{float}
\usepackage{ragged2e}
\usepackage{array, booktabs, ltablex, makecell, threeparttablex}
\usepackage{ragged2e}

\newcolumntype{R}{>{\RaggedRight\hspace{0pt}}X} 
\newcolumntype{L}{>{\raggedright\arraybackslash}X}
\newcolumntype{C}{>{\Centering\arraybackslash}X}
\usepackage[skip=1ex,labelfont=bf,font=small]{caption}
\usepackage{bbm}
\usepackage{upgreek}

\usepackage{siunitx}
\usepackage{changepage}

\graphicspath{{./Pictures/}}

\author{Sayan Mandal}
\title{Deep Learning to Predict Glaucoma Progression using Structural Changes in the Eye}
\supervisor{Felipe A. Medeiros}
\department{Electrical and Computer Engineering} 

\date{December, 08, 2023} 

\copyrighttext{ All rights reserved except the rights granted by the\\
   \href{http://creativecommons.org/licenses/by-nc/3.0/us/}
        {Creative Commons Attribution-Noncommercial Licence}
}

\member{Vahid Tarokh, Co-Supervisor}
\member{Leslie Collins}
\member{Ricardo Henao}
\member{Yiran Chen}

\begin{document}

\maketitle

\abstract

Glaucoma is a group of chronic eye diseases characterized by optic neuropathy, which causes irreversible vision loss. It is caused by progressive degeneration of the optic nerve, leading to gradual loss of the visual field from the periphery to the center, resulting in blindness if left untreated. Since the changes are gradual and the damage progresses generally slowly, glaucoma development is insidious and often diagnosed until it reaches an advanced stage. Early detection of glaucoma progression is necessary to monitor the atrophy and formulate treatment strategies to halt progressive functional vision impairments. The availability of data centric methods have made it possible for researchers to develop computer-aided algorithms for the clinical diagnosis of glaucoma and capture accurate disease characteristics. In this research, we use deep learning models, one such forefront, to identify complex disease characteristics and progression criteria, enabling the detection of subtle changes indicative of glaucoma progression. 

To this end, we investigate the structure-function relationship of glaucoma progression and explore the possibility of predicting functional impairment from structural eye deterioration. We also analyze various statistical and machine-learning methods that have aided previous attempts to estimate progression, including emerging deep-learning techniques that use structural features like optical coherence tomography (OCT) scans to predict glaucoma progression accurately. We show through our investigations that these methods are still prone to confounding risk factors, especially variability due to age, data imbalances, potential noisy labels, lack of gold standard criteria, etc. We developed novel semi-supervised time-series algorithms to overcome these multifaceted challenges using unique data-driven approaches: 

\textbf{Weakly-Supervised Time-Series Learning}: We develop a convolutional neural network-long short-term memory (CNN-LSTM) base model to encode the spatiotemporal features from the OCT scan sequence taken over a fixed follow-up. We model the rest of the deep learning architecture on the fact that original OCT sequences exhibit age-related progression, and reshuffling the sequence order, along with the knowledge of healthy eyes from a positive-unlabeled dataset, can establish robust pseudo-progression criteria for glaucoma. This circumvents the need for gold standard labels for disease progression. 

\textbf{Semi-supervised Time-Series Learning}: We extend the above notion to a labeled case where labels are obtained from Guided Progression Analysis (GPA), a well-known, stable, and accurate functional assessment for glaucoma progression, but might be prone to noisy labels due to nuances in data acquisition. We model the structural progression as a pseudo-identifier for functional glaucoma deficits. We use this knowledge in a contrastive learning scheme where the CNN-LSTM base architecture learns accurate spatiotemporal characteristics from potentially mislabeled data and improves predictions. 

Finally, we compare and show that these methods outperform conventional and state-of-the-art techniques.

\dedication{In the loving memory of my father, a source of strength and inspiration throughout my life.}

\tableofcontents 
\listoftables	
\listoffigures	
\abbreviations


\section*{Symbols}


\begin{symbollist}
    \item[$dB$] Denotes Decibels, A Unit of Visual Field Measure.
    \item[$\mu m$] Denotes Micro-Meter, A Unit of RNFL Thickness Measure.
    \item[$|\cdot|$] Denotes Absolute Value or Number of Items.
    \item[$\Vert \cdot \Vert_2$] Denotes Frobenius Norm of Matrix or Euclidean Norm of Vector.
    \item[$\mathbbm{1}$] Denotes Indicator Function.
    \item[$\Delta$] Denotes Difference between Values.
    \item[$e$] Residuals: Difference between Observed and Estimated Values.
    \item[$P$] Denotes Probability on Events or Distributions.
    \item[$\pi$] Denotes a Function for Random Number Generator.
    \item[$\alpha , \beta , \gamma$] Denotes Factor Constants used in Objective Functions.
    \item[$\delta, \epsilon$] Denotes Small Positive Quantities.
    \item[$T$] Denotes OCT Follow Up Time.
    \item[$x^{(t)}$] Denotes Input (Image) at Time t.
    \item[$X$] Denotes Input (Image) Sequence.
    \item[$y$] Denotes Outcome (Observed) Measure.
    \item[$\hat{y}$] Denotes Estimated (Predicted) Value.
    \item[$z^{(t)}$] Denotes Time Series Encoding at Time $t$.
    \item[$Z_n$] Denotes Spatiotemporal Encoding from CNN-LSTM Network.
    \item[$\tau$] Denotes No. of OCT Bscan Images (Inputs).
    \item[$\mathcal{T}$] Denotes Time at Endpoints for VF GPA.
    \item[$\kappa$] Denotes Cohen's Kappa, A Measure of Agreement Between Variables.
    \item[$r$] Denotes Correlation between Two Variables.
    \item[$H$] Denotes Entropy Function.
    \item[$L$] Denotes Non-Negative Real-Valued Loss Function.
    \item[$\mathcal{L}(\cdot)$] Denotes Likelihood Function.
    \item[$\mathbf{J}(\cdot)$] Denotes Objective Function.
    \item[$\Pi$] Denotes Randomizing Function with k Permutations.
    \item[$\mathcal{X}$] Denotes Input Space in Dataset.
    \item[$\mathcal{Y}$] Denotes Output Space in Dataset.
    \item[$\mathbb{R}^{n}$] Denotes Real-Valued n-Dimensional Space.
    \item[$H, W$] Denotes Image Height and Width.
    \item[$F$] Denotes 1D Feature Width.
    \item[$\mathcal{A}$] Denotes Augmentation Function.
    \item[$\theta$] Denotes CNN-LSTM Model Parameters.
    \item[$\mathbb{H}$] Denotes Hypothesis Space for DL Models.
    \item[$h$] Denotes DL Model Candidate from Hypothesis Space.
    \item[$\mathcal{H}$] Denotes DL Model Function.
    \item[$\phi, \psi$] Denotes Parameters for Projection Heads in the DL Model.
    \item[$\mathcal{D}, \mathcal{S}$] Denotes Dataset and Data Subset.
    \item[$R$] Denotes Risk Function.
    \item[$\mathbb{P}$] Denotes Theoretic Measure of Probability on Countable Sets.
    \item[$\uptau$] Denotes Temperature Parameter for Contrastive Learning.
\end{symbollist}


\section*{Abbreviations}

\begin{symbollist}
    \item[AA] African American
    \item[ACG] Angle-Closure Glaucoma
    \item[AGIS] Advanced Glaucoma Intervention Study
    \item[AI] Artificial Intelligence
    \item[AUC] Area Under [Receiver Operator Characteristic] Curve
    \item[BLUP] Best Linear Unbiased Predictors
    \item[BCE] Binary Cross Entropy
    \item[Bi-RM] Bidirectional Recurrent Model
    \item[CAM] Class Activation Maps
    \item[CAR] Conditional Auto-Regressive Models
    \item[CCE] Categorical Cross Entropy
    \item[CCT] Central Corneal Thickness
    \item[CDR] Cup-Disk Ratio
    \item[CIGTS] Collaborative Initial Glaucoma Treatment Study
    \item[CNN] Convolutional Neural Network
    \item[cpRNFL] Circumpapillary Retinal Nerve Fiber Layer
    \item[CPT] Current Procedural Terminology
    \item[CSLO] Confocal Scanning Laser Ophthalmoscopy
    \item[CTHMM] Continuous Time-Hidden Markov Model
    \item[DL] Deep Learning
    \item[DLS] Differential Light Sensitivity
    \item[DOR] Duke Ophthalmic Registry
    \item[DSF] Dynamic Structure-Function
    \item[EGPS] European Glaucoma Preventing Study
    \item[EHR] Electronic Health Records
    \item[EMGT] Early Manifest Glaucoma Trial
    \item[FC] Fully Connected
    \item[FBDS] Fuzzy Bayesian Detection Scheme
    \item[GAN] Generative Adversarial Network
    \item[GCC] Ganglion Cell Complex
    \item[GCIPL] Ganglion Cell-Inner Plexiform Layer
    \item[GDx-VCC] Glaucoma Detection with Variable Corneal Compensation
    \item[GEM] Gaussian Mixture Model Expectation Maximization
    \item[GLMM] Generalized Linear Mixed Models
    \item[GLT] Glaucoma Laser Trial
    \item[GMM] Gaussian Mixture Model
    \item[GON] Glaucomatous Optic Neuropathy
    \item[GPA] Guided Progression Analysis
    \item[GTN] Gated Transformer Network
    \item[HFA] Humphrey Field Analyzer
    \item[HRT] Heidelberg Retina Tomograph
    \item[ICD] International Classification of Diseases
    \item[ICA] Independent Component Analysis
    \item[IOP] Inter Ocular Pressure
    \item[IRB] Institutional Review Board
    \item[JLSM] Joint Longitudinal Survival Model
    \item[LMM] Linear Mixed Models
    \item[LSTM] Long Short-Term Memory
    \item[MAE] Mean Absolute Error
    \item[M2M] Machine to Machine
    \item[MD] Mean Deviation
    \item[MLE] Maximum Likelihood Estimation
    \item[ML] Machine Learning
    \item[MLP] Multi-Layer Perceptron
    \item[MCC] Matthew’s Correlation Coefficient
    \item[NLP] Natural Language Processing
    \item[NTG] Normal Tension Glaucoma
    \item[OAG] Open-Angle Glaucoma
    \item[OCT] Optical Coherence Tomography
    \item[OCTA] Optical Coherence Tomography Angiography
    \item[OHTS] Ocular Hypertension Treatment Study
    \item[OLS] Ordinary Least Squares
    \item[OLSLR] Ordinary Least Square Linear Regression
    \item[ONH] Optic Nerve Head
    \item[PACG] Primary Angle-Closure Glaucoma
    \item[PCA] Principal Component Analysis
    \item[PLR] Pointwise Linear Regression
    \item[PoP] Permutation of Points
    \item[PoPLR] Permutation of Pointwise Linear Regression
    \item[POAG] Primary Open-Angle Glaucoma
    \item[POD] Proper Orthogonal Decomposition
    \item[PSD] Pattern Standard Deviation
    \item[PU] Positive-Unlabeled Learning
    \item[RGC] Retinal Ganglion Cells
    \item[RNFL] Retinal Nerve Fiber Layer
    \item[RA] Rim Area
    \item[ROC] Receiver Operator Characteristics
    \item[SAP] Standard Automated Perimetry
    \item[SDOCT] Spectral-Domain Optical Coherence Tomography
    \item[SGD] Stochastic Gradient Descent
    \item[SITA] Swedish Interactive Threshold Algorithm
    \item[SLP] Scanning Laser Polarimetry
    \item[SSOCT] Swept-Source Optical Coherence Tomography
    \item[SVM] Support Vector Machine
    \item[TDOCT] Time-Domain Optical Coherence Tomography
    \item[TPA] Trend-Based Progression Analysis
    \item[TSNIT] Temporal, Superior, Nasal, Inferior and Temporal Regions
    \item[UKGTS] United Kingdom Glaucoma Treatment Study
    \item[VA] Visual Acuity
    \item[VAE] Variational Auto Encoder
    \item[VCA-MA] Variational Change Analysis with Markovian A-Priori
    \item[VF] Visual Field
    \item[VFI] Visual Field Index
    \item[VIM] Variational Bayesian Independent Component Mixture Model
    
\end{symbollist}


\acknowledgements

I would like to express my deepest gratitude to my advisor, Felipe Medeiros, whose invaluable feedback, consistent mentoring, and patience, even during my mistakes, have shaped this thesis. Your guidance has been instrumental, and I have learned so much from our discussions and your perspectives.

My mentor at the lab, Alessandro Jammal, has been a tremendous source of support, almost paralleling the guidance I received from my advisor. Your accessibility during various hours and your unwavering assistance were pivotal in my journey. I would also like to thank other lab members, Rizul Naithani, Davina Malek, Aasma Youssif, Vahid Ownagh, and Gary Gan, whose help, although varied, contributed significantly to my research.

To my defense committee members, your feedback during the preliminary exams enriched the final form of this work. Your expertise and insights were much appreciated. I would like to express my gratitude to my co-chair, Vahid Tarokh, for his invaluable support and guidance during my advisors transition.

I extend my heartfelt gratitude to the ECE department for their unwavering assistance throughout my academic journey. Their support in clarifying departmental intricacies, ensuring seamless scheduling, and handling administrative matters played a crucial role in allowing me to focus on my research. Their behind-the-scenes efforts are much appreciated.

To my friends at Duke University and Durham, Harsh Bandhey, Deniz Acil, Viswa Alaparthy, Tyler King, Lalit Yadav, Vidvat Ramachandran, Edward Hanson, Keerti Anand you have been my rock. The late-night hang-outs, study sessions, and moments of respite were possible because of you. Friends Chunge Wang, Francesco Luzi, Catalena Le, Arjun Sridhar, Vani Yadav and Bhavna Gopal deserve special mention for enduring my rants and being there for me. I'm also profoundly grateful to a special individual, Jayoshree Adhikari, who stood by me, providing mental and emotional strength. An exceptional thanks to Anurag Kashyap, who redefined the essence of true friendship for me.

My undergraduate and childhood friends, Sourav Agarwal, Utkarsh Tyagi, Sourav Kundu, Soumyadeep Dasgupta, Vinsea Singh, and Prateeksha Pamshetty, your fun conversations and cherished memories have been my solace, reminding me of simpler times and keeping me grounded.

The pillars of my life, my parents, have been my unwavering supporters even from 8,300 miles away. My late father, whose belief in me led me to pursue this PhD, supported me until his very last breath. My mother, with her boundless love and constant prayers, has been my anchor. She stood by me, talked to me whenever I needed, always providing words of encouragement. My sister, too, has been a beacon of strength, uplifting me during my lowest moments.

I would like to acknowledge the funding agency, Google Inc., whose support was integral to my research and PhD. I also recognize the National Institutes of Health/National Eye Institute grants EY029885 and EY031898 for their generous assistance.

Lastly, I know there are many who have crossed my path, provided support in various capacities, and have not been named here. Please know that I am genuinely grateful for every bit of support and kindness I've received. I am privileged to be surrounded by such incredible individuals, and I aspire to give back in any way I can for the support, no matter how small, that I've received.

%
%
%
\chapter{Introduction}

Vision is the ability of a person to see, interpret, and interact with the world. The eyes provide the sense of sight, capturing images from our surroundings. The brain interprets these images, allowing us to make sense of what we see. The sense of sight is one of the most vital sources of information we receive out of all five senses combined. Many of the movements and tasks we perform and our interactions with our environment rely heavily on vision. Thus, taking care of our vision is crucial, and protecting our sense of sight is fundamental to this care. 

As such, eye diseases may cause vision impairments, often leading to partial or complete loss of sight. When individuals lose the ability to see, basic activities that were once taken for granted, such as reading, driving, or recognizing loved ones, become arduous or even impossible and significantly reduce the quality of life. Among these eye diseases, glaucoma stands out as uniquely insidious. Unlike most other conditions, glaucoma develops slowly and without noticeable symptoms, leading to vision loss before even a person realizes the problem. Thus, detecting the progression of glaucoma early on its inception is necessary. This research focuses on investigating the current state of research in glaucoma progression detection and finding novel methods to detect glaucoma progression accurately and efficiently.

\section{Background}

\subsection{Brief History of Glaucoma}

The term glaucoma was first used in Hippocrates' Aphorisms (\cite{anderson2018glaucoma}), originating from the Greek words '$\Gamma\lambda\alpha\acute{\upsilon}\dot{V}\xi$ – $\Gamma\lambda\alpha\acute{\upsilon}\dot{V}\kappa o\varsigma$' (glaukos) and '$\Gamma\lambda\alpha\acute{\upsilon}\dot{V}\sigma\omega$' (glausso; \cite{tsatsos2007controversies}). 
Glaukos is a noun which is a non-specific term that means diseased bluish-green or light-gray hue on healthy irrides. Glausso, on the other hand, is a verb that means "to glow" or "to shine." Both words have slightly different meanings but point to the same connotation, "some sort of opacification or hardening of the cornea or lens resulting in apparent discoloration of the eye, most commonly occurring in the elderly" (\cite{fronimopoulos1991terms}). However in contemporary medicine the term glaucoma encompassed multiple eye diseases, including cataract (\cite{leffler2015glaucoma}), definitions of multiple eye conditions involving glaucoma such as hardness of the eyeball, increased eye pressure, dilated pupil and reduced vision were reported (\cite{leffler2015early}). It was not until the scientific advancements in ophthalmology during the 1700s and 1800s, that glaucoma and cataracts were recognized as distinct diseases with different pathological mechanisms.  William MacKenzie observed and described the role of interocular pressure (IOP) in glaucoma (\cite{mackenzie1855practical}), which was followed by an illustration of the first glaucomatous understanding of the optic disc by Jaeger (\cite{lazaridis2022deep}). Graefe spearheaded some of the most notable developments in the modern knowledge of glaucoma, such as the structural understanding (deterioration of optic nerve fibers), functional understanding (difference in central and peripheral vision within stages of glaucoma) and different types of glaucoma based on clinical signs of inflammation: acute, chronic or secondary (\cite{v1856ueber}). In the subsequent years, the development of modern scientific apparatus, assessment techniques, and advancements in medicine allowed researchers to define glaucoma more precisely, leading to the current understanding of its underlying pathophysiological mechanisms.

\subsection{Current Understanding}

Today, glaucoma is understood not just as a single disease but as a group of eye conditions with multiple etiologies leading to optic nerve damage. Elevated IOP is found to be a significant risk factor and not the whole cause as previously assumed (\cite{frankfort2013elevated}). It is found that different types of glaucoma have varying pathophysiologies and different treatments. Early detection, primarily through regular eye check-ups, is essential since vision loss due to glaucoma is reversible but can be slowed down with appropriate treatment. Advancements in treatment strategies and the advent of sophisticated diagnostic tools have enabled easy detection and monitoring of the disease's progression. The focus of treatment strategies has also expanded from eyedrops that decrease IOP to surgical interventions to reverse the damage to the optic nerve.

\section{A Clinical Overview of Glaucoma}
\label{sec:clinical}

Glaucoma is a type of chronic optic neuropathy caused by the pathological degeneration of the retinal ganglion cells (RGC) resulting in progressive visual function loss.  It is the leading cause of irreversible blindness worldwide, with an estimated 80 million affected by the disease and a projected 111.8 million people affected by 2040 (\cite{pascolini2012global, mariotti2012global}). 

\subsection{Pathophysiology}

Retina has photoreceptors, which are highly specialized cells that receive light and transmit visual information to the brain through RGC and their axons. The axons from the RGC converge to the optic disc in bundles, forming the retinal nerve fiber layer (RNFL). Glaucoma results from progressive RGC loss and axon degeneration, for which elevated IOP is the most important risk factor. This progressive cell loss gives rise to gradual visual field loss which generally starts from the periphery, but can advance to generalized visual field loss and blindness (\cite{hood2013glaucomatous}). Although the vision loss from glaucoma can be catastrophic, its generally chronic, slow, and discreet onset, with no early symptoms, makes it difficult to be perceived by the patient, thus making methods to detect glaucoma progression paramount, to provide treatment early on and prevent irreversible blindness.

\subsection{Types of Glaucoma}

\begin{figure}[!ht]
\centering
	\includegraphics[width=0.5\textwidth]{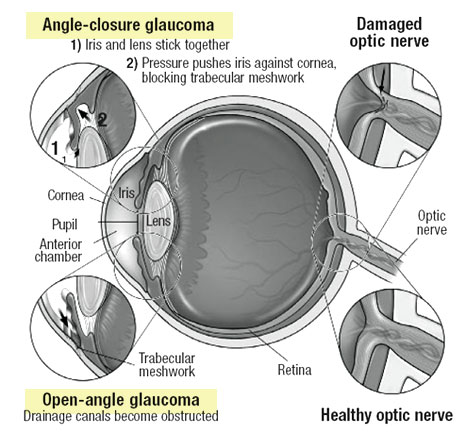}
\caption{Open-Angle and Angle-Closure Glaucoma (\cite{vision}).}
\label{oag}
\end{figure}

Multiple classifications for glaucoma exist, but glaucoma can be generally classified into two large groups: 1) those caused by underlying systemic or ocular disorders (i.e., secondary) and 2) those caused by intrinsic changes in the eye leading to IOP elevation (i.e., primary) and increased susceptibility to IOP damage. Since the former present a combination of multiple diseases with distinct pathophysiological mechanisms, this dissertation focus on primary causes of glaucoma, which can be further classified into two main types based on the anatomical characteristics of the angle between the iris and the peripheral surface of the cornea. The iridocorneal angle is the region where the trabecular meshwork is located and it is responsible to continuously drain most of the aqueous humour which maintains the pressure inside the eye and changes in this structure may increase the IOP and lead to glaucomatous damage (figure \ref{oag}). \cite{morrison2003glaucoma}. 

\subsubsection{Primary Angle-Closure Glaucoma}

Primary Angle-Closure Glaucoma (PACG) is characterized by elevated IOP due to iridocorneal angle blockage, limiting or completely stopping the flow of aqueous humor through the trabecular meshwork \cite{schuster2020diagnosis}. PACG accounts for most ACG cases and is categorized by temporary and sudden apposition of the iris over the angle. leading to a drastic increase in the IOP. Due to the nature of the onset of PACG, this type of Glaucoma may occur suddenly and lead to important clinical manifestations such as ocular pain and redness \cite{schuster2020diagnosis}. 

\subsubsection{Primary Open-Angle Glaucoma}

The large majority of patients who develop glaucoma have Primary Open-Angle Glaucoma (POAG), which is a more silent and slowly-progressive disease. POAG typically affects the optic nerve causing damage followed by visual field loss. It is seen that about 50-60\% of patients develop an initial IOP measurement above 21 mm Hg against the standard population average of 15.7 mm Hg, with some having optic nerve head (ONH) damage with even lower IOP values (\cite{schuster2020diagnosis,dielemans1994prevalence,prum2016primary}). Since this is the most common and important cause of glaucoma worldwide, we focus our efforts on detecting POAG.

\subsection{Common Risk Factors}

Apart from elevated IOP, some other major risk factors for Glaucoma have been identified, such as age, and positive family history for glaucoma (\cite{ramdas2011ocular,ekstrom2012risk,le2003risk,czudowska2010incidence}). Ethnicity has also been found to be a risk factor for Glaucoma, being more prevalent in people self identified as Black or African American (AA) and Hispanics (\cite{tham2014global,racette2005differences,quigley2006number}). Studies focused on identifying risk factors for POAG indicated that advanced age (\cite{leske2007predictors}), elevated IOP (\cite{nouri2004predictive,musch2009visual,founti2020risk,drance2001risk}), smoking (\cite{founti2020risk}), bilateral diseases (\cite{founti2020risk}), and disc hemorrhages (\cite{le2003risk,drance2001risk}) have been associated with faster disease progression. Apart from clinical factors, genome association research showed that genetics also plays a role in glaucoma development and progression (\cite{gharahkhani2021genome}). More recently, degenerative neurological disorders such as Alzheimer's disease, and Parkinson's disease have been suggested as risk factors for Glaucoma (\cite{london2013retina,koronyo2011identification,matlach2018retinal}). 

\subsection{Glaucoma Progression: An Unique Prelude}

Glaucoma is characterized by the loss of RGC, resulting in eye visual field defects. Glaucoma progression, on the other hand, refers to the worsening or advancement of the disease over time. While glaucoma indicates the presence of glaucomatous characteristics such as structural loss or functional impairments, its progression is characterized by active degradation in the RNFL layer, manifesting as an increase in optic nerve damage or a consistent visual field deterioration. Two primary factors differentiating glaucoma progression from glaucoma are the rate of deterioration and treatment strategies. Some patients may have glaucoma that remains stable for years, while others may progress rapidly with a faster vision loss rate. While detecting the disease itself is essential, regular follow-up and assessment are paramount to keep track of visual field deterioration over the patient's lifespan. This allows clinicians to recommend appropriate treatment strategies early on so that it can halt or arrest glaucoma progression. Therefore, it is not only necessary to diagnose glaucoma for appropriate disease intervention but also to monitor its progression to improve the ongoing visual quality of life for patients.

\section{Methods for Glaucoma Diagnosis}

Due to the nature of the disease, patients suffering from acute ACG generally experience pain in the eye, conjunctival hyperemia, nausea, and sudden visual impairment. Immediate treatment is required to prevent ocular damage. In contrast, POAG remains asymptomatic until it reaches an advanced stage and is only detected if the patient's vision deteriorates to a large extent (\cite{crabb2013does,kim2016prevalence}). Since POAG (majority of glaucoma) remains asymptomatic for years, the American Academy of Ophthalmology (AAO) recommends regular eye exams for patients of age 40 onwards (\cite{mowatt2008screening}). Due to the relatively low prevalence of glaucoma, low sensitivity and specificity, and high false positive rates, several tests are devised to diagnose glaucoma.

Eye exam for glaucoma evaluation usually involves Tonometry, Fundus Examination, Perimetry, Gonioscopy, and Pachymetry, each with a specific role in examining the eye.
\begin{itemize}
     \item Tonometry: Since elevated IOP is the main risk factor for glaucoma and used to evaluate treatment strategies, a doctor or technician measures this pressure at a routine checkup by applying pressure to the cornea using a puff of air or the tip of the tonometer probe in contact with the cornea. 
     \item Fundus Examination: Assessment of the retina and optic nerve head in glaucoma involves evaluating the presence of enlarged cup-to-disc ratio (CDR) and diffuse or localized loss of the peripapillary RNFL. A doctor might use a device that magnifies the posterior region of the retina and ONH, directly (ophthalmoscope), fundus photography for documentation and qualitative assessment, and OCT for quantitative analysis.
     \item Perimetry: As patterns of vision loss, usually starting from the periphery, are  characteristic of glaucoma, Computerized Perimetry or a visual field test is used to map the field of vision to determine the degree of vision loss associated to glaucoma.
     \item Gonioscopy: This test is used to evaluate changes in the anatomical features of the angle of the eye and subclassify the disease in different etiology groups, i.e., if the angle between the iris and cornea is closed and blocked (ACG) or wide and open (OAG).
     \item Pachymetry: In this test, a pachymeter is used to measure the thickness of the cornea, which has the potential to influence the IOP readings.
 \end{itemize}

Regular check-ups for Glaucoma include a thorough eye exam and tonometry. Clinical evaluation is followed by the assessment using clinical devices, which falls under two large groups: structural evaluation (fundus photos and OCT) or functional evaluation (perimetry). These are currenty the most effective tools for detecting Glaucoma and have been the mainstay for glaucoma detection. 

\subsection{Structural Tests for Glaucoma}

\begin{figure}[!ht]
\centering
	\includegraphics[width=0.7\textwidth]{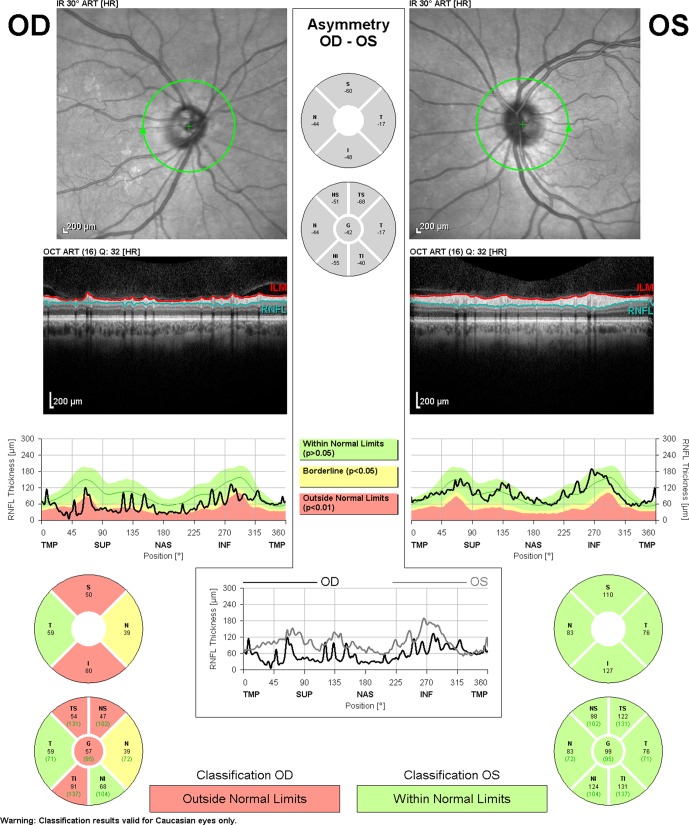}
\caption[Spectralis SDOCT scan report of an eye highlighting glaucomatous damage (\cite{dong2016clinical}).]{A scan report obtained from an eye exam from Spectralis-OCT (Heidelberg Engineering, Heidelberg, Germany) showing glaucomatous damage (\cite{dong2016clinical}).}
\label{sdoct}
\end{figure}

Optic nerve imaging has been the staple for assessment of glaucomatous optic neuropathy (GON). A fundoscopic exam captures the morphological features of the optic disc and RNFL that are linked to glaucoma, such as enlargement of the optic disc cup, localized or difuse thinning of the neuroretinal rim, and RNFL defects (\cite{lucy2016structural}). However, a fundoscopic exam provides only a subjective and qualitative overview of glaucoma more reproducible techniques for evaluation of the optic disc have been developed over the years. Today, OCT has become an essential part of clinical routine eye exam, providing noninvasive objective and quantitative evaluation of the optic nerve head and measurement of the RNFL. There are three types of OCT based on the underlying technology in scanning the Optic Disc and neighboring structures. These are Time-Domain OCT, Spectral-Domain OCT (SDOCT), and more recently the Swept-Source OCT (SS-OCT), of which Spectral-Domain OCT is the most widely-availbale method that produces the most accurate and reproducible results to access glaucoma (Figure \ref{sdoct}; \cite{lazaridis2022deep}).

\subsection{Functional Assessments for Glaucoma}
\label{ss: func}

\begin{figure}[!ht]
\centering
	\includegraphics[width=0.5\textwidth]{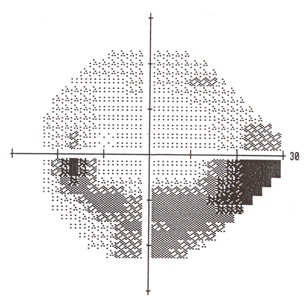}
\caption{A Visual Field Mapping obtained from Visual Field Test (\cite{eyewiki})}
\label{vftest}
\end{figure}

A Functional Diagnosis measures the visual field (VF) impairment due to the loss of optic nerve fibers. As discussed, VF Tests are done using standardized computerized algorithms (Standard Automated Perimetry; SAP), where stimuli are presented to the patient, and their responses are registered. White-on-white SAP is the reference standard to assess visual field loss in glaucoma and changes over time and is measured by mapping the patient's response to a contrast stimulus projected in the eye (Figure \ref{vftest}) (\cite{lucy2016structural}). The assessment is done by measuring the differential light sensitivity on a decibel scale. These values are measured for the entire field of view, mapped into a regular grid, and divided into four regions called quadrants. A major drawback of the VF test is the variability in the measurements due to cognitive function and cognitive load, such as fatigue, distraction, etc., the patient might experience during VF tests. Thus it is recommended that the VF test be done at regular intervals and that visual field defects should be confirmed with a subsequent exam.

\subsection{Structure-Function Relationship}

Studies have shown an association between larger neuroretinal rim area, RGC loss, and VF damage occurring in Glaucoma (\cite{garway2002relationship}). It was found that the relationship between Differential Light Sensitivity (DLS) (dB) and RGC count, and DLS (dB) and neuroretinal rim area follows a curvilinear relation (\cite{garway1997aging,harwerth1999ganglion}). In a study dated Oct 2012, Medeiros et al. showed that this relationship might not be linear nor curvilinear (\cite{medeiros2009prediction}). They used a combination of RNFL assessment with OCT and SAP to show that the amount of neural damage perceived by the OCT and SAP highly depends on the stage of the disease. At early stages, a significant loss in RGC would amount to only a tiny change in Mean Deviation (MD) in SAP which increases as the disease advances. Thus progressive structural changes in the eye are useful indicators of VF loss, and combining VF and OCT would provide a more accurate assessment of rates of deterioration of eye sight early on.

\subsection{Assessment of Glaucoma Progression}

Although many efforts have been made in glaucoma diagnosis, detecting its progression is still challenging. Existing techniques for diagnosis find it hard to distinguish between glaucoma progression, normal age related loss and the variability due to other factors. No unified approach has been established for detecting or evaluating Glaucoma Progression in clinical practice. However, research is underway to develop techniques to assess glaucoma using extensive and complex data. These new techniques use different modalities to detect progression with high accuracy and fewer follow-up periods (\cite{abu2020framework}).

Glaucoma Progression can be grouped into two types based on how tests have accumulated: a) Type of Analysis and b) Unit of Analysis. There are two categories for types of analysis: Trend-based and Event-based analysis and three types for unit of analysis: Global, Sectoral or Pointwise Analysis.

\subsubsection{Trend-based Analysis}

Trend-based analysis for glaucoma determines whether progression is present in a series of functional or structural tests by evaluating longitudinal changes in test parameters over time. Most common methods to evaluate longitudinal changes are by linear regression or some of its variants (\cite{hu2020functional}). It considers both the trajectory and magnitude of change in parameters than only relying on cross-sectional observations thereby providing information not only about the presence of glaucoma progression but also the rate of change over time. This allows clinicians to identify patients who are at a higher risk of rapid deterioration (fast progressors) and adjustment proactively. Trend analysis through linear regression can be done on global indices, cluster of indices are even individual points.

\subsubsection{Event-based analysis}

In event-based analysis, each new measurement is evaluated against two baseline test to determine whether progression (event) is present or not. Progression is identified if the new measurement exceeds the expected test-retest variability and this change persists over multiple tests. For glaucoma progression, search events might include a significant worsening in RNFL thickness, a certain degree of visual field loss, or a change in optic nerve appearance. Once one of these predefined event occurs, it indicates a potential stepwise progression or worsening of the disease.

Similar to trend-based analysis event-based methods rely on global, sectoral or pointwise deviations of multiple sensitivity measurements to define progression. In Global Analysis, all the individual measurements are averaged into a single examination (e.g., MD for SAP). This measurement provides a more concise metric for evaluation. Sectoral analysis, identifies a cluster of regions often dividing into predefined zones (eg. superior, inferior, nasal, temporal etc) where changes are more likely to appear. Pointwise analysis, on the other hand, evaluates all pointwise sensitivity measurements separately to give an estimate of localized changes for a holistic analysis. All these methods have advantages and disadvantages during the examination and must be used carefully.

\subsection{Progression Criteria}
Doctors and Researchers use a set of rules called Progression Criteria to accurately estimate and classify glaucoma progression. Progression Criteria typically employs one or more statistical tests from the diagnostic tests mentioned above to obtain the likelihood and magnitude of glaucomatous change, assuming a null hypothesis of no evolution. Several Subjective and Objective criteria have been developed to identify eyes undergoing glaucoma progression. One such measure is to check for a minimum global RNFL thickness change of more than 0.5 $\mu m/year$, with $p-value < 0.05$, confirmed with two consecutive tests. Similarly, different criteria have been developed for both Structural and Functional assessment of Glaucoma Progression, which uses both Subjective and Objective rules to classify progression in the eye (\cite{thakur2023use}).

\subsection{Challenges in Glaucoma Progression Analysis}

As mentioned earlier, there needs to be standardized criteria to identify glaucoma progression. Doctors and technicians rely on a combination of subjective and objective analysis to assess for progression using clinical, structural, and functional tests which introduces subtle biases in evaluations. Thus an objective standard is necessary to overcome this uncertainty. The most widespread objective standard to detect glaucoma progression has been the GPA used in HFA, which uses a series of functional tests, namely the 24-2 SAP test, to provide inference on progression (\cite{dixit2021assessing,nguyen2019detecting,vianna2015detect}). GPA is a pointwise event-based analysis in which every point in the new VF test is compared with the values from two baseline tests. Points on the VF are flagged with (statistically) significant loss of sensitivity ($p<0.05$) when the measured pointwise pattern deviation becomes more than the expected variability (already derived from a population of stable glaucoma patients). If changes occur at three or more points in two consecutive follow-up tests, the eye is labeled as "possible progression," If these points are repeated in three consecutive tests, the eye is said to be "likely progression" (\cite{giraud2010analysis}). Due to this, GPA becomes a qualitative measure that is relatively simple to implement and accounts for differences in variability associated with VF location, threshold sensitivity, and patient age (\cite{rui2021improving,vianna2015detect}). Although GPA overcomes major challenges faced by subjective analysis, it has the same intrinsic limitations as the SAP test. More details on the GPA is provided in the later sections.

A question remains as to what might be a true objective criterion for Glaucoma Progression. Similar to VF GPA, some commercially available OCT devices provide a pointwise event-based change analysis of the RNFL thickness map and RNFL thickness profile for both global and sectoral RNFL averages (\cite{nguyen2019detecting}). Although OCT GPA is not susceptible to uncertainty due to cognitive stress during tests but significant changes in RNFL thickness do not always translate to true glaucoma progression, due to variability in the measurements and segmentation artifacts. 

Age, another factor, significantly influences the rate at which visual fields and RNFL characteristics change. One of the pioneer studies on glaucoma, the Early Manifest Glaucoma Trial (EMGT), discovered that faster progression is linked with older age (\cite{leske1999early}). This finding is also observed in other research; studies by \cite{vianna2015importance} and \cite{leung2013impact} demonstrated that age is a crucial factor in the deterioration of neuroretinal parameters in healthy individuals. Moreover, age-related changes have been found to skew the progression analysis in glaucoma patients. Subsequent research highlights these findings, indicating that healthy individuals do experience a significant age-related decrease in RNFL over time. Therefore, more accurate estimates of variability or thresholds are necessary to distinguish between age-related RNFL decline and progression from glaucoma (\cite{nguyen2019detecting,vianna2015detect}). In another tangent, the relationship between structure and function also found a similar effect. \cite{zhang2017comparison} showed that structural deterioration (RNFL) loss manifests in early stages, while visual field progression is more prevalent in advanced glaucoma. This discrepancy in the onset of glaucoma progression across various anatomical, functional, and clinical parameters underscores age as an important risk factor in progression analysis.

An ideal method for detecting glaucoma progression should indicate if the eye has glaucoma progression and estimate the rate of deterioration. Previous studies have shown that Joint Longitudinal Modeling can better characterize the relationship between structural and functional tests and improve glaucoma progression detection (\cite{medeiros2011combining,medeiros2012combining,medeiros2012integrating,medeiros2014evaluation}). Joint modeling allowed for a decrease in measurement error as longitudinal changes that would not have been significant in the VF test, might have shown significance in the OCT test (\cite{medeiros2014evaluation,strouthidis2011longitudinal,artes2005longitudinal}). Structural OCT tests have shown to detect glaucoma progression accurately in early stages while the sensitivity of functional VF tests increases in advanced glaucoma (\cite{zhang2017comparison}). The same study showed that although RNFL might not able to detect progression, the ganglion cell complex (GCC) can be useful in all stages. This discrepancy between structure and function, coupled with the potential for combined modeling, highlights a research gap in progression detection. Thus in this research, we utilize the relationship between structural and functional deterioration in eyes to develop models that can detect progression accurately.

\section{Problem Statements and Motivations}

As mentioned before in Section \ref{sec:clinical}, glaucoma, often described as a silent thief of sight, is a group of optic neuropathies characterized by progressive optic nerve deterioration. Early detection and continuous monitoring of its progression are pivotal, not only in preventing irreversible blindness but also in managing the risk of functional impairment to improve the quality of life of affected individuals. While it is understood that functional progression of glaucoma directly impacts patient's vision-related quality of life, structural changes to the ONH and RGC often serve as precursors of these functional outcomes. However, a significant challenge in detecting glaucoma progression is the need for a standardized, unified metric for progression detection. This gap makes it difficult for researchers to track and predict the progressive characteristics across patients consistently. In light of the current advancements in the structural assessment of glaucoma, obtaining accurate and precise measurements of the ONH and RGC has become more accessible than before. But the need for a unified metric is still pressing. Coupled with the availability of functional outcomes of glaucoma progression through visual field tests, it would be valuable to develop algorithms that seamlessly integrate structural or functional data in detection process. Such advancements could not only help clinicians or researchers to pinpoint and predict progressive damage with ease but also reduce the reliance on clinical expertise. Hence, we lay in the following motivations for this research.

\begin{enumerate}
    \item \textbf{Urgency of Early Detection}: Due to the insidious nature of glaucoma, a significant damage to the ONH can occur before any noticeable functional impairment symptoms arise. It is important to detect glaucoma progression early on to allow for timely interventions and treatment that can slow or halt disease progression.
    
    \item \textbf{Structure-Function Relationship}: Research has shown that structural deterioration in the eye due to glaucoma progression often precedes detectable functional changes in vision. Although the exact relationship between structure and function in glaucoma is not yet fully understood, identifying the association between them through modern methods and repeated assessments can help researchers accurately assess glaucoma progression.

    \item \textbf{Precision of Structural Tests}: The Advent of modern imaging technologies for structural tests, particularly OCT, have not only proven to be reliable indicators of glaucomatous changes, but also provides repeated accurate and precise measurements. Analyzing the structural tests can help clinicians accurately identify functional progressive characteristics of glaucoma.

    \item \textbf{Age-Related Progressive Characteristics}: The ONH in the eyes naturally undergoes structural deterioration as individuals age. Since the expected variability in glaucoma-induced changes increases with age, age-related changes can often be confused with the inherent test-retest variability observed in glaucoma progression. Identifying the distinction between progression due to age and progression due to glaucoma is challenging but, at the same time, holds potential for more accurate glaucoma progression insights.

    \item \textbf{Unified Metric for Progression}: The lack of a universal reference standard for glaucoma progression makes consistent detection challenging. By leveraging surrogate methods such as age-related deterioration patterns, we aim to provide a basis for improved detection without the reliance on metrics for progression detection. This approach not only holds promise to achieve results comparable to some standardized methods like the GPA but also has the potential to refine universal detection criteria.

    \item \textbf{Advancements in Medical Image Analysis}: In pursuit to resolve some of the challenges in glaucoma progression detection, we would also confront fundamental challenges in medical image analysis, including imbalance data sets and noisy labels. By utilizing powerful and innovative deep learning methodologies and data driven methods, we aim to resolve these foundational challenges in data and provide holistic methods that can be transferable to other research.
\end{enumerate}

In summary, our motivation for this research stems both from the clinical urgency of early and accurate glaucoma progression detection as well as the promise to resolve some fundamental challenges in medical image analysis and glaucoma research. In light of this, the research objectives can be summarized as follows

\subsection{Research Objectives}

\begin{itemize}
    \item \textbf{Survey of Current Landscape}: Deep dive into the present state of glaucoma progression detection research and identifying the potential of deep learning methods for the same.

    \item \textbf{Data Driven Solution}: Develop, analyze and exploit Glaucoma Progression Data set, comprising of clinical characteristics from different modalities such as OCT scans and VF tests, to understand and predict glaucoma progression more holistically.

    \item \textbf{Deep Learning Technologies}: Develop a series of novel algorithms and models that leverage the spatiotemporal aspects of longitudinal structural assessments of the eye such as OCT scans and the underlying structure-function relationship to predict glaucoma progression.
\end{itemize}

\subsection{Research Contributions}

We will use the motivations drawn from the above research objectives to make the following contributions:

\begin{enumerate}
    \item \textbf{State of the Art Survey}: I conducted an extensive survey of the current state of research in glaucoma progression detection using both traditional and machine learning methods. I cover a comprehensive review of both conventional and emerging technologies to understand glaucoma's structural and functional attributes. This allowed me to pinpoint the promising frontiers of glaucoma research and a potential paradigm shift towards using deep learning assisted imaging techniques for early detection of glaucoma progression.

    \item \textbf{Duke Eye Dataset and Methodology}: I took a data driven approach to gain insights into the structure and function in glaucoma progression. This was achieved by developing, analyzing and utilizing the Duke Eye dataset. Specifically, I aggregated a data set comprising of longitudinal structural assessment of patient's eyes through sequences of OCT scans from the Duke Ophthalmology Registry. A rigorous analysis of the demographic and clinical attributes of this dataset was performed, to provide insights on structural deterioration using RNFL thickness slopes at the observation level. Additionally, a progression dataset was curated by integrating data obtained from the glaucoma registry. This was done by matching the longitudinal OCT scans with the corresponding GPA events derived from VF SAP tests, which served as the objective criteria for glaucoma progression in the subsequent deep learning analysis. All insights and glaucoma research were done on this dataset.

    \item \textbf{Novel Deep Learning Algorithm}: I developed a novel deep-learning algorithm based on CNN-LSTM model. This model was specifically designed to capture the spatiotemporal features from a sequence of longitudinal OCT scans to identify glaucoma progression. Identifying that eyes exhibit age-related structural deterioration naturally and that susceptibility to glaucoma increases with age, this algorithm captures a pseudo-progression criterion in longitudinal structural data. The approach uses reshuffling individual longitudinal OCT scan sequences of patients and the knowledge of healthy eyes from a separate standout data set to establish robust pseudo-progression criteria for glaucoma, eliminating the dependence on gold standard labels for disease progression prediction.

    \item \textbf{Advanced Deep Learning Model with GPA}: Building upon the previous model, I incorporated labels obtained from GPA on the CNN-LSTM model to obtain an accurate detector of glaucoma progression. Given that GPA, although accurate, can have label noise due to its susceptibility to false positives. Owing to the nuances due to data acquisition, the deep learning (DL) model was designed to use age-related structural progression as a pseudo-identifier. A contrastive learning scheme was used to teach the base CNN-LSTM classifier accurate spatiotemporal characteristics of glaucoma progression from potentially noisy labels and highly imbalanced data to provide refined predictions.
    
\end{enumerate}

\section{Thesis Structure}

Chapter 2 provides a literature review and an extensive survey of the current research in glaucoma progression detection. This also covers the recent developments in machine learning (ML) and DL approaches for glaucoma progression. In Chapter 3, we present the Duke Ophthalmic Registry dataset along with basic demographic and clinical characteristics analysis. We also introduce the basic methodology for DL analysis, input features, reference standard, baseline comparative methods and post-hoc analysis. We present our novel weekly supervised deep learning algorithm to detect glaucoma progression using unlabeled longitudinal OCT scans in Chapter 4. This is followed by Chapter 5, in which we develop an advanced deep learning model assisted by noisy labels obtained from GPA to classify disease progression accurately. Finally, we provide the concluding remarks on the current research and discuss the future work in Chapter 6.

\chapter{Literature Review}
\label{chap:guide}

This chapter provides a comprehensive review of the current advancements in glaucoma progression detection. We begin with a meta-review of survey papers and an examination of some landmark studies. This is followed by an introduction to clinical, traditional, and machine-learning methods to detect disease progression, which covers the structural and functional aspects of evaluation techniques. Through this, we aim to learn the latest advancements and the challenges and pitfalls in tracking glaucoma. We then transition to the complex structure-function relationship and emphasize how it has improved the performance of various traditional and machine-learning detection methods. We will conclude the chapter by presenting an in-depth review of the current state of the art in deep learning approaches for detecting progression, which will form the basis of our research.

\section{Background}

Glaucoma, specifically glaucoma progression, has been a significant area of research in Ophthalmology. Recent advancements in artificial intelligence (AI) algorithms and the availability of large medical data sets have shown potential and a growing interest in its application for glaucoma progression detection and management. Numerous studies have tried to explore the effectiveness of AI in improving the diagnosis of glaucoma progression over standard clinical practice and traditional methods. These studies have relied on regular follow-up assessments of several clinical parameters such as demographics (age, sex), electronic health records (EHR; IOP, central corneal thickness - CCT, cup-disc ratio - CDR) data, structural properties (ONH change, RNFL thickness) and functional characteristics (VF sensitivity) to identify glaucomatous change. While these features have improved the predictive accuracy, it still needs to be determined which set of data offers the best representation of glaucoma progression, balancing data acquisition, ease of access, and minimal expertise interpretation. 

Modeling glaucoma progression has been challenging since combining spatial and temporal aspects with a single technique, even with AI, is difficult. Confounding risk factors, test-retest variability, and secondary diseases also make it hard to categorize glaucoma progression. Another contention amongst all studies is the inconsistent definition of progression and how it's measured. Although established methods exist, such as event-based GPA or trend-based point-wise linear regression (PLR), there are no universal criteria to define progression, which makes it difficult to develop AI algorithms for detection. Nonetheless, clinicians have overcome many challenges applying AI, especially deep learning techniques, to predict glaucoma progression using imaging and health records. We provide a brief review of such studies, discussing both the advantages and challenges of AI in glaucoma research shedding light on how AI overcomes limitations. 

\section{Review}

Several surveys have reviewed primary studies focusing on methods used to detect glaucoma progression using either structure, function, or both. These papers present various perspectives on applying AI and deep learning in tracking glaucoma progression. For instance, an article by \cite{thompson2020review} lays a solid foundation, exploring how advances in computing technology and the availability of large datasets have enhanced glaucoma diagnosis. This paper reviewed 91 studies on traditional and machine learning methods, illustrating how the techniques can be integrated into clinical practice. The survey found that while evaluations have become more accurate with imaging data, there remains a drawback in the inconsistent definitions of the reference standard. This underscores the importance of rigorous validations alongside expert opinions before AI techniques are deemed suitable for screening. Although the primary focus of this article is on glaucoma diagnosis, studies on tracking its progression have drawn similar conclusions. These techniques have yet to be incorporated into clinical decision-support systems and require thorough research before implementation.

A review by \cite{mirzania2021applications} on AI for glaucoma detection has emphasized the influence of data quality for improving the DL algorithm's performance. They imply that datasets with better demographic, clinical, structural, and functional representations obtain better performance. In a more specific case, \cite{asaoka2023prediction} have reviewed 108 papers, and \cite{hu2020functional} have reviewed 207 papers on the prediction of functional progression of glaucoma using visual fields. Both these surveys highlight that although VF is highly effective in predicting progression, the importance of model adjustment to specific nuances of experiment design (e.g., irregular testing) before applying AI is paramount. Both papers have outlined that none of the techniques can yet be incorporated in clinical settings due to a lack of validation with a universal ground truth. Similarly, a paper reviewing 30 papers by \cite{bussel2014oct} demonstrating the importance of structural parameters like RNFL from OCT scans in detecting disease progression also fails to identify a common reference standard. Even though RNFL is a precise measure, providing AI methods with accurate estimates for structural progression, it faced challenges like limited reliable datasets, outliers, and the absence of long follow-up data. On the upside, these papers have prompted the use of combined structure and function index for improved glaucoma progression.

A more holistic view of DL approaches to detect and monitor glaucoma progression is provided in the works of \cite{mayro2020impact}, 71 papers and \cite{thakur2023use}, 108 papers. Both articles offer distinct but overlapping perspectives of DL methods using structure, function, and clinical data, highlighting innovative techniques for progression detection. The first paper demonstrates the transformative potential of DL methods to generate new image data, thereby forecasting disease trajectory. Various methods discussed in this paper underscore the utility of DL algorithms in efficiently processing information from multiple modalities, providing accurate predictions. It succeeds in pinpointing several limitations regarding data-centric modeling techniques, critiquing the algorithm's inability to effectively analyze confounding factors, sensitivity to disease manifestations, and image quality. The second paper, on the other hand, adopts a broader approach to analysis, covering various ML and DL methods with multiple modalities (clinical, structure, function) for disease progression. It differs from the first in its ability to highlight variability in results, inconsistencies in datasets, lack of standard progression definitions, and the ability to provide a nuanced view of methods. This paper sets its narrative review specifically for glaucoma progression detection and provides evidence for the translational potential of current research in the field. Although all these papers offer critical reviews of the evolving AI field, laying down the groundwork for research, a more granular analysis that would help clinical understanding and provide a computational and data-centric perspective is necessary.

\section{Landmark Glaucoma Studies}

Several foundational glaucoma studies done over the years have highlighted the importance of early detection and intervention in glaucoma progression. Table \ref{tab:studies} summarizes some landmark studies demonstrating the importance of timely intervention. The seminal work in The Early Manifest Glaucoma Trial (EMGT) showed that individuals are at a higher risk of progression if timely treatment is not provided (\cite{leske1999early}). It is one of the first studies that identified age as a significant risk factor in VF deterioration. The Advanced Glaucoma Intervention Study (AGIS) emphasized the persistent risk of visual field loss throughout the follow-up, even after intervention (\cite{agis}). The Ocular Hypertension Treatment Study (OHTS) proved that the risk of eyes evolving into POAG extended over 15 years, making glaucoma one of the most prolonged chronic progressive diseases (\cite{ohts}). Both AGIS and the OHTS showed that lowering IOP can substantially delay the onset of POAG, underscoring the importance of preemptive care. 

Further studies, like the Collaborative Initial Glaucoma Treatment Study (CIGTS), compared the efficacy of surgical and medical interventions in IOP management and their implications for vision-related quality of life (\cite{cigts}). On a tangential yet similar note, the Collaborative Normal-Tension Glaucoma Study (CNTGS) illustrated that IOP reduction is paramount for therapeutic impact on disease progression. The European Glaucoma Preventing Study (EGPS) supported the findings of other studies by recognizing the influence of IOP as a risk factor (\cite{egps}). The Glaucoma Laser Trial (GLT) by \cite{glts} and the United Kingdom Glaucoma Treatment Study (UKGTS) by \cite{ukgts} affirmed the above studies in determining the efficacy of IOP management through medical interventions. It showed that IOP management is an effective way to reduce the rate of visual field progression during the initial POAG stages. Collectively, these pioneering studies elucidated the relationship between structural nuances such as IOP fluctuations and their effects on functional components such as the visual field deterioration, thereby providing a holistic understanding of components in glaucoma progression. These studies also showed that early glaucoma progression detection and monitoring are pivotal for efficient and targeted therapeutic interventions.

\begin{landscape}
    \small
    \sffamily
    \setlength\tabcolsep{2pt}    
\begin{tabularx}{\linewidth}{@{} *{9}{R} >{\linespread{0.84}\selectfont}R @{}}
    \caption{A brief review of some landmark glaucoma studies}
    \label{tab:studies}        \\
    
    \toprule

    \thead[bl]{Name}   & \thead[bl]{Study\\ Type} & \thead[bl]{Citation}  & \thead[bl]{Design}   & \thead[bl]{Dataset}   & \thead[bl]{Follow-up\\ Period}  & \thead[bl]{Treatment}  & \thead[bl]{Outcome\\ Measures} & \thead[bl]{Findings}   \\
    
    \midrule

    Early Manifest Glaucoma Trial (EMGT)  
    & \cite{leske1999early} 
    & Treatment vs No Treatment 
    & Randomized, Clinical Trial.   
    & 255 Patients with mean age 68.1yrs; 66\% women   
    & \textgreater 4 years    
    & Laser Trabeculoplasty + Betaxolol  
    & VF loss in 3 consecutive C30-2 Humphrey tests or Optic Disc Changes  
    & Higher progression rate in controls group; High IOP frequently followed by progression; Old age was associated with faster progression. \\

    Advanced Glaucoma Intervention Study (AGIS)  
    & \cite{agis}       
    & Treatment vs Treatment    
    & Randomized Clinical Trial  
    & 591 Patients (789 eyes) age 35-80yrs; 54.3\% Female; 56.2\% AA  
    & 7 years & Series of Laser Trabeculoplasty and Trabeculectomy  
    & VF tests using 24-2 Humphrey Field Analyzer and Visual Acuity Tests   
    & Over half the eyes showed VF progression throughout the follow-up after intervention.  \\

    Ocular Hypertension Treatment Study (OHTS) 
    & \cite{ohts} 
    & Treatment vs No Treatment 
    & Randomized Clinical Trial   
    & 1636 Patients age 40-80yrs; 57\% Female; 25\% AA  
    & \textgreater 5 years  
    & Topical Ocular Hypotensive Medication  
    & Reproducible 30-2 Humphrey VF deterioration and reproducible stereoscopic optic disc deterioration  
    & Risk of developing POAG continues over at least 15 yrs of follow-up; High risk individuals more likely to respond to treatment. \\

    Collaborative Initial Glaucoma Treatment Study (CIGTS) 
    & \cite{cigts}  
    & Treatment vs Treatment 
    & Randomized, Controlled Clinical Trial      
    & 607 Patients age 25-75yrs; 45\% Female; 38.1\% AA  
    & \textgreater5 years  & Topical Ocular Hypotensive Medication vs Immediate Trabeculectomy 
    & Sustained Progression in VF loss in 24-2 HFA.  
    & Both treatments effective in slowing down POAG; Bpth group displayed similar rates of VF loss over time.  \\

    Collaborative Normal-Tension Glaucoma Study (CNTGS) 
    & \cite{cntgs} 
    & Treatment vs No Treatment 
    & Randomized, Controlled Clinical Trial.    
    & 145 Subjects age 20-90 yrs;     
    & 5 years    
    & Topical Medication or Laser Trabeculoplasty                       
    & VF Progression and Stereoscopic Optic Disc deterioration  
    & Reducing IOP in patients with Normal-Tesion Glaucoma significantly reduced progression in VF.  \\

    European Glaucoma Preventing Study (EGPS) 
    & \cite{egps}          
    & Treatment vs No Treatment 
    & Randomized, Double-Blinded, Controlled Clinical Trial 
    & 1077 Subjects age $\ge 30$ years; 54.4\% Female  
    & 5 years    
    & Topical Ocular medication                                      
    & IOP, VF Tests and Stereoscopic Optic Discs 
    & Lower IOP associated with reduced risk of glaucoma oset;  Similar clinical characteristics of POAG group with OHTS.  \\

    Glaucoma Laser Trial (GLT)   
    & \cite{glts}    
    & Trearment vs Treatment    
    & Randomized Clinical Trial.   
    & 271 Patients (542 eyes) age \textgreater35 yrs; 56\% Female; 43\% AA 
    & \textgreater2 years (\textless9 years) 
    & Topical Medication vs Laser Trabeculoplasty  
    & IOP, VF Tests and Optic Disc Changes   
    & Laser Trabeculoplasty had significant effect in lowering IOP than eyes with topical medication. \\

    United Kingdom Glaucoma Treatment Study (UKGTS) 
    & \cite{ukgts} 
    & Treatment vs No Treatment 
    & Randomized, Controlled Multicenter Treatment Trial. 
    & 516 Subjects mean age 66 \textpm 11 yrs; 47.1\% Female  
    & 2 years    
    & Topical Ocular Medication   
    & VF 24-2 HFA II Testing, Confocal Scanning Laser Ophthalmoscopy, Scanning Laser Plarimetry, OCT, Monoscopic Optic Disc Photograph, Fundus Photographs, Tonometry 
    & IOP treatment significantly reduced VF progression in early POAG; Pointwise Linear Regression helpful in detecting progression. \\
    
    \bottomrule
    
\end{tabularx}  
\end{landscape}

\section{Glaucoma Progression Detection using Assessments of Visual Fields}

\subsection{Clinical Assessments and Event-Based Methods}
\label{ss:clinicgpa}
Visual field tests are usually used to assess the glaucoma progression leading to functional impairments. VF tests are obtained by measuring the patient's response to contrast stimuli using various perimetric data as shown in Section \ref{ss: func}. It is often measured as the intensity of light stimulus on a logarithmic scale (dB). VF has complex properties; therefore, assessment of progression requires additional analysis of the perimetric data such as clinical gradings, glaucoma change probability scores, event-based analysis, or trend-based analysis (\cite{hu2020functional}). Clinical judgments are subjective and have been found to have moderate intraobserver agreement $(\kappa = 0.45-0.55)$ (\cite{aref2017detecting}). Landmark studies have implemented some form of scoring system to detect progression. The most commonly used and explicit scoring criteria is the glaucoma change probability criteria developed by EMGT (\cite{leske1999early}), which detected early progression. AGIS and CIGTS have used similar methods, which have been repurposed for identifying progression by researchers (\cite{KATZ1999391}). Studies have shown that all these methods have comparable statistics for detecting progression but have various tradeoffs between specificity and sensitivity (\cite{hu2020functional, 10.1001/archopht.117.9.1137}). The Glaucoma Change Probability (GCP) analysis and later the GPA were inspired by the EMGT and is now a commonly used event-based analysis. Research on the comprehensive analysis of the GPA found that there is a strong correlation between GPA and thorough, objective clinical criteria with 93\% sensitivity and 95\% specificity (\cite{arnalich2009performance}), which has shown to be a promising method for detecting VF loss. Some studies showed that GPA was conservative in declaring VF sequences progressing (\cite{roberti2022glaucoma, TANNA2012468, doi:10.5301/ejo.5000193}). These papers showed that event-based GPA has a moderate agreement with expert clinical judgments (\cite{TANNA2012468}). In other studies, the GPA method was found to be susceptible to high false positive rates when predicting "possible progression" in patients with high test-retest variability (\cite{ARTES20142023, 10.1167/tvst.7.4.20}). Future studies have tried to modify GPA, integrate sophisticated analysis, or add structural GPA to improve specificities (\cite{TANNA2012468, medeiros2012integrating, leung2010evaluation}). Ubiquitously, all papers showed GPA could detect early signs of VF deterioration than other methods but are met with the same intrinsic limitations as SAP tests. Authors recommend using a sufficient number of VF tests, which can be time-consuming, require effort, or are prone to test-retest variability.

\subsection{Linear Regression and Trend-Based Methods}

Trend-based analysis has been widely used to detect glaucoma progression utilizing some form of linear regression on VF test parameters or indices. They predict progression by quantifying the VF deterioration as a rate of change over time. This not only helps detect progression but also identifies the onset of progression, rapid progressors, time to change, etc. An eye is classified as progressing using trend-based analysis if the rate of change is negative (usually $<0dB/yr$) and is statistically significant (usually $p<0.05$). Unlike GPA, which generally needs global context for analysis, trend-based methods can be applied to global (MD), sectoral, or individual points (pattern standard deviation - PSD or visual field index - VFI) (\cite{hu2020functional}). Trend-based methods agreed with GPA and clinical judgments (\cite{casas2009visual, medeiros2023validation, nouri2004prediction}). Specialized analysis software using linear regression methods such as PROGRESSOR also showed similar sensitivities for detecting progression (\cite{de2012comparison}). Adding mixed-effects, prior distribution, and clinical priors improved predictive accuracy (\cite{zhang2015log, nouri2004prediction, swaminathan2022rates}). Trend analysis using different parameters of perimetric tests such as VFI, MD, or PSD was also found to correlate well, identifying a similar proportion of eyes progressing. Still, they have intrinsic limitations (\cite{cho2012progression, fitzke1996analysis}). The trend-based analysis detected progression accurately in patients with longer follow-ups, a common pattern observed in multiple studies (\cite{casas2009visual}). When comparing how spatial information helps in detecting progression, generally, PLR or cluster trend analysis (CTA) performed better, but no consensus was observed (\cite{viswanathan1997early, vesti2003comparison, mayama2004statistical}). All these methods had limitations and trade-offs between experiment design, parameters to be used, and test results, often affected by confounding factors. Another study by \cite{quigley1996rate} indicated that linear regression might not be sufficient to detect progression in the presence of bilateral disease. Adding statistical methods into linear regression has been shown to improve sensitivity in VF compared to linear regression and PLR alone. For example, the ANSWERS (Weibull Error Regression) method by \cite{zhu2014detecting} obtains 15\% less prediction error than permutation of PLR (PoPLR). Overall, trend-based methods were moderately accurate, but more emphasis is needed on the experiment design and how to analyze trend data.

\subsection{Probabilistic Methods}

Table \ref{tab:dlfunction} summarizes complex modeling strategies used in glaucoma progression detection. Probabilistic methods provide a framework for modeling and understanding uncertainties in data. In the context of detecting glaucoma progression, these methods offer a robust and adaptive approach, taking into account the inherent variability and intricacies of ocular data.

Probabilistic methods, especially those based on Bayesian frameworks, have been extensively explored for detecting glaucoma progression using visual fields (VF). \cite{medeiros2012combining} showcased the Bayesian Hierarchical Model's effectiveness, which integrates trend and event analyses. This approach significantly outperformed ordinary least squares (OLS) regression, with a 98\% detection rate at 96\% specificity. Concurrently, studies like \cite{warren2016statistical} and \cite{berchuck2019diagnosing} employed Conditional Autoregressive (CAR) models, each showing strong agreements with expert judgment on glaucoma progression. \cite{betz2013spatial} further reinforced CAR's utility, demonstrating it outperforms standard PLR methods. In another study, the Empirical Bayes Estimates of Best Linear Unbiased Predictors (BLUP) introduced by \cite{medeiros2012improved} significantly reduced mean squared error in predicting future VF impairments compared to traditional OLS. In the same domain, \cite{montesano2021hierarchical} proposed a Hierarchical Bayesian Analysis model that showed improved VF progression estimation for glaucoma progression.

Studies also demonstrated probabilistic methods can be repurposed for classification and unsupervised learning. For example, \cite{yousefi2014learning, yousefi2018detection} series of studies from 2014 to 2018 utilized Gaussian Mixture Models (GMM) and Variational Bayesian techniques for detecting progression. These models, especially GMM Expectation Maximization with Permutation of Points (GEM-PoP), were found to detect with high sensitivity and outperformed several baseline and clinical methods in predicting progression. \cite{goldbaum2012progression} introduced a Variational Bayesian Independent Component Mixture Model (VIM), adding valuable clinical insights. Collectively, these probabilistic methods displayed superior performances against established reference standards and highlighted their potential to offer better interpretability and robustness in tracking glaucoma progression compared to conventional methods.

\subsection{Machine Learning Methods}

ML methods have gained significant traction in detecting glaucoma progression using visual fields (VF). \cite{o2012visual} introduced the Permutation Analysis PLR (PoPLR), which reordered permutations to yield a continuous estimate of deterioration while maintaining control over specificity. \cite{shuldiner2021predicting} employed traditional classifiers such as Support Vector Machines (SVM) and Fully Connected (FC) networks to predict rapid progression using initial VF tests, achieving a notable AUC of 0.72. In another study, \cite{jones2019identifying} combined Principal Component Analysis (PCA) with a Soft Voting Classifier, using both longitudinal IOP and VF data, which enhanced accuracy in predicting rapid progression with an AUC of 0.83. \cite{sample2005unsupervised} applied Independent Component Analysis (ICA) in an unsupervised setting, outperforming traditional criteria like AGIS and EMGT. Unique methods such as  Archetypal Analysis were also developed (\cite{wang2019artificial} to identify representative progression patterns, outperforming multiple established methods. \cite{saeedi2021development} used numerous ML classifiers, from Random Forests to CNNs, to distinguish progressing from non-progressing eyes, all of which achieved high sensitivity and specificity, outperforming conventional approaches.

When compared with probabilistic methods, ML techniques, especially when combined with unsupervised learning or ensemble classifiers, are adept at handling the intricacies of VF data for glaucoma progression. While probabilistic methods usually need prior knowledge to incorporate uncertainty, machine learning can learn the complex non-linear patterns directly from the datasets and produces superior performance in terms of accuracy and generalizability for detecting glaucoma progression using VF data.

\subsection{Deep Learning Methods}

DL methods have significantly advanced in detecting glaucoma progression using VFs, particularly by leveraging the data's complex spatial, temporal, or spatiotemporal relationships. Several models, such as CascadeNet-5 utilized by \cite{wen2019forecasting} and the generalized Variational Autoencoder (VAE) by \cite{berchuck2019estimating}, focus primarily on spatial patterns in VF, predicting future Humphrey VF (HVF) points and trajectories of progression with commendable accuracy.

On the other hand, we saw that models that use both spatial and temporal aspects of VF data provided some of the most promising results. For instance, \cite{dixit2021assessing} developed a Convolutional Long Short-Term Memory (ConvLSTM) network, which captures spatiotemporal relationships in VF and demonstrated that combining VF data with clinical measures further improves predictive accuracy. \cite{hosni2023bidirectional} use of Bidirectional Recurrent Model (Bi-RM) and \cite{sabharwal2023deep} implementation of ConvLSTM techniques further emphasized the efficacy of incorporating time-series data in defining VF progression patterns by producing state of the art results.

Unsupervised DL methods are another such technique that is shown to capture subtle characteristics of progression. For example, \cite{berchuck2019estimating} developed a VAE latent space by reconstructing longitudinal VF data to effectively predict rates and trajectories of functional progression. The unsupervised Deep Archetypal Analysis by \cite{yousefi2022machine} is another method that illustrates DL-based unsupervised methods can predict early signs of glaucoma and potential rapid progression accurately.

Compared to the previously discussed probabilistic and traditional ML techniques, DL models, especially those integrating spatiotemporal features, were found to be more holistic and produced a nuanced understanding of glaucoma progression, presenting an essential tool for glaucoma progression detection for researchers.

\subsection{Time Series Analysis and Forecasting Methods}

Time series and forecasting models emphasizing predicting future VF deterioration have also been researched in glaucoma progression. Specifically, models like the intrinsic CAR technique in studies by \cite{warren2016statistical} and \cite{betz2013spatial} were inherently geared towards spatiotemporal data, making it proficient at capturing the intricacies of VF changes over time. \cite{berchuck2019diagnosing} extended the CAR framework and integrated it with Bayesian and Generalized Linear Mixed Models (GLMM) to improve the model's predictive capability, especially highlighting the improvements made by spatiotemporal modeling over strictly spatial ones.

Outside the traditional probabilistic CAR domain, \cite{garcia2019using} employed traditional estimation techniques such as the Kalman Filter Estimator, a classical time-series forecasting approach. This model accounts for measurement uncertainties and inaccuracies, offering holistic forecasting results for disease trajectory, mainly seen with normal tension glaucoma (NTG).

A holistic comparison of all the above methods shows that DL methods provide adaptability and potential in handling large and complex datasets. Probabilistic schemes like CAR and Time-series models like the Kalman Filter provide a more structured, principled approach to predicting progression. Traditional models incorporate domain knowledge well, ensuring outcomes align with known clinical expectations and thus ensuring greater trustworthiness in clinical settings. DL model, on the other hand, although it provides better accuracies, needs to be carefully tuned to obtain desired outcomes.

\begin{landscape}
    \small
    \sffamily
    \setlength\tabcolsep{2pt}    
\begin{tabularx}{\linewidth}{@{} *{11}{R} >{\linespread{0.84}\selectfont}R @{}}
    
    \caption{A Review of Methods for Detection and Prediction of Glaucoma Progression using Visual Field Tests}
    \label{tab:dlfunction} \\
    
    \toprule

    \thead[bl]{Citation}   
    & \thead[bl]{Setup} 
    & \thead[bl]{Algorithm/\\Method} 
    & \thead[bl]{Dataset} 
    & \thead[bl]{Data\\Type}   
    & \thead[bl]{Follow-up\\Period}     
    & \thead[bl]{Reference\\Standard}  
    & \thead[bl]{Baseline\\Method} 
    & \thead[bl]{Model\\Output} 
    & \thead[bl]{Outcomes} 
    & \thead[bl]{Summary}   \\
    
    \midrule

    \cite{medeiros2012integrating}                                                    
    & Regression                
    & Bayesian Hierarchical Model integrating trend and event                       
    & 711 eyes - glaucoma and suspects; 55 eyes - stable              
    & VFI from SAP with GPA prior                    
    & 5 reliable tests, 5 years                                                 
    & 2 consecutive VF test has PSD \textless 0.05 from baseline and GPA           
    & OLS regression slopes                         
    & Bayesian slopes of change for VFI                                           
    & 98\% by Bayesian vs 50\% OLS for 96\% Specificity                               
    & Integrated Bayesian Model significantly better than OLS \\

    \cite{warren2016statistical}                              
    & Regression                
    & Intrinsic Conditional Autoregressive Model                                    
    & Train - 191 eyes, Test - 100 eyes                           
    & 24-2  VF tests from HFA                        
    & 5 follow-ups, 4.34 years                                                  
    & Clinicians Grading and GPA                                                   
    & Regression Models                             
    & Deviance Information Criterion,VF Progression Slopes                        
    & 0.8 AUC, 92\% Sensitivity, 95\% Specificity                                     
    & Model agrees with Progression by Experts \\

    \cite{o2012visual} 
    & Regression                
    & Permutation Analysis PLR                                                      
    & 944 eyes                                                    
    & 24-2 HFA II SAP tests                          
    & 10 exams, 8 years                                                         
    & Slopes \textless{}0 and \textless -1dB/y with p\textless{}0.05               
    & PLR, MD Linear Regression                     
    & PoPLR slopes and p value                                                    
    & 12\%, 29\%, 42\% hit-rate at 5th, 8th and final exams                           
    & PoPLR uses permutation reordering to get continuous estimate of deterioration, allows for control over specificity \\

    \cite{medeiros2012improved}                                            
    & Regression                
    & Empirical Bayes Estimates of BLUP                                             
    & 643 eyes                                                    
    & VFI from SAP Tests                             
    & 10 tests, 6.5 years                                                       
    & significant negative slope (alpha = 0.05)                                    
    & OLS Regression                                
    & Slopes of change with future VFI predicitions                               
    & MSE 32.3 vs 13.9 for Bayesian vs OLS                                            
    & BLUP is beneficial in predicting future impairment \\

    \cite{betz2013spatial}                                                                            
    & Regression                
    & Conditional Autoregressive (CAR) Model                                        
    & 194 eyes                                                    
    & Full SAP threshold, SITA exams                 
    & 7.5 tests, 2.5 years                                                      
    & Clinical judgements on VF reports                                            
    & PLR Slopes of change                          
    & Progression: Significant Slopes of change with varying alpha                
    & 68\% sensitivity, 73\% specificity                                              
    & Model outprforms PLR method \\

    \cite{berchuck2019diagnosing}
    & Regression                
    & Bayesian CAR with GLMM                                                        
    & 191 eyes                                                    
    & VF SAP tests from HFA II                       
    & 7.4 tests, 2.5 years                                                      
    & signifiant LR slope at all VF locations                                      
    & PLR Method                                    
    & Progression: Signigficant LR slopes                                         
    & 0.74 AUC for SpatioTemporal Method                                              
    & Spatiotemporal method better than Spatial method in predicting progression \\

    \cite{montesano2021hierarchical} 
    & Regression                
    & Hierarchical Bayesian Analysis                                                
    & Modeling: 146 eyes, Analysis: 3352 eyes                     
    & 24-2 SAP test from SITA HFA III                
    & 10 VFs, 4 years                                                           
    & P-score \textgreater 0.5, significant negative slopes                        
    & OLSLR and PoPLR                               
    & Progression Score (one sided p-value for slope)                             
    & 57\% Hit-rate at 95\% Specificity                                               
    & Bayesian Model better estimates VF progression \\

    \cite{bryan2017bayesian}                                                               & Regression                
    & Bayesian Hierarchical Model                                                   
    & 276 eyes                                                    
    & 52 points from 24-2 SAP tests from HFA         
    & 10.5 years                                                                
    & -                                                                            
    & -                                             
    & VF Slopes, Posterior Predictiive Checks, Deviance Information Control (DIC) & DIC=1075212, slope = -0.31dB/year for progressing                               
    & Two-stage modeling is beneficial to explore progression \\

    \cite{wen2019forecasting}                                                                          & Regression                
    & CascadeNet-5 and other DL models                                              
    & 8263 eyes (80\% train, 10\% test)                           
    & 24-2 SAP tests from SITA HFA II                
    & 3.6 visits, 3.5 years                                                     
    & Mean and STD of rate of progression from EMGT                                
    & -                                             
    & future HVF points, PMAE                                                     
    & 2.47 dB PMAE, correlation with MD: 0.92                                         
    & DL can predict future spatiotemporal HVF \\
    
    \cite{garcia2019using}                                                                    
    & Regression                
    & Kalman Filter Estimator                                                       
    & 263 eyes NTG, 601 eyes HTG                                  
    & demographics, IOP, VF MD, VF PSD               
    & 5.9 years NTG, 6.3 years HTG                                              
    & -                                                                            
    & Null Model, Linear Regression                 
    & future MD, PSD and IOP values                                               
    & Predictive accuracy: 87.2\% KF-NTG, 86\% KF-HTG vs 86.4\% Null Model, 72.7\% LR & KF forecasts disease trajectory with NTG \\

    \cite{dixit2021assessing}                                              
    & Regression and Classification 
    & Convolutional LSTM with VFI, MD, PLR slopes                                   
    & 11242 eyes                                                  
    & 54 points 24-2 VF tests, IOP, CCT, CDR         
    & 4 follow-ups                                                              
    & statistical significant negative slopes                                      
    & MD Slope, VFI Slope, PLR Slope                
    & stable and progressing                                                      
    & 0.89-0.93 AUC with VF and Clinical, 0.79-0.82 AUC with VF                       
    & ConvLSTM captures spatiotemporal relationships accurately, VF with clinical data more accurate than VF alone \\

    \cite{hosni2023bidirectional} 
    & Regression and Classification 
    & Bidirectional Recurrent Model (Bi-RM)                                         
    & Train: 5413 eyes, Test: 1272 eyes                           
    & 54 points 12-2 HFA using ITT                   
    & 6 tests, 3 years                                                          
    & -                                                                            
    & Linear Regression, Term Memory                
    & progressing vs non-progressing, VF points                                   
    & 92.6\% AUC, 3.61 dB Prediction MAE                                              
    & Bi-RM is predictive of VF progression \\
    
    \cite{hemelings2023predicting}                                                            
    & Regression and Classification 
    & ResNet50 with OLSLR or Huber                                                    
    & Train: 1839, Valid: 272, Test: 271 eyes                     
    & 24-2 VF HFA tests                              
    & 8.5 tests, 5.37 years                                                     
    & significance at different negative slope cutoffs                             
    & -                                             
    & progression: significant negative slopes                                    
    & AUC: 0.67 with OLS at 0.65 with Huber                                           
    & DL predicts VF progression using baseline fundus \\

    \cite{goldbaum2012progression}                                          
    & Classification            
    & Variational Bayesian Independent Component Mixture Model (VIM)                    
    & 2085 eyes                                                   
    & 24-2 SAP test from HFA II, SITA                
    & 6.7 tests, 4 years                                                       
    & GPA outcomes, stereophoto clinical judgements                                
    & GPA outcomes                                  
    & Progression: LR slopes from POP estimates                                   
    & 26.3\% vs 14.5\% accuracy for PGON eyes wrt GPA                                 
    & POP adds information to clinicians to detect VF progression \\

    \cite{sabharwal2023deep}                  
    & Classification            
    & Convolutional LSTM                                                            
    & 8705 eyes (80\% train,10\% valid, 10\% test)                
    & VF SAP tests                                   
    & 12 VF tests, 12 years                                                     
    & event  (GPA) and trend-based (LR) methods                                    
    & clinical assessment (EHR at final VF)         
    & progressing vs non-progressing                                              
    & 0.78-0.94 AUC with all VF                                                       
    & DLM defines VF worsening successfully \\

    \cite{shuldiner2021predicting} 
    & Classification            
    & SVM, FC Network, Random Forest                                                
    & 22925 eyes                                                  
    & 24-2 VF tests                                  
    & 5 VF tests                                                                
    & significant negative MD slope \textless -1dB/y                               
    & Logistic Regression, naive Bayes              
    & rapid and non-rapid progressors                                             
    & AUC: 0.72 with SVM and FC Net                                                   
    & MLA predicts rapid progression with initial VF tests \\

    \cite{saeedi2021development}
    & Classification            
    & MLCs: Logistic Regression, RF, xgBoost, SVC, CNN, FCN                         
    & 131156 eyes (80\% train, 10\% test), 161 eyes clinical test 
    & 24-2 VF SAP from SITA tests                    
    & 6.9 follow-ups, 6.3 years                                                 
    & majority vote by MD slope, VFI slope, AGIS criteria, CIGTS score, PLR, PoPLR 
    & -                                             
    & progressing vs non-progressing                                              
    & 0.83-0.88 sensitivity, 0.92-0.96 specificity                                    
    & MLC had better and balanced predictions than conventional methods \\

    \cite{yousefi2014learning}                                         
    & Classification            
    & GEM + longitudinal slopes                                                     
    & 76 eyes progress, 91 eyes stable                            
    & 52 points from 24-2 SITA SAP tests from HFA II 
    & progressing: 5.5 follow-ups, 2.7 years, stable: 4.7 follow-ups, 4.2 weeks & expert grading of serial stereo photos                                       
    & SAP GPA, VFI LR, MD LR                        
    & stable vs progressed                                                        
    & 28.9\% Sensitivity at 95\% Specificity                                          
    & Accuracy of GEM outperforms baseline and clinical methods \\

    \cite{jones2019identifying}                                           
    & Classification            
    & PCA + Soft Voting Classifier                                                  
    & 571 patients (80\% train, 10\% test)                        
    & IOP, MD, PSD data                              
    & 6 years                                                                   
    & statistical significant negative slopes \textless -1dB/y                     
    & -                                             
    & rapid and non-rapid progressors                                             
    & 0.83 AUC for predicting rapid progression                                                      
    & IOP and VF data improves accuracy for Rapid Progression \\

    \cite{yousefi2018detection}
    & Unsupervised              
    & Gaussian Mixture Model                                                       
    & 2085 eyes - glaucoma                                        
    & VF 24-2 SAP tests                              
    & 5 follow-ups, 9 years                                                     
    & PSD \textless 0.05 in SAP tests wrt Baseline and reproducible at least once  
    & Global, Region, Point wise Linear Regression  
    & Time to detect progression using ML slope and p-value                       
    & 3.5 years vs 3.9 years to detect progression                                    
    & ML outperformed other methods \\

    \cite{wang2019artificial}                                       
    & Unsupervised              
    & Archetypal Analysis                                                           
    & method: 12217 eyes + validation: 400 eyes                   
    & 24-2 VF SAP Tests                              
    & 5 follow-ups, 5 years                                                     
    & Event-based and trend-based clinical assessment                              
    & MD Slope, AGIS score, CIGTS score, PoPLR      
    & Rate of Archetype change ONL                                                
    & 51\% accuracy and 77\% correct rejection rate                                   
    & Achetype significantly ouperformed all methods \\

    \cite{sample2005unsupervised}                           
    & Unsupervised              
    & Independent Component Analysis (ICA)                                                
    & 191 eyes - glaucoma and suspect                            
    & 24-2 and 30-2 VF SAP Tests                     
    & 3 follow-ups, 6.24 years                                                  
    & Clinical assessment of Stereoscopic Photos + PGON                            
    & AGIS and EMGT scoring                         
    & Progressed vs Non-progressed                                                
    & 16.7\% progressing, 31\% PGON                                                   
    & vB-ICA outperforms AGIS and EMGT criteria \\
    
    \cite{berchuck2019estimating}                          
    & Unsupervised              
    & generalized Variational Autoencoder                                           
    & 3832 eyes (80\% train, 10\% valid, 10\% test)               
    & 24-2 SAP VF Test                               
    & 7.61 visits, 4.95 years                                                   
    & significant negative rate of change (alpha = 0.05)                           
    & PW Linear Regression and OLSLR slopes with MD & 24-2 SAP VF points, rates of change using latent dimension                  
    & MAE: VAE 5.14dB vs PW 8.07dB                                                    
    & VAE can predict rates ad trajectories of progression \\

    \cite{yousefi2022machine}                             
    & Unsupervised              
    & Unsupervised Deep Archetypal Analysis                                         
    & 205 eyes from OHTS                                          
    & 30-2 VF from SITA HFA                          
    & 16 years                                                                  
    & OHTS expert-identified patterns                                              
    & -                                             
    & MD of 18 VF clusters, GEE LR slopes                                         
    & -2.7 dB MD at glaucoma, -5.5dB MD at last visit                                 
    & Automated ML system predicts early sign and rapid progression \\

    \cite{yousefi2016unsupervised}             
    & Unsupervised              
    & GEM Progression of Patterns and VIM 
    & 2143 eyes                                                   
    & 52 points in 24-2 SITA SAP tests               
    & progress: 14 follow-ups, 9.1 years                                       
    & expert grading of serial stereo photos                                       
    & signifiacnt PoPLR, MD LR, VFI LR slopes      
    & stable vs progressed                                                        
    & AUC: 0.86 for GEM-PoP,  0.82 VIM-Pop                                            
    & GEM-PoP was significantly more sensitive to PGON \\                                                                 
    \bottomrule
    
    \end{tabularx}  
\end{landscape}

\section{Detection of Structural Progression}

Several methods have been developed for detecting structural progression in glaucoma by objectively and quantitatively measuring the anatomical changes in the eye, particularly the ONH and the RNFL. Comparison of advanced imaging techniques for glaucoma progression, such as OCT, scanning laser polarimetry (SLP), confocal scanning laser ophthalmoscopy (CSLO), etc., have been done previously in \cite{miki2012assessment}. The study found that although precision in the measurement of structure is generally good with imaging techniques, there was no uniform agreement regarding the most appropriate evaluation method. Several studies have shown and argued the usefulness of OCT in glaucoma progression detection due to its ability to visualize the retinal substructure (\cite{geevarghese2021optical, bussel2014oct, abe2015use, tatham2017detecting}). This is because of the ability of OCT to focus on the circumpapillary RNFL (cpRNFL) thickness measurements, which are the most widely used parameters in clinical practice. \cite{abe2015use}, \cite{tatham2017detecting} and \cite{kotowski2011clinical} further reviewed different OCT assessment techniques and found that the SDOCT has more precision and reproducibility of RNFL measurements than other OCT methods, which is quintessential to evaluate glaucoma progression, with consistently high sensitivity in detection. Detection of glaucoma progression using different imaging techniques such as SLP from GDx-VCC software (\cite{dada2014scanning}), CSLO from Heidelberg Retina Tomograph (HRT) (\cite{maslin2015suppl}) exist. Still, again, agreement on the methods needed to be found. Although not widely researched, researchers have used fundus photos for glaucoma progression detection and found appreciative results (\cite{medeiros2021detection}). These methods highlight various challenges in diagnosing glaucoma progression based on structural changes. These challenges include differentiating between age-related variations and actual glaucomatous changes, a decreased detection rate in advanced glaucoma (floor effect), high costs of tests, and the need for specialized clinical expertise to operate the equipment. Nevertheless, structural diagnostic methods continue to be extensively researched because they provide detailed information about the RNFL in Optic Discs. In the subsequent sections, we will explore various methods and techniques researchers and clinicians have developed to identify structural changes in glaucoma progression.

\subsection{Clinical and Conventional Methods}

In clinical practice, experts identify glaucoma progression by analyzing changes in stereoscopic optic disc photographs. However, there's often only poor to moderate agreement in assessments between different experts (\cite{o2010glaucomatous}). Like event and trend-based analysis in visual fields, objective methods have also been used to evaluate glaucoma progression using structural parameters. Similar to VF, GPA can be translated to structural progression by measuring the RNFL test-retest variability with baseline measurements where progression means exceeding a predetermined criterion of this change. Research by \cite{kaushik2015comparison}, which compared GPA using SAP parameters vs GPA using OCT measures, found that OCT detects progression in early glaucoma better but performs poorly in advanced stages with overall RNFL GPA obtaining lower sensitivity than SAP GPA. Because of these inconsistencies and the continuous advancements in OCT technologies, the application of GPA for structural evaluation is rare, and there is limited research in this area.

Trend analysis is another paradigm often explored for evaluating glaucoma progression due to its ease of implementation and ability to track both presence and rate of progression. A study by \cite{lin2017trend} showed that cpRNFL rates of change using the PLR are informative of VF loss in glaucoma but not necessarily associated with VF defects. Several other research showed promise for ganglion cell-inner plexiform layer (GCIPL) thinning rates being applicable for objective assessment of glaucoma progression (\cite{lee2017trend, wu2020wide}). Another research comparing trend-based progression analysis (TPA) with RNFL GPA found higher sensitivity for the TPA method - 48.8\% vs 27.1\% at 84.2\% and 81.7\% specificities respectively (\cite{yu2016risk}). Using global and cluster-wise linear regression, \cite{lee2011trend} showed that the localized OCT RNFL thinning rate also indicates progressive loss with high sensitivity (62\% at 95\% specificity). In other research by \cite{thompson2021agreement}, a poor agreement was found between trend-based and qualitative assessment ($\kappa = 0.0135-0.1222$), so the performance of trend methods can be debated. Like functional progression, trend analysis using structure also has the same limitations, requiring longer follow-ups for accurate modeling (\cite{miki2012assessment}) and is susceptible to age-related loss (\cite{jammal2020effect,leung2013impact}), etc.

\subsection{Probabilistic Methods}

Table \ref{tab:dlstructure} summarizes some methods to detect glaucoma progression using structural parameters using advanced techniques, specifically focusing on structural progression. In the following sections, we will discuss some of these sophisticated methods and draw insights into how structural assessments have shaped glaucoma progression detection.

Regression-based approaches have been studied in detail for predicting progression. A study by \cite{nagesh2019spatiotemporal} utilizing the Continuous Time-Hidden Markov Model (CT-HMM) on a dataset of 135 eyes with SDOCT images showed impressive results predicting future RNFL with a mean absolute error (MAE) of 3.48 vs 4.06 for VFI over 4.9 years follow-up. This method notably surpassed the performance of Linear Regression predictions. In \cite{mohammadzadeh2021estimating}, a Hierarchical Longitudinal Bayesian Regression was applied on 112 eyes for predicting GCC thickness, resulting in a notable negative correlation between slope and baseline ranging from $r = -0.43 to -0.50$ over 2-4.2 year follow-up. In another study, \cite{su2023spatially} incorporated spatially varying hierarchical random effects on 111 eyes to predict progression, emphasizing that including visit effects reduced estimation errors. This study demonstrated that the new model obtains an accuracy of 21.4\% vs. 18\% to detect significant negative slopes over the same follow-up duration with linear regression.

Belghith et al. explored several Bayesian methods to develop classification-based models for glaucoma progression. In a study, \cite{belghith2014joint} utilized the Fuzzy Bayesian Detection Scheme (FBDS) on a training set of 25 eyes and a test set of 117 eyes, achieving a 64\% sensitivity at 94\% specificity over three years. \cite{belghith2014glaucoma}, in another study, showed an improved 70\% sensitivity at 94\% specificity in its study over 2.2 years, indicating the Bayesian Fuzzy Detection Scheme performs better in shorter follow-ups. The Variational Change Analysis, as used in \cite{belghith2013glaucoma} on a dataset of 267 eyes, obtained a sensitivity of 86\% at 96\% specificity in the yet shorter follow-up of 0.5 years. In another research, \cite{belghith2015learning} explored the Bayesian-Kernel Detection Scheme and demonstrated a 78\% sensitivity at 94\% specificity for non-progressing conditions in its 117-eye dataset over 1.7 years.

Some unsupervised models were also investigated for detecting progression. \cite{yousefi2015unsupervised} applied GEM on a large dataset of 2274 eyes with a short five-week follow-up and achieved a sensitivity of 78\% at 95\% specificity. Meanwhile, \cite{huang2021detection} merged GEM with longitudinal slopes on another massive dataset (3485 eyes), reported a 38.6\% accuracy, and showed that the ML model generally outperforms linear regression over numerous follow-ups.

The various probabilistic methods detailed in the studies for detecting structural glaucoma progression have demonstrated considerable promise. Many of these methods outperformed traditional linear regression models, showcasing advancements in prediction accuracy, sensitivity, and specificity. Regression and classification techniques achieved impressive results on diverse datasets over varying follow-up durations. These structural assessments are crucial as they offer direct, quantitative insights into glaucoma progression, which can be complementary to functional assessments. While functional assessment remains the clinical gold standard for monitoring glaucoma progression, these probabilistic structural assessment techniques underscore the potential of providing an earlier and possibly more nuanced understanding of disease progression. Combining structural and functional assessments may offer the most comprehensive view of glaucoma progression as technology and algorithms advance.

\subsection{Machine Learning Techniques}

Machine Learning methods have shown significant potential in detecting glaucoma progression through structural assessments. A study by \cite{balasubramanian2012localized} used Proper Orthogonal Decomposition (POD) for regression on topographic measurements of HRT II from 246 eyes. This approach could predict progression with high sensitivity and specificity, even for a smaller follow-up. Another study by \cite{christopher2018retinal}, who applied unsupervised learning using PCA combined with Logistic Regression on swept-source OCT (SSOCT) images, showcased impressive AUC results, demonstrating its efficiency in identifying structural progression. \cite{mohammadzadeh2022detection} similarly employed a multi-layer perceptron (MLP) classifier on macular OCT Angiography (OCTA) images and showed that it outperforms logistic regression, emphasizing machine learning's capability in improving progression detection.

Comparatively, machine learning methods learn efficiently from intricate data patterns, allowing for potentially higher predictive accuracy than probabilistic methods. ML methods were found to be more proficient in handling complex datasets and generally required lower follow-up time. Unlike the previously discussed probabilistic methods, which primarily depend on statistical models, ML techniques can easily extract meaningful features from the data. However, a potential limitation might be that they require larger datasets and computational power. While both probabilistic and ML methods offer valuable insights, integrating both could lead to more holistic and accurate detection of glaucomatous progression.

\subsection{Deep Learning Approaches}

Many Deep learning methods have been developed in glaucoma progression detection, particularly in structural assessments. These methods are more intricate than probabilistic and traditional ML models, and their results often exceed the latter in both learning from complex data and accuracy.

CNNs are among the most common choices for researchers to learn spatial representations from data. \cite{mohammadzadeh2022detection} implemented a CNN classifier on macular OCTA images from 134 patients taken over four visits (2.4-5.5 years) and achieved an AUC of 0.81 vs 0.66 in logistic regression, a significant performance improvement. Similarly, \cite{mariottoni2023deep} applied a CNN with an FC network on cpRNFL thickness from SDOCT from 816 eyes six visits (in 3.5 years) and obtained an impressive 87.3\% sensitivity at 86.4\% specificity. \cite{mandal2023noise} developed a CNN-LSTM model based on SDOCT B-scans from 3253 eyes across five follow-ups, predicting progression with a 48\% sensitivity at 95\% specificity.

In a more specialized case, DL models have been modified to use structural information uniquely to draw insights and improve accuracy. \cite{medeiros2021detection} developed a ResNet50 M2M model that predicted RNFL thickness from fundus photos trained on a dataset with 8831 eyes (6.2 visits in 5.5 years). The subsequent trend-based analysis using the RNFL thickness measure achieved an impressive AUC of 0.86. In another study, \cite{hassan2020conditional} used a Conditional Generative Adversarial Network (GAN) on macular OCT volume scans to predict future glaucoma development in OCT scans in 109 eyes. Researchers found that the generative model could obtain better reconstruction on the 3rd visit than on the 2nd one. \cite{HOU2023854} implemented a Gated Transformer Networks (GTN) that analyzed RNFL measurements from OCT scans of 4211 eyes and achieved an AUC of 0.97 with the Majority Voting scheme. In another research, \cite{bowd2021individualized} integrated Deep Learning Autoencoders with cpRNFL thickness map data from the OCT, which was able to identify progression with 90\% sensitivity, demonstrating that focusing on specific regions of interest can notably improve predictive accuracy.

DL methods often exhibit superior performance and efficiency in processing large complex datasets compared to probabilistic and ML techniques. Their capacity to extract complex salient features from image datasets is a distinctive advantage. However, a limitation of this method is that it often requires large datasets to generalize and significant computational power to train. Nevertheless, with specialized models like M2M, Conditional GAN, GTN, etc., DL methods can replace conventional glaucoma progression detection practices, potentially minimizing the need for frequent expert evaluations.

\subsection{Time Series Approaches in Structural Assessment}

Advanced algorithms and models are rapidly becoming essential in glaucoma progression detection. For instance, the Gated Transformer Networks (GTN) from \cite{HOU2023854} utilizes OCT scans to offer holistic insight into progression, signifying the growing importance of transformer architectures in medical applications. The LSTM approach in \cite{mandal2023noise} also stands out by predicting progression using SD-OCT B-scans, highlighting the recurrent model's ability to capture temporal dependencies in ophthalmic data. These sophisticated models underscore the potential of modern DL architectures to provide both timely and accurate glaucoma progression assessments by efficiently processing time-series information.

\begin{landscape}
    \small
    \sffamily
    \setlength\tabcolsep{2pt}    
\begin{tabularx}{\linewidth}{@{} *{11}{R} >{\linespread{0.84}\selectfont}R @{}}
    
    \caption{A Review of Methods for Detection and Prediction Glaucoma Progression using Structural Assessments}
    \label{tab:dlstructure} \\
    
    \toprule

    \thead[bl]{Citation}   
    & \thead[bl]{Setup} 
    & \thead[bl]{Algorithm/\\Method} 
    & \thead[bl]{Dataset} 
    & \thead[bl]{Data\\Type}   
    & \thead[bl]{Follow-up\\Period}     
    & \thead[bl]{Reference\\Standard}  
    & \thead[bl]{Baseline\\Method} 
    & \thead[bl]{Model\\Output} 
    & \thead[bl]{Outcomes} 
    & \thead[bl]{Summary}   \\
    
    \midrule

    \cite{nagesh2019spatiotemporal}     
    & Regression     
    & Cont. Time - Hidden Markov Model                      
    & 135 eyes                                                               
    & SDOCT Images and Avg RNFL Thickness                 
    & 7.9 visits, 4.9 years                
    & VFI Progression                                           
    & Linear Regression Predictions                          
    & Change in RNFL Thickness Maps and VFI State                      
    & 3.48 MAE RNFL and 4.06 MAE VFI                                  
    & OCT CT-HMM significantly outperforms LR and Avg RNFL CT-HMM  \\
    
    \cite{asaoka2021joint}             
    & Regression     
    & Latent Space LR - Deep Learning                       
    & Cross-Sectional: 746 eyes, Longitudinal: Train 998 eyes, Test 148 eyes 
    & OCT Image data: GCC thickness, macular RNFL, OS+RPE 
    & 8 VF tests, 5.6 OCT tests, 5.9 years 
    & RMSE and Signigicant Slopes in LMM                        
    & MLR, SVM, DL, CNN-TR, PLR, DLLR                        
    & 68 points in HFA 10-2 test, 52 points in HFA 24-2 test           
    & RMSE cross-section 6.4dB, longitudinal 4.4dB and 3.7dB          
    & LSLR-DL predicts both cross-sectional and longitudinal VF \\

    \cite{medeiros2021detection}        
    & Regression     
    & ResNet50 (M2M model)                                  
    & 8831 eyes (Test: 1147 eyes)                                            
    & Color Fundus, SDOCT RNFL thickness                  
    & 6.2 visits, 5.5 years                
    & statistical significant negative slopes                   
    & -                                                      
    & SDOCT global RNFL measurements, rate of change in RNFL thickness 
    & AUROC 0.86 for predicting progressors                           
    & M2M predicts RNFL thickness and can monitor progression \\

    \cite{balasubramanian2012localized} 
    & Regression     
    & Proper Orthogonal Decomposition                       
    & 246 eyes                                                               
    & Topographic measurements of HRT II                  
    & 4 follow-ups, 4.1 years              
    & SAP GPA or stereo-photo assessment                        
    & Topographic Component Analysis significant change      
    & Progression: follow-ups observed positive rate greater than OPR  
    & 100\%, 78\%, 78\% sensitivity, 0\%, 86\%, 86\% specificity      
    & POD with k-family-wise error rate reduces number of follow-ups to predict progression \\

    \cite{mohammadzadeh2021estimating}  
    & Regression     
    & Hierarchical Longitudinal Bayesian Regression         
    & 112 eyes                                                               
    & GCC thickness from Macular OCT Scans                
    & 4 tests, 2-4.2 years                 
    & significant negative slope at Bayesian p\textless{}0.025  
    & -                                                      
    & GCC thickness estimates, macular slopes                          
    & Negative correlation between slope and baseline: -0.43 to -0.50 
    & Bayesian method is efficient method for estimating macular rates \\

    \cite{su2023spatially}              
    & Regression     
    & Spatially Varying Hierarchical Random Effects         
    & 111 eyes                                                               
    & GCC thickness from Macular OCT Scans                
    & 4 tests, 2-4.2 years                 
    & significant slope when 95\% CI is less or greater than 0  
    & Simple Linear Regression                               
    & GCC thickness estimates, macular slopes                          
    & 21.4\% vs 18\% signifiance negative slopes                      
    & Including visit effects reduces estimaton errors \\

    \cite{belghith2014joint}            
    & Classification 
    & Fuzzy Bayesian Detection Scheme                       
    & Train - 25 eyes, Test - 117 eyes                                       
    & 3D SD-OCT Images                                    
    & 3 tests, 3 years                     
    & stereo photograph grading and VF GPA                       
    & RNFL-SVM and RNFL-ANN                                  
    & Progressing ns Non-progressing                                   
    & 64\% Sensitivity at 94\% Specificity                            
    & FBDS using image features outperforms ANN and SVM using RNFL classifiers \\

    \cite{belghith2014glaucoma}         
    & Classification 
    & Bayesian Fuzzy Detection Scheme                       
    & 117 eyes                                                               
    & 3D SDOCT voxel images                               
    & 2 follow-ups, 2.2 years              
    & EMGT Criteria and Progression in stereophoto              
    & RNFL-SVM, RNFL-ANN, MRF                                
    & progressor or non-progressor                                     
    & 70\% sensitivity at 94\% specificity                            
    & BFDS has higher diagnostic accuracy  \\

    \cite{belghith2013glaucoma}         
    & Classification 
    & Variational Change Analysis with Markovian-a-priori   
    & 267 eyes                                                               
    & Heidelberg Retina Tomograph (HRT II)                
    & 4 follow-ups, 0.5 years              
    & progressing by stereophoto or VF change                   
    & Topographic CA, VCA                                    
    & progressing vs non-progressing                                   
    & 86\% sensitivity, 96\% specificity                              
    & Detection formulated as missing data problem. VCA-MA outperforms other methods \\

    \cite{li2022deep}                   
    & Classification 
    & DiagnoseNet and PredictNet                            
    & 3003 train, 422 valid, 337 test for progression                        
    & Color Fundus Photographs                            
    & 34.8-41.7 months                     
    & 3 VF points worse than 5\% baseline in 2 consecutive test 
    & -                                                      
    & progressing vs non-progressing                                   
    & 82\% Sensitivity at 59\% Specificity                            
    & DL model useful in early detection of glaucoma progression \\

    \cite{belghith2015learning}         
    & Classification 
    & Bayesian-Kernel Detection Scheme (BFDS)               
    & 117 eyes                                                               
    & 3D SDOCT volume scans                               
    & 2.21 tests, 1.7 years                
    & Longitudinal VF testing, stereophoto assessment           
    & RNFL-SVM, RNFL-ANN, KDS, KDS, GBKDS, RBF-BKDS          
    & progressing vs non-progressing                                   
    & 78\% Sensitivity, 94\% Specificity for non-progressing          
    & BKDS outperforms other methods, Only healthy and non-progressing eyes can produce high accuracy \\
    
    \cite{mariottoni2023deep}           
    & Classification 
    & CNN and FC Model                                      
    & 692 stable, 124 progressing eyes                                       
    & RNFL thickness peripapillary SDOCT                  
    & 16.2 tests, 6 visits, 3.5 years      
    & Clinical judgements on longitudinal SDOCT reports         
    & trend-based analysis                                   
    & progressing vs non-progressing                                   
    & 87.3\% sensitivity, 86.4\% specificity                          
    & DL model agreed with experts, provided likely location of change \\

    \cite{mohammadzadeh2022detection}   
    & Classification 
    & CNN and MLP classifiers                               
    & 134 patients                                                           
    & macular OCTA images                                 
    & 4 visits, 2.4-5.5 years              
    & significant negative MD slope                             
    & Logistic Regression                                    
    & progressing vs non-progressing                                   
    & AUC: 0.81 DL vs 0.66 with LR                                    
    & DL could extract and enhance progression detection \\

    \cite{HOU2023854} 
    & Classification 
    & Gated Transformer Networks (GTN) 
    & 4211 eyes (2666 patients) 
    & OCT scans 
    & 5 tests, 1.2 - 4.7 years 
    & significant negative slopes in SAP trend-based methods, SAP GPA 
    & LMM, naive Bayes Classifiers 
    & progressing vs non-progressing 
    & AUC: 0.97 with Majority Voting (M6) 
    & GTN outperforms conventional methods; Ensemble methods improve performance \\

    \cite{mandal2023noise}              
    & Classification 
    & CNN-LSTM Classifier                                            
    & 3253 eyes                                                              
    & SD-OCT B-scans                                      
    & 5 follow-ups, 3.1 years              
    & -              
    & OLSLR on global mean RNFL thickness                                                  
    & progressing vs non-progressing                                  
    & 48\% sensitivity at 95\% specificity                            
    & DL model identifies structural progression from age-related changes without reference standard \\

    \cite{hassan2020conditional}        
    & Generative     
    & Conditional GAN: 3DCNN, UNet, PatchGAN                
    & 109 eyes                                                               
    & macular OCT volume scan                             
    & 4 tests, 6 months apart             
    & -                                                         
    & -                                                      
    & OCT Macular Volume Scan, SSIM, PSNR                             
    & 0.8325 SSIM with 3 visits, 0.8336 with 2                        
    & GAN can predict future glaucoma development in OCT scans \\

    \cite{bowd2021individualized}       
    & Unsupervised   
    & DL Autoencoders                                       
    & 44 progressing, 303 nonprogressing, 109 healthy eyes                   
    & cpRNFL thickness maps from OCT                      
    & 4 visits, 2.9 years                  
    & significant LMM slope for Region of Interest cpRNFL       
    & global slopes from LMM                                 
    & Progressing: significant LMM slopes                              
    & 90\% sensitivity, progression slope -1.28 {\textmu}m/y                  
    & ROI from DL-AE can boost progression accuracy \\

    \cite{christopher2018retinal}       
    & Unsupervised   
    & Principle Component Analysis with Logistic Regression 
    & 179 eyes                                                               
    & SSOCT images                                       
    & 7-8.7 tests, 1.7-2.2 years           
    & significant negative slope                                
    & LR with cpRNFL, SAP MD, FDT MD & progressng vs non-progressing  
    & AUC: 0.95 for RNFL PCA                                          
    & Computational method can identify structural progression \\

    \cite{yousefi2015unsupervised}      
    & Unsupervised   
    & Gaussian Mixture Model (GEM)                          
    & 2274 eyes                                                              
    & RNFL Thickness from SD-OCT                          
    & 5 tests, 5 weeks                     
    & event-based SAP GPA                                                   
    & Linear regression                                      
    & progressing vs non-progressing                                   
    & 78\% sensitivity, 95\% specificity      
    & GEM predicts RNFL patterns for glaucoma progression \\

    \cite{huang2021detection}           
    & Unsupervised   
    & GEM + Longitudinal Slopes                             
    & 3485 eyes                                                              
    & RNFL thickness maps from OCT scans                  
    & 9 follow-ups                         
    & statistical significant negative slopes                    
    & Linear Regression                                      
    & stable vs progressed                                             
    & 38.6\% accuracy with ML model                                   
    & ML model predicts progression better than LR  \\                                                

    \bottomrule
    
    \end{tabularx}  
\end{landscape}

\section{Structure-Function Relationship in Glaucoma Progression Detection}

\subsection{Clinical Methods for the Assessment of Glaucoma Progression Using Structure and Function}

As discussed previously, several studies indicated that OCT is more sensitive than VF in the early stages of glaucoma progression detection, but this sensitivity decreases as progression advances (\cite{zhang2017comparison, abe2016relative}). Some research (\cite{swaminathan2021rapid, gracitelli2015association}) exploring the effect of progressive RNFL loss found an association with loss in visual fields. Similar conclusions were drawn from studies where changes in macular thickness were used to find an association with central visual field loss (\cite{mohammadzadeh2020longitudinal}). Another research by \cite{suda2018evaluation} found an appreciative correlation (R=0.589) between SAP and OCT results for glaucoma progression. In a more novel approach, both the structure and function were used to obtain rates of change of RGC count as indicators for neural damage in glaucoma progression, paving the way for combined structure-function index (\cite{medeiros2012structure, hirooka2016estimating, wu2021simplified}). A survey by \cite{lisboa2013combining}, which analyzed several studies in the same field, found that a combined approach is more effective in detecting glaucoma progression even though there can be disagreements between detection using structural or functional measures alone. \cite{gardiner2012effect} further found that even though the structure-function is still affected by inter-test variability, it is still feasible to combine them for progression assessment.

\subsection{Sophisticated Methods to Detect Glaucoma Progression Using Combined Structure and Function}

Studies in probabilistic methods for glaucoma progression detection using combined structure-function demonstrated improved predictive accuracy in general (Table \ref{tab:dlstructfunc}). \cite{medeiros2011combining}, showed in their research using the Bayesian Hierarchical Model on 434 glaucoma-suspected eyes that probabilistic methods obtain higher sensitivity with VFI  and temporal, superior, nasal, inferior, and temporal (TSNIT) RNFL averages than traditional OLS regression standards. Similar results were also seen in \cite{medeiros2012combining}, which used the Bayesian Joint Regression Model and applied to 242 eyes, which outperformed the OLS linear regression in predicting MD and rim area (RA) from SAP and CSLO data. In another study, \cite{russell2012improved} implemented Bayesian Linear Regression on 179 eyes, which indicated higher performance for short time series, but traditional OLSLR was more accurate for longer follow-ups when analyzing VF. Overall, Bayesian methods have shown superior performance, though the input data type and follow-up time might influence detection accuracy for progression.

Following the studies on probabilistic methods, researchers have focused on developing innovative statistical methods for detecting glaucoma progression, leveraging both structural and functional indicators. In research by \cite{bilonick2008evaluating}, Latent Class Regression was developed, which was able to identify accurately baseline RNFL characteristics that indicate progression. Meanwhile, the study by \cite{hu2014prediction} employed a Dynamic Structure-Function (DSF) model, which demonstrated an improved performance over the conventional OLS regression, especially in shorter follow-ups. Drawing insights from glaucoma suspects, a study by \cite{meira2013predicting} showed the advantages of the Joint Longitudinal Survival Model (JLSM) with RGC Estimation, which outperformed estimation with structure or function measurements alone. \cite{medeiros2014evaluation}, analyzing 492 eyes, reiterated this by linking progressive RA loss conclusively to VF loss, combining JLSM with RA and VF to predict progression. Further, \cite{zhalechian2022augmenting} in their study with a Kalman Filter Estimator for estimating future MD with RNFL measures indicated that global RNFL just marginally enhanced MD predictions as compared to utilizing MD by itself, showing confounding results.

These statistical methods illustrate that a holistic, combined structural and functional analysis is paramount for accurate detection of glaucoma progression. Time-series forecasting methods, such as the DSF model, showed the importance of shorter follow-up for detection and were helpful in frequent monitoring. However, when compared with the probabilistic Bayesian approaches, these methods, while promising, do show that the Bayesian techniques often have better sensitivity. Thus, it is essential to choose an appropriate model in clinical settings to detect progression based on the type of data and how long the patient has been monitored.

\subsection{Artificial Intelligence Utilising Combined Structure and Function Relationship to Evaluate Glaucoma Progression}

Machine learning techniques, especially when they combine both structural and functional data, have shown better results in identifying glaucoma progression. In classification methods, \cite{nouri2021prediction} research used an Elastic net logistic regression and ML classifier to achieve AUCs between 0.79 and 0.81, suggesting an improvement of predictive power using baseline and longitudinal structural data on visual field (VF) progression. This was reflected by multiple studies notably \cite{bowd2012predicting} using Relevance Vector Machine, \cite{yousefi2013glaucoma} implementing a multitude of classifiers (including Bayesian Net and Lazy K Star), and \cite{lee2020machine} employing Random Forest and Extra Trees, where all emphasized the significant role of RNFL measurements and baseline parameters for detecting progression. Notably, the study by \cite{kamalipour2023combining} showed an impressive AUC of 0.89 by integrating OCT and OCTA features via a Gradient Boosting Classifier, demonstrating its efficacy in forecasting clinical VF progression.

On the other hand, regression methods by \cite{lee2022predictive} using Random Forest to measure the rate of change in RNFL thickness against baseline emphasized how such models improve the predictive accuracy of baseline ONH characteristics. While the outcomes are varied, Random Forest, Gradient Boosting, and Elastic Net methods tend to produce superior performance, often surpassing traditional regression and decision trees. It is crucial to note that these studies used diverse data types, from cpRNFL and GCIPL thickness to OCT scans and VF parameters, allowing for nuanced analysis. While these methods were shown to be great at identifying complex patterns in the data, there's still a worry that they might overfit smaller datasets. Therefore, while ML methods provide a holistic view, a balance between model complexity and data characteristics is essential for optimal results.

Deep Learning techniques, being able to learn from complex data, provide advanced modeling capacity, especially when leveraging both structure and function. A study by \cite{sedai2020forecasting} used a 3DCNN with traditional ML forecasting techniques, primarily utilizing cpRNFL from OCT, age, and 24-2 VF tests as input, which obtained the lowest MAE across healthy, suspect, and glaucoma subjects, showcasing the method's reliability in real-world applications. \cite{lee2021predicting} in their research introduced an innovative Machine to Machine (M2M) method paired with JLSM, which used Color Fundus and SDOCT RNFL thickness data to detect longitudinal changes, showed this method can distinguish between converter and non-converter groups in glaucoma. In a notable multimodal approach, \cite{pham2023multimodal} combined ResNET50 and LSTM to forecast future VF points using the 30-2 VF HFA SITA and RNFL thickness map to demonstrate its predictive accuracy in noisy data environments. Among these, the incorporation of time-series data, as seen in LSTMs, highlights a paradigm shift towards capturing sequential data and temporal dependencies, improving prediction accuracy. However, while DL methods have shown superior predictive capabilities, the need for larger datasets for generalization and computational power for training is a significant drawback. Moreover, DL, being a black-box model, cannot provide interpretable results, especially in clinical settings where understanding model decisions is crucial.

Elaborating on the DL methods, techniques like 3DCNNs, M2M methods, and complex architectures like ResNET50 with LSTM have been adopted efficiently to detect glaucoma progression. These models utilize large datasets to extract intricate patterns from both structural and functional data to produce improved performance. In comparison, traditional ML models, like Random Forest and Gradient Boosting, although sophisticated, often require a certain degree of manual feature extraction and may not capture intricacies as effectively as DL models. When compared with probabilistic and purely statistical models, such as PLR methods or linear regressions, both machine and deep learning offer superior predictive capabilities than the former. However, probabilistic and statistical methods provide interpretable results and a clear insight into data dynamics, often making them preferred choices for direct interpretability and better understanding of the data.

\begin{landscape}
    \small
    \sffamily
    \setlength\tabcolsep{2pt}    
\begin{tabularx}{\linewidth}{@{} *{11}{R} >{\linespread{0.84}\selectfont}R @{}}
    
    \caption{A Review of Methods for Detection and Prediction Glaucoma Progression using both Structure and Function}
    \label{tab:dlstructfunc} \\
    
    \toprule

    \thead[bl]{Citation}   
    & \thead[bl]{Setup} 
    & \thead[bl]{Algorithm/\\Method} 
    & \thead[bl]{Dataset} 
    & \thead[bl]{Data\\Type}   
    & \thead[bl]{Follow-up\\Period}     
    & \thead[bl]{Reference\\Standard}  
    & \thead[bl]{Baseline\\Method} 
    & \thead[bl]{Model\\Output} 
    & \thead[bl]{Outcomes} 
    & \thead[bl]{Summary}   \\
    
    \midrule

    \cite{bilonick2008evaluating}   
    & Regression     
    & Latent Class Regression                                                    
    & 106 eyes                                
    & MD, PSD, AGIS score, VFI, RNFL                                                                
    & 5 tests, 5 years                                        
    & significant negative slope (p\textless{}0.05)                                 
    & -                                                            
    & LCR Model slope, AIC                                                    
    & 2494.8 AIC                                                                            
    & Baseline RNFL was indicative of progression \\

    \cite{medeiros2011combining}    
    & Regression     
    & Bayesian Hierarchical Model                                                
    & 434 eyes - glaucoma and suspect             
    & VFI from SAP, GDx TSNIT average from Optic Disc Stereophotographs, Scanning Laser Polarimerty 
    & 3 reliable tests, 4.2 years                             
    & OLS regression slope of VFI, observer disagreement of stereophotograph change 
    & OLS regression slopes on VFI and TSNIT with p \textless 0.05 
    & Bayesian slopes of change for VFI and TSNIT                             
    & 22.7\% vs 12.8\% using VFI, 74\% vs 37\% with optic disc at 100\% specificity         
    & Bayesian method obtains significantly higher sensitivity for progression with VFI and TSNIT 100\% specificity \\

    \cite{hu2014prediction}         
    & Regression     
    & Dynamic Structure-Function Model                                           
    & 220 eyes from DIGS and ADAGES           
    & MS from 24-2 SAP Tests and Rim Area from scanning laser ophthalmoscopy                        
    & 3 follow-ups, 8.4 years                                 
    & Glaucoma Criteria based on ADAGES                                             
    & OLS regression slopes                                        
    & Prediction in future visits and MSE                                      
    & Lower MSE 70\% eyes at visit 4 and 60\% for 5,6,7                                     
    & DSF outperforms OLS in shorter intervals  \\

    \cite{medeiros2012combining}    
    & Regression     
    & Bayesian Joint Regression Model                                            
    & 242 eyes - glaucoma                     
    & MD from SAP Test,  RA from CSLO                                                               
    & 4 follow-ups, 6.4 years                                 
    & PSD \textless 0.05 in SAP tests wrt Baseline and reproducible at least once   
    & OLS linear regression slopes with MD and RA                  
    & Bayesian regression slopes of change with MD and RA                     
    & 5.13 vs 11.2 MSE predicting MD and 0.016 vs 0.027 MSE predicting RA                   
    & Bayesian method outperforms OLS \\

    \cite{meira2013predicting}      
    & Regression     
    & Joint Longitudinal Survival Model (JLSM) with RGC Estimation               
    & 288 glaucoma suspect eyes               
    & Avg RNFL from OCT and MD from SAP                                                             
    & 4 follow-up, 3.8 years                                  
    & Significant Slopes from JLSM                                                  
    & Isolated Structure or Function Measurements                  
    & Combined Structue Function Index, RGC Estimates                         
    & -18,987 cells/y progressors, -8,808 cells/y non progressors                           
    & Joint Longitudinal Esitmates better than Structure or Function alone \\

    \cite{russell2012improved}      
    & Regression     
    & Bayesian Linear Regression                                                 
    & 179 eyes                                
    & MS from 24-2 VF tests, RA from CLST HRT                                                       
    & 8 follow-ups, 5.8 years                                 
    & Significant negative slope (p\textless{}0.05)                                 
    & OLSLR                                                        
    & Rate of change in MS                                                    
    & RMSE 0.14dB smaller than OLSLR                                                       
    & BLR with MS and RA outperforms OLSLR for short time series, OLSLR more accurate for long time series with only VF \\

    \cite{sedai2020forecasting}     
    & Regression     
    & 3D CNN + ML Forcasting                                                     
    & Train: 859 subjects, Test: 230 subjects 
    & cpRNFL from OCT, age, IOP, 24-2 VF tests  
    & 3 visits, 3.65 years                                    
    & Glaucoma (2 consecutive test ONL) and abnormal ONH                            
    & Linear Trend Based Estimation                                
    & RNFL Global and Sectoral Means                                          
    & MAE: 1.10, 1.79, 1.87 for healthy, suspect and glaucoma                               
    & Model consistent across suspect and glaucoma subjects \\

    \cite{medeiros2014evaluation}   
    & Regression     
    & JLSM with RA and VF                                                        
    & 492 eyes suspect                        
    & RA from CSLO, SAP 24-2 VF, IOP                                                                
    & 5 CSLO tests, 2 years                                   
    & 3 consecutive abnormal VF tests with PSD (p\textless{}0.05)                   
    & -                                                            
    & RA loss rate, Survival adapted R2, proportion of treatment effect (PTE) 
    & R2 62\% univariate model, 81\% multivariate, PTE 65\%                                 
    & Progressive RA loss predictive of VF loss \\

    \cite{lee2021predicting}        
    & Regression     
    & Machine to Machine (M2M) with JLSM                                         
    & 1072 eyes                               
    & Color Fundus, SDOCT RNFL thickness, 24-2 SAP tests                                            
    & 4.2 fundus tests, 9.6 VF tests, 5.9 years               
    & 2 consecutive abnormal VF (PSD with p\textless{}0.05)                         
    & -                                                            
    & progression: significant slopes from M2M predictions                    
    & -1.02 um/y vs -0.67um/y between converter and non-converter                           
    & Longitudinal changes from DL can predict progression \\

    \cite{pham2023multimodal}       
    & Regression     
    & ResNET50 and LSTM                                                          
    & Train: 266 eyes, Test: 99 eyes          
    & 30-2 VF HFA SITA tests, RNFL map from cirrus OCT tests                                        
    & Train: 5.7 visits, 5 years; Test: 2.3 visits, 3.3 years 
    & -                                                                             
    & -                                                            
    & Future VF points                                                        
    & MAE 3.31, RMSE 4.58                                                                   
    & VF and OCT data improves predictive performance, model is useful with noisy data \\

    \cite{zhalechian2022augmenting} 
    & Regression     
    & Kalman Filter Estimator                                                   
    & 362 subjects                            
    & demographics, IOP, VF MD, VF PSD, global RNFLT                                                
    & 10.6, 19.9, 11.7 follow-ups, 13.6, 12.1, 5.7 years      
    & -                                                                             
    & OLS Linear Regression                                        
    & future MD, PSD and RNFL values                                          
    & Predictive accuracy: 73.5\% vs 58\% with LR                                           
    & global RNFL minimally improved MD prediction than with MD alone \\

    \cite{lee2022predictive}        
    & Regression     
    & Random Forest                                                              
    & 712 eyes with OAG                       
    & demographics, IOP, LCCI, peripapillary CT, global RNFLT, VF MD, PSD, AXL, CCT                 
    & 5.3-11.5 years                                          
    & -                                                                             
    & Regression and Decision Trees                                
    & Rate of change in RNFL Thickness wrt Baseline                           
    & MAE: 0.075, 0.115, 0.128 for RF, LR, DT                                               
    & baseline ONH characteristics predict risk of faster progression \\

    \cite{yousefi2013glaucoma}      
    & Classification 
    & ML Classifiers: Bayesian Net, Lazy K Star, Meta Classifiers, AD Tree, CART 
    & 107 eyes - progressing, 72 - stable     
    & 52 + MD + PSD VF SAP Points, Global + Sectoral OCT RNFL Thickness                             
    & 4.3 follow-ups, 2.2 years                               
    & PGON criteria + GPA                                                           
    & -                                                            
    & progressed vs non-progressed                                            
    & 0.88 AUC for Random Forest with RNFL and SAP; 0.88 AUC for Lazy K Star with only RNFL 
    & RNFL measurements provide more discriminating power \\

    \cite{lee2020machine}           
    & Classification 
    & Random Forest and Extra Trees                                              
    & Train: 110 eyes, Test: 45 eyes               
    & IOP, CCT, 30-2 SAP VF tests, mGCIL thickness, cpRNFL thickness HDOCT                          
    & 6.2 follow-ups, 3.39 years                              
    & event based GPA                                                               
    & Linear Regression Slopes                                     
    & progressing vs non-progressing                                          
    & 0.881 AUC for Extra Trees, 0.811 AUC for Random Forest                                
    & ML Classifiers can predict progression effectively in young myopic patients \\

    \cite{nouri2021prediction}      
    & Classification 
    & Elastic Net Logistic Regression, MLC                                       
    & 104 eyes                                
    & cpRNFL, GCIPL thickness from Macular SD-OCT                                                   
    & 5 tests, 3years                                         
    & PLR deterioration on 24-2 SAP tests \textless -1dB/yr, p\textless{}0.01       
    & PLR Method                                                   
    & progressing vs non-progressing                                          
    & AUC 0.79 with ENR, 0.81 with ML                                                       
    & VF progression can be predicted from baseline and longitudinal structural data \\

    \cite{bowd2012predicting}       
    & Classification 
    & Relevance Vector Machine (RVM)                                                  
    & 264 eyes (10 fold CV)                   
    & 117 CSLO points from HRT II and 52 SAP points from 24-1 SITA HFA II                           
    & 5.35 years                                              
    & SAP GPA or stereo-photo assessment                                            
    & Glaucoma Probability Score                                   
    & progressing vs non-progressing                                          
    & AUC: 0.640, 0.762, 0.805 using CSLO, SAP, combined parameters                         
    & RVM with baseline parameters predicts future progression \\

    \cite{kamalipour2023combining}  
    & Classification 
    & Gradient Boosting (GB) Classifier                                               
    & 110 eyes                                
    & OCT scans, OCTA scans                                                                         
    & 3 follow-ups, 4.1 years                                 
    & SAP GPA, significant negative slope of VF MD, PPLR event                      
    & -                                                            
    & progressing vs non-progressing                                          
    & AUC 0.89 with both OCT and OCTA features                                              
    & ML with OCT and OCTA predicts clinical VF progression \\ 
    
    \bottomrule
    
    \end{tabularx}
\end{landscape}

\section{Glaucoma Progression Detection with EHR and Clinical Data}

Deep learning has also been successfully applied to detect glaucoma progression using structured clinical data and text notes from Electronic Health Records (EHR). These techniques offer a paradigm shift from traditional uses of deep learning methods using structure or functional progression characteristics and have been shown to predict progression accurately. Research by \cite{wang2023machine} and \cite{baxter2019machine} explores several machine learning classifiers using EHR data to predict if the data is indicative of progressive characteristics and if the patient needs surgery. The models obtained a moderate AUC of 0.623-0.673 on the test sets, suggesting further research is required during inference, especially for different demographics. Another research \cite{tao2023predicting} by the same group studied survival-based AI algorithms to predict if a patient is showing characteristics of glaucomatous progression to surgery. They showed that using more complex algorithms such as DeepSurv (DL method) has a better predictive AUC of 0.802 due to its ability to capture information from high-dimensional data. The addition of clinical text notes with EHR data has also boosted the performance of deep learning models in classifying whether patients undergoing surgeries show signs of progression, obtaining an AUC of 0.873 (\cite{jalamangala2023predicting}). Exploring the utility of DL methods to predict progression from unstructured data, a Natural Language Processing (NLP) based DL algorithm was developed to predict progression from free-text clinical notes (\cite{wang2022deep}). Although a combination of free text and EHR data obtained a higher AUC of 0.73 in the NLP model, only free-text data also had an appreciative AUC of 0.70. Making the DL model more complex has been shown to enhance its accuracy. For example, research by \cite{hu2022predicting} demonstrated this by using BERT-based models on clinical notes obtained from ophthalmologists. This approach predicted which patients might need glaucoma surgery with a reliability score, AUC of 0.734. Although the performance was not better than DL methods with just structured EHR data, this research showed the potential of using massive pre-trained models to predict progression using free text notes, which are abundantly available.

\section{Conclusion}

In this chapter, we have seen various models and algorithms using different modalities or combinations of modalities to detect, predict, and forecast glaucoma progression from longitudinal data. We observed various advantages of using complex techniques, trade-offs between structure and function, and challenges in the modeling of medical data. Identifying a holistic method for predicting glaucoma progression with a reliable reference standard is found to be quintessential. In addition, trade-offs between structural and functional assessment for progression suggest the need for a precise, reproducible, and comprehensive feature set that can accurately represent glaucomatous characteristics with the ability to detect and separate age-related variability. Obtaining longitudinal image data can be time-consuming and expensive, which adds another layer of constraints to the model development. However, with the advent of complex models such as deep learning, the availability of computational resources and data from large cohorts can be used to make powerful models that can predict glaucoma progression accurately and with minimal clinical expertise. In subsequent chapters, we introduce and formulate novel deep-learning strategies for detecting glaucoma progression within a longitudinal cohort characterized by data ambiguity, leveraging anatomical information of progression characteristics.
\chapter{Methodology}
\label{chap:example}

\section{Dataset Overview}

The study, acquisition and documentation of the dataset in this chapter was carried in collaboration with Alessandro A. Jammal, MD, PhD and Felipe A. Medeiros, MD, PhD. 

\subsection{Duke Ophthalmic Registry}

The data set used in this study is obtained from the Duke Ophthalmic Registry (DOR), currently the largest single Institution clinical database for ophthalmic records in the world. The database contains several millions of clinical and imaging data for patients with eye diseases. Taking over three decades of routine follow-up in a multi-ethnic and culturally diverse group of people in central-eastern North Carolina, the database has been used for multiple AI studies by researchers at the Duke Eye Center at Duke University. The DOR database is an extension of the Duke Glaucoma Registry (DGR) (\cite{jammal2021rates}) by the Vision, Imaging, and Performance (VIP) laboratory at the Duke Eye Center whose main aim was to aggregate a large population and create an accessible pool of 'real-world' clinical information database of glaucoma. The DOR contains patient eye disease information collected at the main eye center and six satellite eye clinics of the Duke University Health Clinics (DUHS), boasting over 485,339 patient data undergoing routine ophthalmic care. Advanced imaging data stored in the DOR is one of the largest sources of longitudinal studies in glaucoma progression and has been the foundation of many researches involving AI and DL applications in evaluating glaucoma progression.

The DOR is an amalgamation of comprehensive health information of the patients undergoing ophthalmic care. It consists of demographic data (age, sex, location, etc.), medical history, clinical diagnosis and encounters, lab test results, and complete ophthalmic examination acquired over several visits of Medical Eye Care of each patient. The ophthalmic examination contains critical clinical and imaging data used in ophthalmic care, such as IOP, visual acuity (VA), fundus photographs, SDOCT scans, and SAP tests. Data hierarchies managed by the DOR were extracted using the Duke Enterprise Data Unified Content Explorer (DEDUCE). This web-based query tool efficiently searches patient Care information compiled using EHR. A protected Analytics Computing Environment (PACE), a virtual network space for data analysis, adds another layer of protection by de-identifying or masking patient data during research, in adherence with the standard of ethics in academic research. A de-identified population characteristics data from the DOR is provided in the following section.

\subsection{Population Characteristics}

\begin{figure}[!ht]
\centering
	\includegraphics[width=0.75\textwidth]{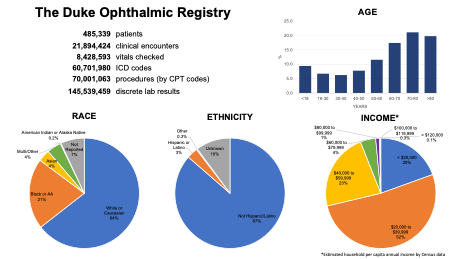}
\caption{Demographic Characteristics of patients in the Duke Ophthalmic Registry.}
\label{fig:demopop}
\end{figure}

The DOR is a growing retrospective database of all available electronic health records from patients' visits at the Duke Eye Center and its satellite clinics undergoing medical care. A wide array of clinical data from diverse populations was collected using relaxed inclusion criteria approved by the Duke Institutional Review Board (IRB). The Dior represents an unbiased sample of the population with demographic characteristics similar to the population of North Carolina \cite{uscensus}. This can be seen by comparing the population from DOR and the US Census Bureau, which shows racial and ethnic similarity. For example, 20.7\% of the population self-reported as Black or African American in DOR vs. 22.2\% in US Census, 64.3\% were White or Caucasian vs. 69.9\% in US Census and 3.5\% Asian vs. 3.6\% in US Census, with a majority of the population being non-Hispanic or Latino (Figure \ref{fig:demopop}). 

\begin{figure}[!ht]
\centering
	\includegraphics[width=0.7\textwidth]{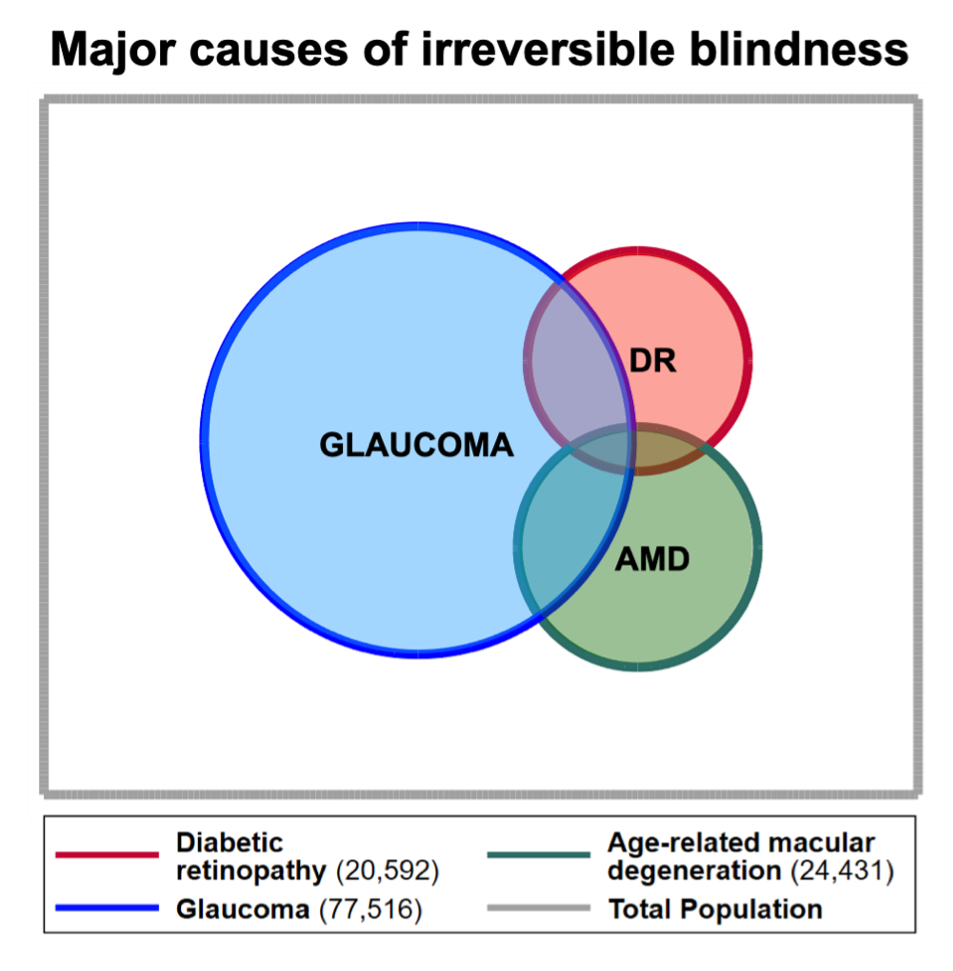}
\caption{Comparison of top three causes of irreversible blindness in the DOR database.}
\label{fig:diseasefreq}
\end{figure}

Aimed to provide better care for improved eye health and vision quality of life, the DOR has a collection of rich samples of common and rare eye diseases. Analyzing the significant health aspects, 77,516 subjects in the database have or are suspected of glaucoma diagnosis, 24,431 subjects have age-related macular degeneration (AMD), and 20,692 subjects have diabetic retinopathy (DR) - the top three causes of irreversible blindness (Figure \ref{fig:diseasefreq}). The database also includes information on more than 111,170 cataract surgeries, one of the most common surgical procedures worldwide, and a wide array of rare eye diseases affecting select vulnerable populations and diseases with a risk of blindness and eye disease progression. This makes DOR one of the most unique and critical sources of ophthalmic data for clinicians, especially at Duke Eye Center, for developing innovative data-centric solutions for eye care. In this research, we will use data from the DOR database to study and establish DL algorithms for glaucoma progression with implications for improving the vision quality of life.

\section{Experiment Design and Setup}

Building on previous research, we describe our study design as a longitudinal retrospective cohort study. Specifically, we develop innovative DL models to discern progression patterns in glaucomatous eyes. The data for this study is derived from the DOR database, organized as longitudinal sequential data for each eye across all the patients involved. Structural assessment is used to define the input features for the DL models. This is because advanced imaging techniques in structural assessment have been shown to effectively capture important anatomical features of the eye, such as RNFL and ONH, which are indicative of progressive changes. This approach is favored as it offers a more precise and reproducible dataset for DL, thereby minimizing the possibility of errors. Since the ultimate objective of this research is to improve the visual quality of life, we use functional outcomes as the benchmark to assess the performance of the solutions derived from our study.

\subsection{Input Features for the Model: Longitudinal SDOCT Scans}

\begin{figure}[!ht]
\centering
	\includegraphics[width=0.7\textwidth]{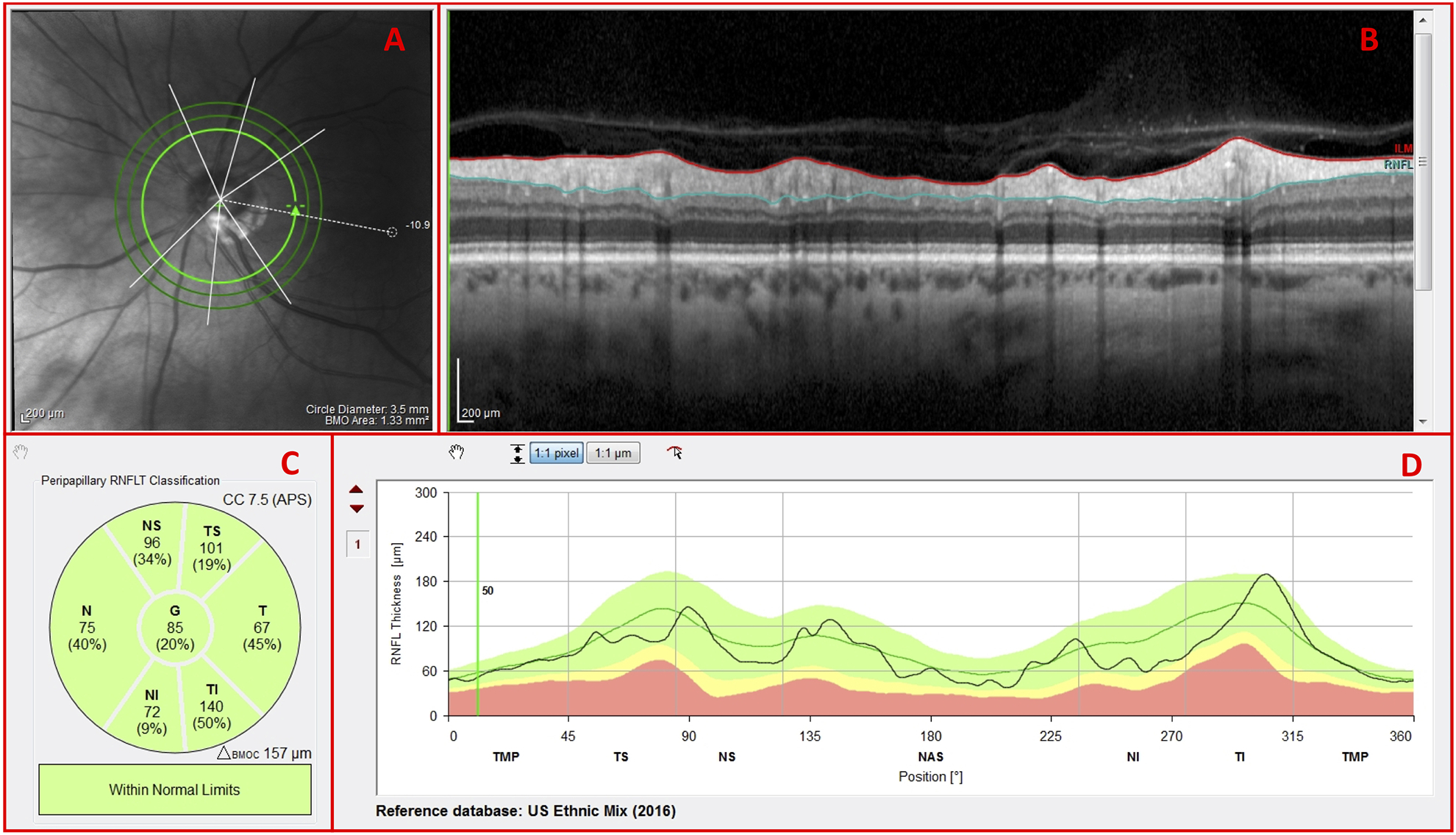}
\caption[A comprehensive SDOCT report of a normal eye obtained from Spectralis (\cite{zemborain2020optical}).]{An SDOCT report of a normal eye showing (A) infrared projection of disc, (B) circular B-scan, (C) cpRNFL pie chart and (D) cpRNFL thickness profile obtained from Spectralis (Heidelberg Engineering, Heidelberg, Germany) (\cite{zemborain2020optical}).}
\label{fig: spectral}
\end{figure}

Longitudinal scans of the retina, specifically around the ONH, obtained from the Spectralis SDOCT (Heidelberg Engineering, Heidelberg, Germany) as part of standard clinical care are used as the primary input for the study. The Spectralis OCT is an advanced imaging system that combines CSLO from Heidelberg Retina Angiography with the dual beam SDOCT to obtain micrometer accurate representation of RNFL, GCIPL, Bruch's membrane (BM), and other layers that form the RGC in the ONH (\cite{leite2011comparison}). Spectralis OCT uses a real-time eye-tracking mechanism to adjust for eye movements and Ensure consistent retina scanning during the visit. It generates a set of different scans with a peripapillary circular scanning pattern of diameter 3.5mm around the ONH, a gold standard scan pattern for detecting structural glaucomatous damage in the RNFL (\cite{chen2009spectral}). These scans include Amplitude-scans (A-scans) - one-dimensional, depth-resolved reflection profiles of the tissue across the ONH, B-scans - two-dimensional cross-sectional images of the tissue by combining multiple A-scans, 3D Volume scans - combining consecutive B-scans, infrared (IR) photograph of the optic disc, cpRNFL thickness pie chart and cpRNFL thickness profile (Figure \ref{fig: spectral}) along with various secondary data (\cite{zemborain2020optical}).

\begin{figure}[!ht]
\centering
	\includegraphics[width=0.85\textwidth]{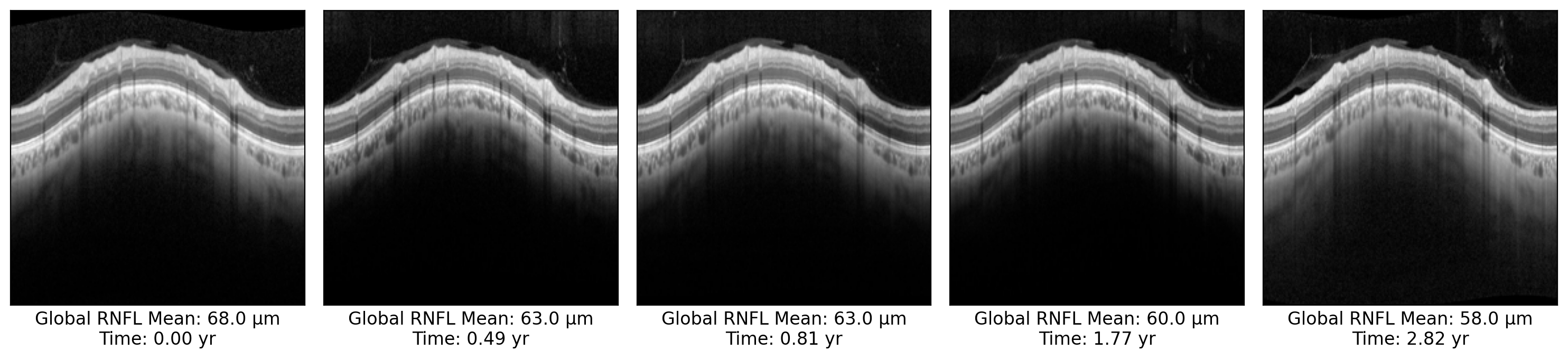}
\caption{A longitudinal sequence of SDOCT B-scan images used as DL model input (resized to $224 \times 224$ pixels).}
\label{fig: lsdoct}
\end{figure}

For relevance, 2D SDOCT B-scan images were used as the input features for the model due to their high precision and reproducibility to capture the complete RNFL profile characteristics in a micrometer scale (\cite{sampani2020comparison}). The study included B-scan images for each patient's eyes with scan rates of 768 and 1536 A-scan points, and all resized to 768 x 496 points using bilinear interpolation. The global RNFL mean thickness for each scan was also recorded from the OCT report to develop baseline models. Scans with segmentation or artifact errors were discarded. Scan quality scores less than 15 were also excluded according to the manufacturer's recommendations. This process was repeated for all eyes across patient visits, obtaining a sequence of multiple OCT scans spaced over a follow-up period (Figure \ref{fig: lsdoct}).

\subsection{Reference Standard: Guided Progression Analysis}
\label{ss:refst}

Glaucoma progression assessment can be a complex and nuanced process whose understanding is often challenging to non-experts. The intricate nature of the disease and its subtle manifestations make it essential to create an objective and simplified method to communicate its progression effectively. A clear and analytical reference is not just beneficial for clinicians but also helpful in explaining patients and stakeholders who are unfamiliar with glaucoma. This underscores a need for an analytical approach to detect progression by tracking deterioration in vision quality of life. Therefore we use Visual Field Guided Progression Analysis (GPA) on SAP tests from the Humphrey Field Analyzer (Carl Zeiss Meditec, Inc., Dublin, CA) as a reference standard due to its ability to detect or predict glaucoma progression in a structured way and overcomes uncertainty.

GPA is a point-wise event-based analysis in which every point in the new VF test is compared with the values from two baseline tests. Points on the VF tests are flagged with (statistically) significant loss of sensitivity ($p<0.05$) or "events" when the measured point-wise pattern deviation becomes greater than a predefined expected variability (derived from repeated tests from a population of stable glaucoma patients). The GPA algorithm then marks the points based on the number of times the "events" repeat at the same location in consecutive tests as:
\begin{itemize} 
    \item \textbf{Empty Triangles}: Locations with a significant change ($p<0.05$) from baseline observed once. 
    \item \textbf{Half Triangles}: VF loss change at that point is confirmed with a second test. 
    \item \textbf{Solid Triangles}: Significant change at the same point is reconfirmed with a third VF test.
\end{itemize} 

\begin{figure}[!ht]
\centering
	\includegraphics[width=0.7\textwidth]{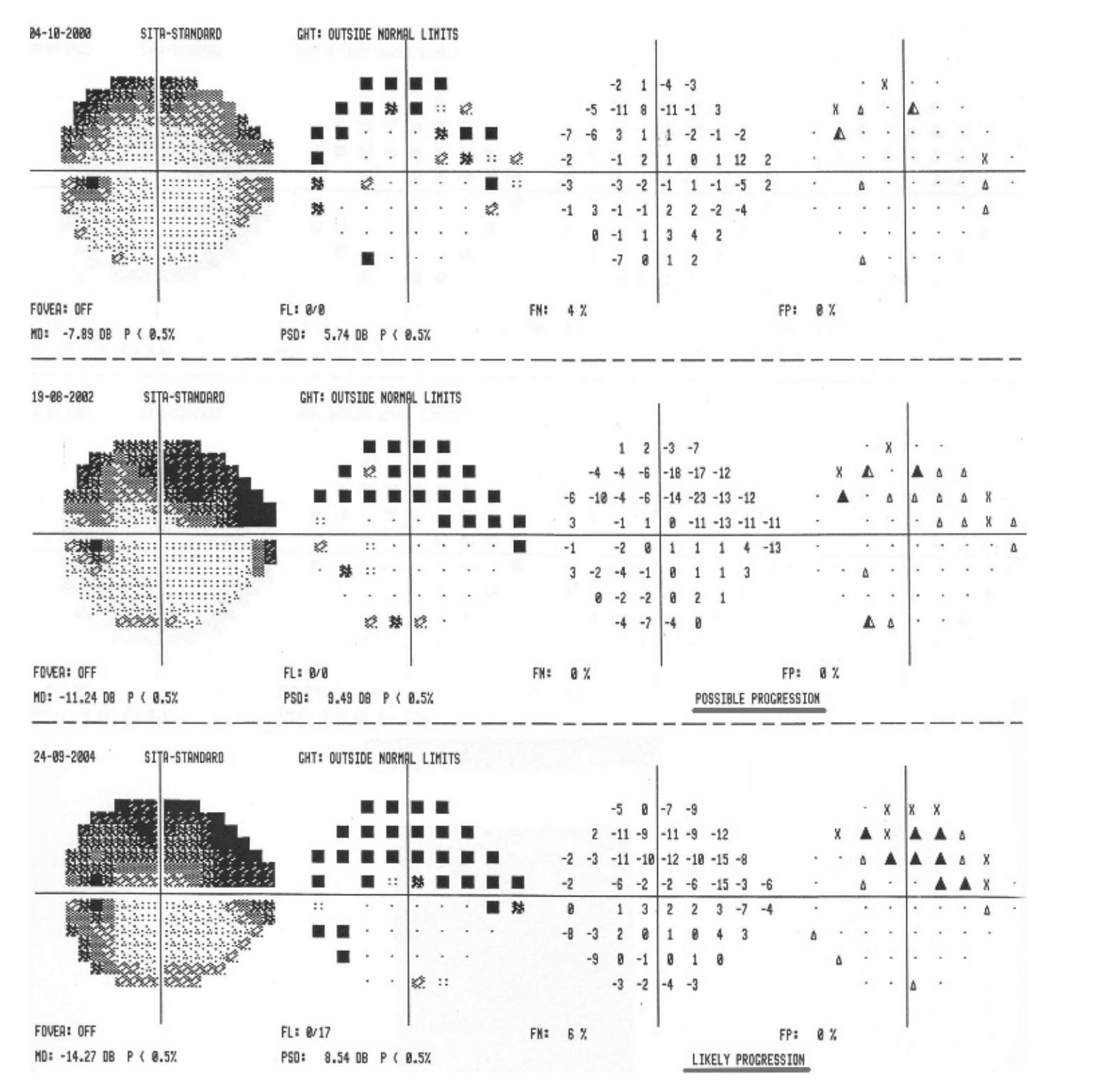}
\caption[A GPA report representing different progression events (\cite{diaz2009detection}).]{A GPA report representing different progression events: no-progression, possible progression and likely progression (top to bottom) (\cite{diaz2009detection}).}
\label{fig: gpatest}
\end{figure}

All the flagged points are locations where the GPA algorithm identifies a potential disease progression. If GPA observed changes at three or more points (at least two solid triangles) in two consecutive follow-up tests, the eye is labeled as "\textit{possible progression},." If changes in these points (at least three solid triangles) are repeated in three consecutive tests, the eye is marked as "\textit{likely progression}" (\cite{vianna2015detect}). This makes GPA a reliable qualitative measure that is relatively simple to implement and accounts for variability associated with VF location, threshold sensitivity, and patient age (\cite{hood202224, giraud2010analysis}). Besides events, the GPA can also report statistical parameters (probability plots), which can help understand the significance of the changes (Figure \ref{fig: gpatest}).

GPA's method to directly contrast follow-up results with a stable baseline helps reduce test-retest variability and provides a better estimate of progression. Moreover, a standardized approach such as GPA ensures consistency, making it easier to compare results across different settings. For simplicity, we formulate our problem in the binary classification scheme with labels as "progression and no-progression," even though GPA produces three outcomes: "no progression," "possible progression," and "likely progression."

\subsection{Baseline Comparison: Ordinary Least Squares Regression Method}

Our baseline comparison method, unless otherwise specified, is the Ordinary Least Squares (OLS) Linear Regression model, which provides trend estimates of progression by quantifying its rate of change. OLS is a foundational statistical method used to analyze linear relationships between dependent variables with one or more independent variables. The output of OLS is the best-fitting linear equation line that describes the data after minimizing the sum of squares of the difference between observed and estimated values. In the context of Glaucoma Progression, OLS LR can be applied to fit global, sectoral, or point-wise values from diagnoses (e.g., sequence of global RNFL thickness values in $\mu m$ from SDOCT taken during follow-ups) with time (yr) to obtain the slopes for the rate of change of values (deterioration in global RNFL thickness in $\mu m/yr$). The definitions of eyes detected as progressing with glaucoma vary across research, but obtaining a statistically significant negative slope ($p < 0.05$) is the most widely used reference.

\subsection{Evaluation Metrics}

Binary classification in disease progression prediction refers to the process of categorizing subjects into dichotomous outcomes: "progressing" or the positive class and "non-progressing" or the negative class, as evaluated by models. For classes $C_1$ and $C_2$ with class prior probabilities $P(C_1)$ and $P(C_2)$, the total data distribution and total probability for binary classification can be modeled as 

\begin{subequations}
	\label{tprob}
	\begin{align}
		\label{prob}
		P(\text{data}) = P(C_1) \cdot P(\text{data}|C_1) + P(C_2) \cdot P(\text{data}|C_2) \\
		\label{sum}
		P(C_1)+P(C_2) = 1
	\end{align}
\end{subequations}

Where $P(\text{data}|C_1)$ represents the conditional distribution of data given class $C$. Predictive models for classification try to assign the most probable "class" of data by categorizing it into one of the two classes. Maximum Likelihood Estimation (MLE) is used by complex models to estimate the parameters of the model. Assuming the true data distribution assigns data points to $C_1$ with probability $y$ and to $C_2$ with probability $1-y$ \eqref{sum} and the predictive model assigns data points to $C_1$ with probability $p$ and assigns to $C_2$ with probability $1-p$, we can derive the cross entropy of our classification problem as: 

\begin{subequations}
	\label{eq:cee}
	\begin{align}
		\label{eq:ce1}
		H(y,p) = -\sum\limits_{i=1}^{2}P(C_1)\log P(data|C_1) \\
		\label{eq:ce2}
		H(y,p) = -y\log(p)-(1-y)\log(1-p)
	\end{align}
\end{subequations}

Which gives the binary cross entropy from the total probability perspective. Here, $P(\text{data}|C_1)$ is viewed as observing the expected encoding length using the predicted distribution for events from the true distribution. Without going into much details, we can also derive the binary cross entropy (BCE) loss from MLE as:

\begin{equation}
\label{eq:bce}
\mathbf{J}(\theta) = -\sum\limits_{i=1}^{N} [y_i \log(p_i) + (1-y_i) \log(1-p_i)]
\end{equation} 

minimizing which gets the best estimate for model parameters $\theta$ to predict class $C_1$ and $C_2$.

\begin{figure}[!ht]
\centering
	\includegraphics[width=0.65\textwidth]{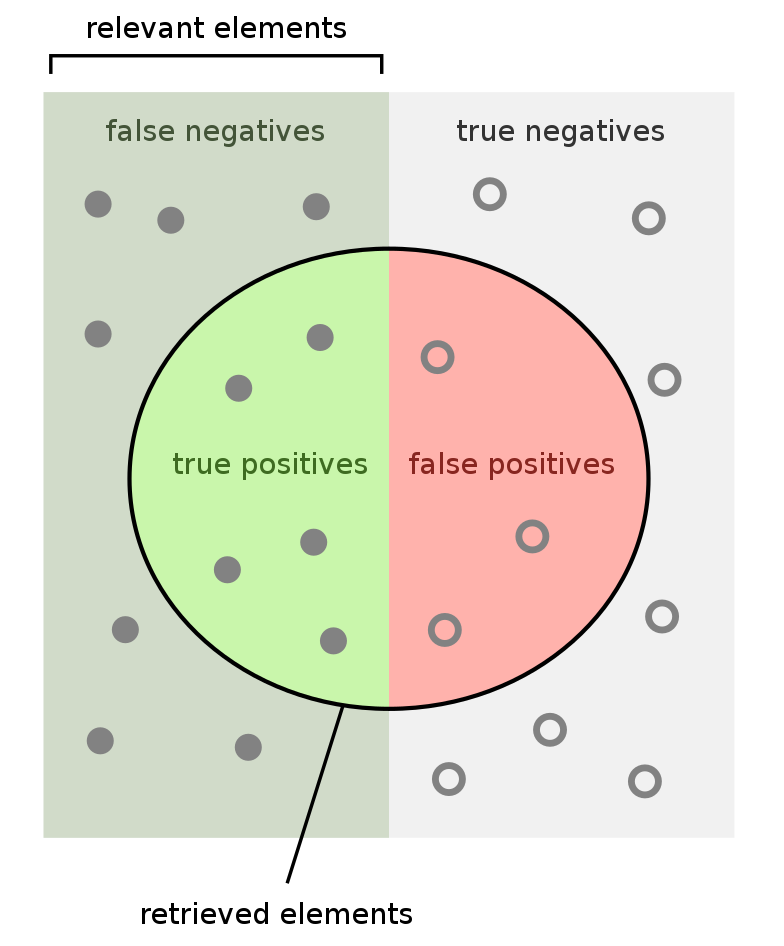}
\caption{Classification Confusion Matrix based on items retrieved from all relevant examples (\cite{prwiki}).}
\label{fig: cfmat}
\end{figure}

Evaluating the model's predictive performance in medical data analyses is not only important for disease progression detection but also critical due to frequent encounters with class imbalance. It is imperative to get the "progressing" samples categorized correctly while ensuring extra care is taken to prevent false alarms (calling progressing samples "non-progressing") (\cite{hicks2022evaluation}). The predictive performance of classification models can be estimated from the confusion matrix by comparing predictions to ground truth. A confusion matrix is given by (Figure \ref{fig: cfmat}): 

\begin{equation}
\text{Confusion Matrix} = 
\begin{pmatrix}
\text{True Positive (TP)} & \text{False Positive (FP)} \\
\text{False Negative (FN)} & \text{True Negative (TN)}
\end{pmatrix}
\end{equation}

Sensitivity, or the true positive rate, is one of the important metrics used to evaluate models for medical analysis. It calculates the rate of classifying positive (progressing) samples correctly: 

\begin{equation}
\text{Sensitivity (or True Positive Rate)} = \frac{\text{TP}}{\text{TP} + \text{FN}}
\end{equation} 

Specificity, an equal if not greater significance than sensitivity, evaluates the model's performance to classify the negative (non-progressing) samples correctly. It is also called the true negative rate and is given by:

\begin{equation}
\text{Specificity (or True Negative Rate)} = \frac{\text{TN}}{\text{TN} + \text{FP}}
\end{equation} 

Additionally, given the emphasis on minimizing false alarms (false positives) in medical settings, there are often benchmarks like "sensitivity at 95\% specificity" to provide a clinical view of model evaluation. Clinicians obtain this value by calculating the sensitivity (hit-ratio) by changing the probability or significance threshold to equalize specificity to 95\%. This allows the model to balance between capturing true cases and avoiding overdiagnosis. Unless otherwise stated, our research will use these values extensively to report model performance.

Extending the above criterion, the Area Under the Curve of the Receiver Operator Characteristics (AUC-ROC) will be used to measure the classification model's discerning capability across different probability thresholds, with higher scores denoting higher classification power and 0.5 equating random guessing. Finally, since the accuracy provides a holistic view of both progression and non-progression samples classified correctly, it will be used as the basic metric for initial evaluations of the model: 

\begin{equation}
\text{Accuracy} = \frac{\text{TP} + \text{TN}}{\text{TP} + \text{TN} + \text{FP} + \text{FN}\text{ (Total Observations)}}
\end{equation}

The ensemble of the metrics discussed in this section will provide enough evidence for a detailed and robust evaluation of the model's clinical utility.

\subsection{Post-Hoc Statistical Analysis}

Post-hoc statistical analyses play a pivotal role in evaluating the clinical relevance of medical image classification systems. Once the predictive model has been developed and validated on a dataset, it is crucial to discern its real-world clinical applicability and acceptability. Linear Mixed Models (LMM), often used in this context, are suitable for analyzing clinical datasets with hierarchical or nested data structures, such as multiple measurements from the same patient or, specifically in ophthalmology, from each eye of the patient. LMM can model the predictions of classification systems with covariates like clinical or demographic features nesting random effects at the patient or eye level. The outputs of the LMM can be used to understand the behavior of the classifier against a covariate across different subgroups. Researchers can ascertain the clinical relevance of the classifier model if the LMM for model predictions obtains a similar performance or significance with true values as determined by the reference standard. This method often offers a nuanced understanding of the classifier's performance and can help identify areas of improvement from a clinical perspective. 

Comparing population characteristics is also an important factor for ascertaining the medical applicability of the model. Chi-squared tests play a crucial role in determining if there are significant differences in populations, especially well-observed demographics like gender and race. This ensures the predictive model isn't inadvertently biased towards any particular demographic or subgroup, owing to generalizability and fairness. Additionally, McNemar's test can be used for a quick comparison of the performance of the developed model against traditional methods, like OLS Linear Regression, highlighting significant improvements from the latter (if it exists). The posthoc statistical analysis thus ensures that medical image classification systems, once developed, not only guarantee robustness beyond predictive accuracy but also have clinical applicability in real-world medical applications.

\section{The Time-Series Deep Learning Model}

Glaucoma progression detection is a longitudinal study of disease progression. In our research, we focus on longitudinal structural assessments (SD-OCT B-scan images) taken over a follow-up period to determine if the OCT exams are indicative of subsequent progressive loss of visual fields by acceptable clinical metrics. For this, we develop time-series deep learning models that use the longitudinal OCT image sequence to learn both the spatial features within each image as well as the temporal evolution across image sequences. Unlike traditional image classification, we define our dataset $\mathcal(D)$ with $N$ points as a time-series model where each datapoint contains $\tau$ B-scans $x^{(t)}$ taken over a time period $T$. We can express $n^{th}$ datapoint in the time-series dataset as ${(x^{(1)}, x^{(2)}, \ldots, x^{(\tau)})}_n \rightarrow y_n$, where $y_n$ is the class label for that point. Outcome $y$ is derived from the reference standard, which is typically obtained at endpoints after the final follow-up time $\mathcal{T}$, where $T \le \mathcal{T}$. Therefore, the time-series model or "oracle" for the dataset can be written as: 

\begin{equation}
f(x^{(1)}, x^{(2)}, \ldots, x^{(\tau)}) \rightarrow y
\end{equation}

Where $x^{(t)}$ represents the spatial features derived from SD-OCT B-scan image at time instance $t$, $\tau$ denotes the total number of instances in the series, and the output $y$ is the classification output of the longitudinal data. The function $f(\cdot)$ is the oracle function which maps the temporal vector $X_n = (x^{(1)}, x^{(2)}, \ldots, x^{(\tau)})_n$ to ground truths $y_n$. Given that we are dealing with binary classification, our labels have $K=2$ classes, which can be represented as a $K$-dimensional vector using $1-\text{of}-K$ encoding.

Our objective is to develop a time-series deep learning model, $\mathcal{H}$, that can encode the spatiotemporal feature vector $X_n$. The goal is to optimize the parameters of $\mathcal{H}$ so that the predicted labels $\hat{y}$ closely match the true probability distributions of $y$. We use MLE to find the optimal parameter values $\theta$ of the DL model $\mathcal{H}$ that maximizes the likelihood of observing data given model. The probability of observing the dataset $\mathcal{D}$ is the likelihood (density) function given as:

\begin{equation}
\mathcal{L}(\theta | \mathcal{D}) = \prod\limits_{i=1}^{N} P(y_i | X_i, \theta)
\end{equation}

Rewriting $P(\cdot)$ with its probability distribution (binomial), replacing probabilities with appropriate estimates $\hat{y}$ and taking the likelihood function's negative log-likelihood, we get a more tractable function:

\begin{equation}
\mathbf{J}(\theta) = -\log \mathcal{L}(\theta | \mathcal{D}) = -\sum\limits_{i=1}^{N} [y_i \log(\hat{y}_i) + (1-y_i) \log(1 - \hat{y}_i)]
\end{equation}

Which is the exact BCE formula in Equation \eqref{eq:bce} derived in the previous section. Minimizing this loss (maximizing likelihood) with respect to model parameters $\theta$ obtains the most optimal parameters.

In the following subsections we discuss the specific parts of our time-series DL model which makes it unique.

\subsection{Pre-trained CNN Networks: 3DCNN + ResNet50}

The first layer of our deep learning architecture is a 3D Convolutional Neural Network (3D-CNN). We use 3D-CNN because of its intrinsic ability to capture not only the spatial features of each image but also the short-term temporal dynamics across the longitudinal image data. This dual encoding is crucial, especially in the first layer, since the input sequence contains high-dimensional images where the feature space is large (\cite{manttari2020interpreting}). By concentrating on contiguous spatial patches in the image sequence, the 3D-CNN extracts the most salient features for both spatial and temporal learning. These features are then passed on to a subsequent pre-trained ResNet50 model. This way, the 3D-CNN ensures that the ResNet50 model receives the most relevant and concise spatial-temporal encodings.

\begin{figure}[!ht]
\centering
	\includegraphics[width=0.9\textwidth]{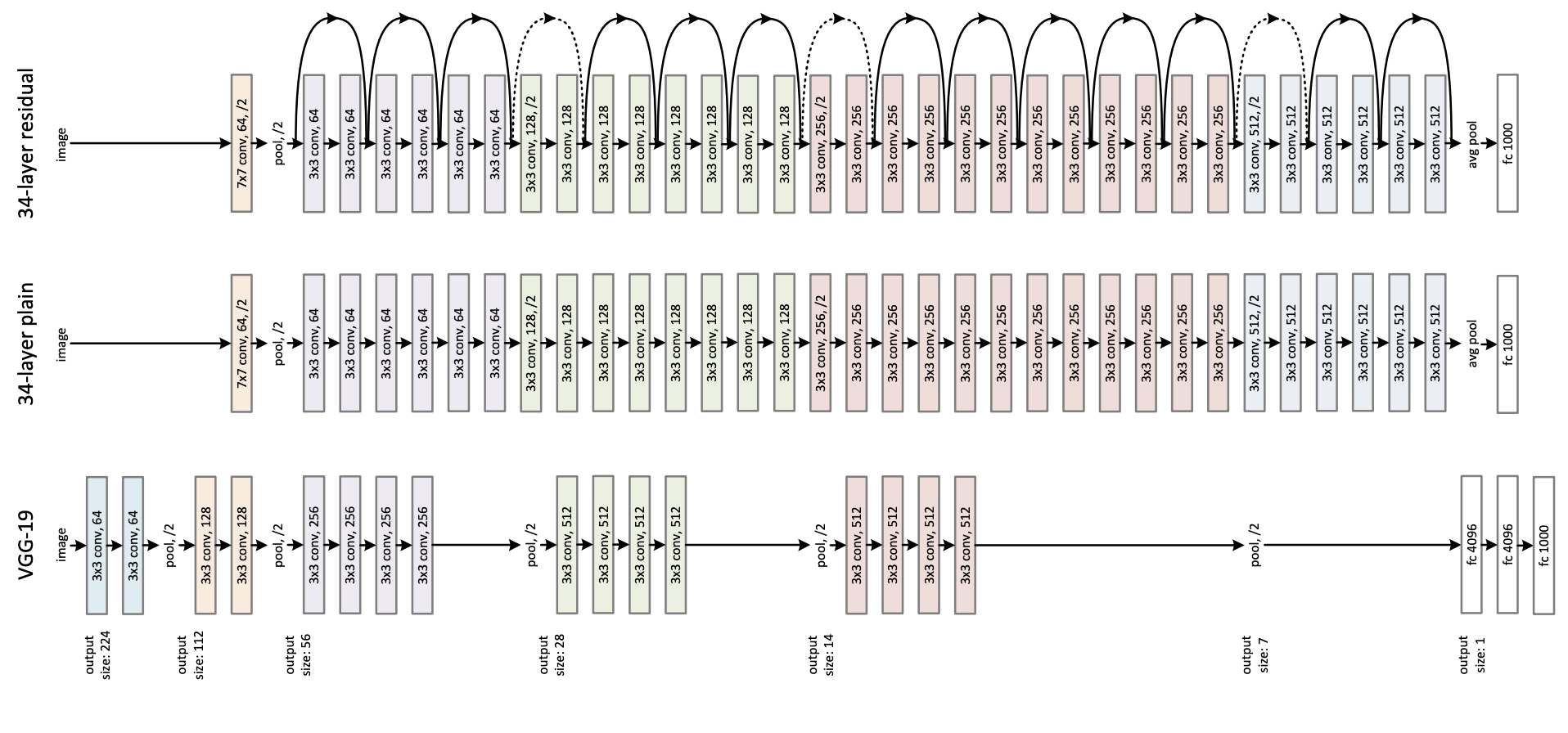}
\caption[Representative example of 34-layer Deep Residual Network architecture (\cite{he2016deep}).]{Representative example of 34-layer Deep Residual Network architecture (top) compared to a plain network (middle) and VGG-19 (bottom) (\cite{he2016deep}).}
\label{fig: rnet34}
\end{figure}

A pre-trained ResNet50 model follows the initial 3DCNN layer to encode the spatial features further. ResNet50, a 50-layer deep residual neural network (DNN), is known for its capability to extract complex spatial patterns and hierarchies from image data due to its ability to solve the vanishing gradient problem in deep networks (\cite{he2016deep}). ResNet50 has been widely used in the development of novel deep learning algorithms and transfer learning tasks in smaller datasets, producing state of the art results. Due to this, Residual Networks have also been widely used in the field of medical image analysis. Thus, we develop our spatial encoder by leveraging the weights and architecture of ResNet50 pre-trained on a large-scale ImageNet1K dataset. This way, our model can use prior knowledge and complex spatial feature representations to enhance the performance of the DL model on the glaucoma progression detection task. A representation of 34-layer ResNet model architecture is shown in Figure \ref{fig: rnet34}.

\subsection{Sequence Learning: LSTM Networks}

\begin{figure}[!ht]
\centering
	\includegraphics[width=0.7\textwidth]{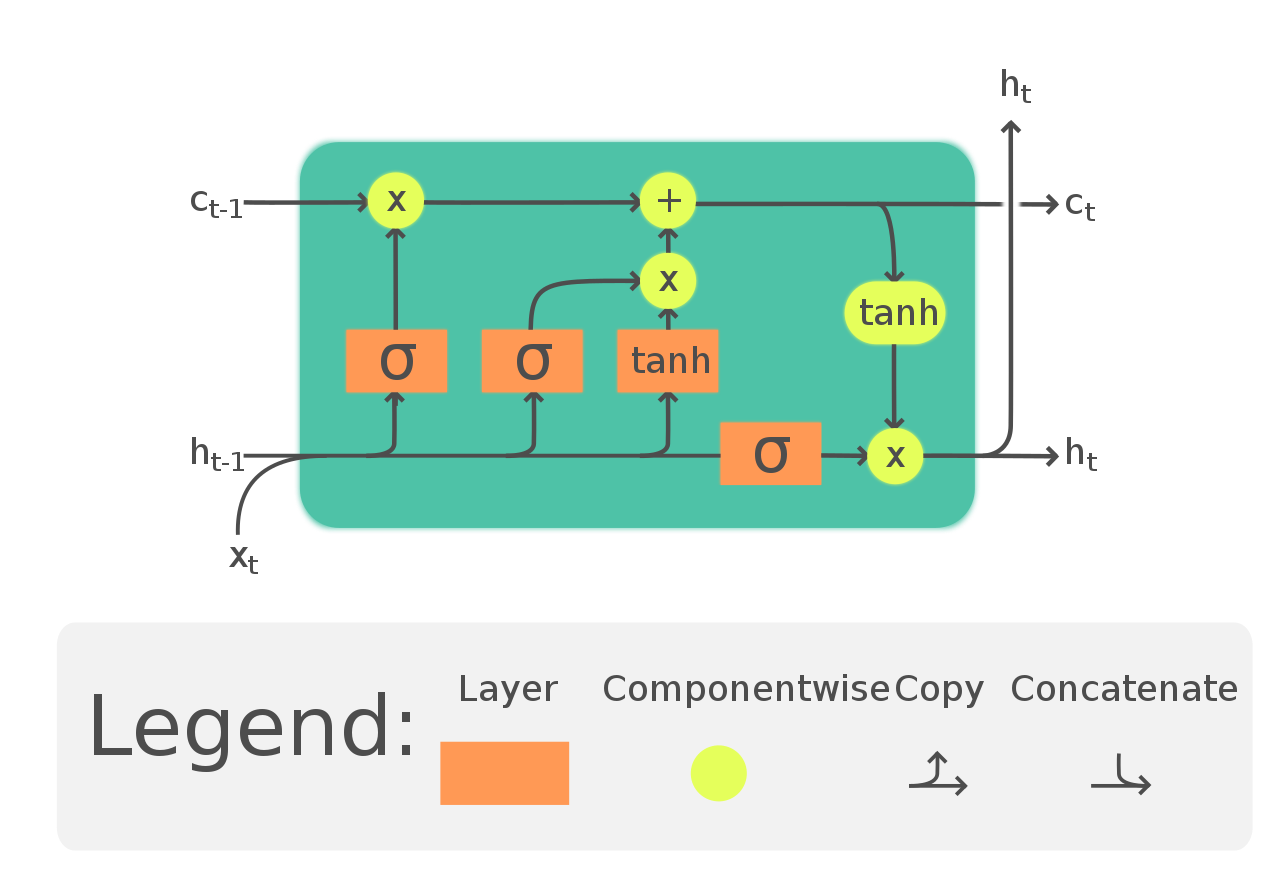}
\caption{A schematic of the Long Short-Term Memory cell explaining components of the Recurrent Neural Networks (\cite{lstmwiki}).}
\label{fig: lstm}
\end{figure}

Long Short-Term Memory (LSTM) networks, an advanced type of recurrent neural network, are then used to capture the temporal evolution of the spatial encodings derived from the ResNet50 model. An LSTM model is made of memory cells controlled by gate mechanisms (\cite{hochreiter1997long}) (Figure \ref{fig: lstm}). Due to this, LSTMs are excellent in encoding the long-term dependencies in the time-series data, making them particularly suitable for handling the longitudinal OCT image sequence. By processing the sequential spatial features derived from the ResNet50, the LSTM layer learns to recognize intricate temporal patterns, capturing both intra-patient variabilities — such as the progression rate within the image sequence of an individual — and inter-patient variabilities — like differences in progression patterns across individuals (\cite{mousavi2019inter}). This nuanced yet thorough understanding of temporal patterns allows for a more comprehensive representation of the glaucoma progression, allowing accurate representation of spatiotemporal features $z$.

\subsection{Spatio-temporal Learning: Combining CNN and LSTM}

Spatio-temporal learning effectively integrates both the CNN and LSTM networks to analyze both spatial and temporal features in longitudinal datasets together. In our model, the CNN module is made of a 3D-CNN layer followed by a pre-trained model. ResNet50 extracts the spatial and short-term temporal patterns from image data, ensuring important features are highlighted. These encodings are then fed into the LSTM network, which captures the long-term dependencies to recognize evolving patterns over time, combining both methods overall. This combined approach offers a unique perspective of the disease, leveraging both CNNs and LSTMs to achieve the most accurate representations of glaucoma progression.

\subsection{Classification Head}

\begin{figure}[!ht]
\centering
	\includegraphics[width=0.85\textwidth]{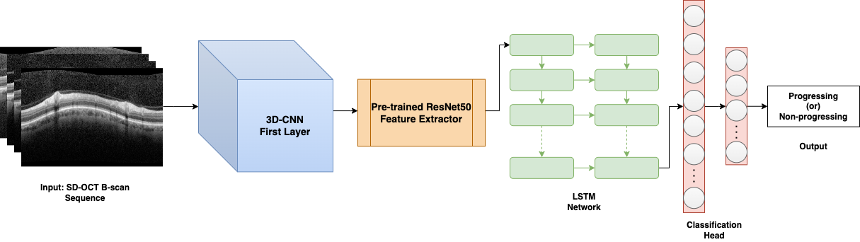}
\caption[An overview of the combined CNN-LSTM Network with a Classification Head.]{An overview of the combined CNN-LSTM Network with a Classification Head: CNN-LSTM encodes spatiotemporal features and Classification Head predicts progression from representations.}
\label{fig: cnnlstm}
\end{figure}

A classification head is used at the end of the learning task to convert the complex feature representations, parsed by the CNN and LSTM layers, into a clinically relevant diagnostic outcome for glaucoma progression. The classification head typically contains fully connected layers to compress high-level encodings, activation functions to introduce non-linearities for complex patterns and relationships, and a final \textit{softmax} function to get logits or probabilities to generate predictions for the binary classification task. Assuming the model contains parameters $\theta$, the label distribution of classes $k$ from the softmax becomes:

\begin{equation}
P_{\theta}( y=k | X_n ) = \frac{e^{\mathcal{H}_{\theta}(X_n)}}{\sum\limits_{k \in K} e^{\mathcal{H}_{\theta}(X_n)}}
\end{equation}

where $\mathcal{H}_{\theta}(X_n)$ is the spatiotemporal encodings obtained from the model till the final layer and $P_{\theta}( y=k | X_n )$ signifies the probability of the model predicting data $X_n$ to $k^{th}$ class. Therefore, the model's output can be interpreted as the likelihood of the model to predict glaucoma progression for an input image sequence. An overview of the combined CNN-LSTM network for spatiotemporal encoding is given in Figure \ref{fig: cnnlstm}

We observe that the longitudinal data for each eye obtained from the DOR database has variable follow-up lengths depending on the patient's follow-up time. We employ various strategies to make the sequence size uniform and empirically select a sequence of 5 OCT images as the input for the model. Model-specific input sequence generation processes are shown in their respective chapters. In the following sections, we will explain in brief two of the most prominent problems in the detection of disease progression and subsequently formulate our research solution.

\section{Modelling with Healthy Patient Data: Modified Positive Unlabeled Learning}
\label{sec: npu}

The accuracy of the DL model for glaucoma progression detection and its clinical relevance highly depends on the reference standard. Since there is no consensus for a universally accepted reference gold standard, we use surrogate methods to evaluate predictions. A change detection model was developed by \cite{belghith2015learning}, which used OCT image pairs to identify regions in the OCT scans that were likely to predict progression. Other research by \cite{leung2013impact} and \cite{jammal2020effect} showed that the cpRNFL layer naturally undergoes age-related deterioration, which can increase the susceptibility of identifying glaucoma progression. We extend these concepts to our dataset consisting of longitudinal SD-OCT image sequences and predict glaucomatous progression by analyzing the structural change characteristics observed in the data. This dataset contains longitudinal data with a diagnosis of open-angle glaucoma but is unlabeled for glaucoma progression. To reinforce the notion of progressing eyes from non-progressing eyes, we introduce a new subset of normal eyes from a small cohort of patients identified as healthy using the International Classification of Diseases (ICD) and Current Procedural Terminology (CPT) codes from the DOR database (the exact inclusion/exclusion criteria is defined in the following chapters). The problem effectively becomes Positive Unlabeled (PU) learning, which is one of the most common yet complex learning paradigms for discriminating positive and negative classes from a partially labeled positive dataset (\cite{li2005learning, bekker2020learning}). It's important to note that in our scenario, the terminology for PU becomes counterintuitive as the "positive" set contains "healthy eyes," which are typically "negative" samples in traditional settings, but this does not change the inner working of PU learning (technically Negative-Unlabeled Learning). In this section, we formally introduce the modified PU learning and attempt to obtain error bounds for the learning task to show the feasibility of the method.

Unlike traditional PU methods, which rely on the calculation of a class prior to modeling classification systems, we let the DL model estimate class densities directly (\cite{bekker2020learning}).  This is done by a modified PU learning task, which breaks down the learning process into two tasks:

\begin{itemize}
    \item \textbf{PU Learning Phase}: A one-class classifier identifies the negative class (healthy eyes) from the unlabeled set (eyes that might or might not be progressing).
    \item  \textbf{Noise Learning Phase}: The unlabeled samples are treated in two subsets: a pseudo-progressing group based on the original sequence showing systematic time-related changes in glaucoma and a non-progressing group that is shuffled with time to remove any temporal dependencies. Noise Learning effectively becomes a binary classification to distinguish between the positive (pseudo-progressing) class and the negative (non-progressing) class.
\end{itemize}

\subsection{Hypothesis and Mathematical Formulation}

Let $\mathcal{S}_{healthy}$ be the set of healthy eyes (positive set in the PU learning context) and $\mathcal{S}_{unlabeled}$ (negative set in PU learning context), therefore PU dataset:
\begin{equation}
    \mathcal{D}_{PU} = \mathcal{S}_{healthy} \cup \mathcal{S}_{unlabeled}
\end{equation}

\subsubsection{Positive Unlabeled (PU) Learning Task:}

\textbf{Objective:} To differentiate between  $\mathcal{S}_{healthy}$ and $\mathcal{S}_{unlabeled}$ using:
\begin{equation}
    f_{PU} : \mathcal{D}_{PU} \rightarrow \{0,1\}
\end{equation}
where 
\begin{align}
f_{PU}(x) &= 1 \implies x \in \mathcal{S}_{\text{unlabeled}}, \\
f_{PU}(x) &= 0 \implies x \in \mathcal{S}_{\text{healthy}}.
\end{align}

We formulate our loss objective $L_{PU}$ for PU learning as BCE over all samples of PU dataset (equation \eqref{eq:bce})

\subsubsection{Noise Learning Task:}

We synthetically create a pseudo-progressing criteria for glaucoma progression. Assuming $\mathcal{S}_{unlabeled} = \{X_1, X_2, \dots, X_{N'}\}$ are the $N'$ unlabeled samples in the set. Shuffling with a random shuffling function $\Pi$ with $k$ permutation yields a set $\mathcal{S}_{shuffled} = \mathcal{S}_{\Pi(unlabeled)} = \{\Tilde{X}_{1}, \Tilde{X}_{2}, \dots, \Tilde{X}_{N'}\}$ where each $\Tilde{X}_{i} = \{X_{\Pi(i), 1}, X_{\Pi(i), 2}, \dots, X_{\Pi(i), k}\}$ contains $k$ randomly shuffled input sequences. Here random shuffling is defined as $X_{\Pi(i)} = (x^{\pi(1)}, x^{\pi(2)}, \ldots, x^{\pi(\tau)})_i$, where $\pi(i)$ is a random number in ${1, 2, \dots, \tau}$. 

Based on the definitions discussed above:

\begin{itemize}
\item $\mathcal{S}_{\text{pseudo-progressing}} = \text{Original sequence in } \mathcal{S}_{\text{unlabeled}} \text{ (pseudo-positives)}$.
\item $\mathcal{S}_{\text{non-progressing}} = \text{Shuffled sequences from } \mathcal{S}_{\text{shuffled}} \text{ (pseudo-negatives)}$.
\end{itemize}

Therefore, the shuffled dataset becomes:

\begin{equation}
    \mathcal{D}_{shuffled} = \mathcal{S}_{pseudo-progressing} \cup \mathcal{S}_{non-progressing}
\end{equation}

\textbf{Objective:} To differentiate between  $\mathcal{S}_{pseudo-progressing}$ and $\mathcal{S}_{non-progressing}$ using:
\begin{equation}
    f_{noise} : \mathcal{D}_{shuffled} \rightarrow \{0,1\}
\end{equation}

where, 

\begin{align}
f_{noise}(x) &= 1 \implies x \in \mathcal{S}_{\text{pseudo-progressing}}, \\
f_{nosie}(x) &= 0 \implies x \in \mathcal{S}_{\text{non-progressing}}.
\end{align}

Similar to PU learning, we model the loss objective $L_{noise-model}$ for the noise learning as BCE (equation \ref{eq:bce}) over the shuffled dataset. 

\subsubsection{Joint Objective Function for Noise-PU Model}

Combining both the losses, we model the objective for the joint learning task as the weighted sum of the two losses:

\begin{equation}
\label{eq: jointnpu}
    J (\theta, \mathcal{D}_{PU}, \mathcal{D}_{shuffled}) = L_{PU} (\theta, \mathcal{D}_{PU}) + \alpha \cdot L_{noise-model} (\theta, \mathcal{D}_{shuffled})
\end{equation}



Where $\theta$ represents the CNN-LSTM model paramenters and $\alpha$ is the weighting factor for our learning scheme, which instructs the algorithm to focus on a particular learning task.

\subsection{Objective Rationale}

To show that joint training offers an advantage over standard Positive-Negative learning formulation, we need to identify the tradeoff between the performance of deep learning model on the unlabeled samples in the PU model and the negative samples in the noise model. In general, assuming that the shuffled sequences provide non-progressing information, it is expected that optimising for combined loss for the PU and Noise Model would be lower than the loss for PU model alone, especially if the pseudo-negative sequences can capture non-progressing conditions effectively. Since the $i \cdot i \cdot d$ conditions are lost when shuffling the time-series sequence in the Noise Model, it is challenging to quantify the improvements observed in joint training in terms of error bounds. Modeling error bounds requires information about the data characteristics and assumptions, which is out of scope. In the following chapters, we provide empirical evidence to show that the joint training model with PU and Noise learning offers a competitive advantage in learning inherent progressive characteristics in medical times-series data.

\section{Modelling DL with External Labels: A Contrastive Learning Approach}
\label{sec: conmodel}

U.S. Food and Drug Administration (FDA) recommends event-based analysis of SAP deterioration as endpoints for progression (\cite{weinreb2011glaucoma}). Inspired by the EMGT study, GPA has been gaining traction as one of the reliable reference standards for event-based progression due to its clinical acceptance, objectivity, high sensitivity, and supportive research base (\cite{arnalich2009performance}). Many pieces of research have shown GPA progression has moderate to good agreement with glaucoma experts with potentially detecting early progression (\cite{aref2017detecting, TANNA2012468}). We have already shown the relevance of GPA as an endpoint for reference standards in Sections \ref{ss:clinicgpa} and \ref{ss:refst}. Using the modeling procedures described in Section \ref{sec: npu}, we extend the definitions of the DL process to a labeled case where the labels are obtained from an external source. Specifically, we propose a DL approach that uses longitudinal sequences of SDOCT B-scan images from the DOR database to predict VF GPA indicative of glaucoma progression. The SDOCT images are derived from the detailed longitudinal structural assessments in the DOR, whereas VF GPA labels, which are binary (originally ternary - non-progressing, possible progressing, and likely progressing), come from VF SAP tests belonging to the same patients. Potential mismatches in the visit dates for SDOCT scans and SAP tests and the inherent discrepancy between early and late-stage progression detection between structure and function might introduce noise or other inconsistencies in labels. Furthermore, challenges in obtaining enough data points during VF SAP follow-ups lead to imbalances, making the modeling process complex. To overcome these difficulties, we use a three-step training method with a DL model that's built on the SimCLR-based contrastive learning framework.

 In the first step, the base model acts as a binary classifier, differentiating between progression and non-progression using the original SD-OCT image sequences and their corresponding VF GPA labels. Parallelly, a second step focuses on discerning VF-derived glaucoma progression (GPA) with non-progressing by introducing controlled randomness (shuffling) in the sequence of images, aiming to mimic non-progressing eyes. Images are adversarially augmented to regularize the model for structural invariance. In addition, label-smoothing binary cross-entropy is used to address imbalances that arise from the shuffling criteria. In the last phase, the latent representations from both training stages are mapped into two distinct spaces using two projection heads. We then use the SimCLR framework to extract contrastive features from these spaces, enhancing the model's precision in differentiating true VF-derived glaucoma progression from test variability due to non-progressing eyes. This tripartite training process is designed to improve the model's performance in detecting progression while solving important data ambiguities observed in medical datasets. The processes are elaborated in the following subsections.

 \subsection{Hypothesis and Mathematical Formulation}

 Extending the definitions in Section \ref{sec: npu}, we describe SDOCT-GPA dataset $\{X_i,y_i\} \sim \mathcal{D}$, where $X_i = (x^{(1)}, x^{(2)},\dots x^{(\tau)})_i$ is the $i^{th}$ longitudinal SDOCT image sequence with followup $T$, and $y_i \in \{0,1\}$ is the VF GPA label representing glaucoma progression at an endpoint $\mathcal{T}$ where $\mathcal{T} \ge T$. Here $0$ represents non-progressing while $1$ represents progressing. 
 
 Using the above data as input, we want to develop a time-series DL model $h$ with parameters $\theta$ that can predict glaucoma progression using $\mathcal{D}$. The $h$ model is a CNN-LSTM network discussed earlier, which consists of three parts:
 \begin{itemize}
     \item CNN: $\mathbb{R}^{T \times H \times W} \rightarrow \mathbb{R}^{T \times F}$ Is a 3DCNN + pretrained ResNet50 deep CNN that extracts feature vector of dimension $F$ from $T$ images of size $H \times W$.
     \item LSTM: $\mathbb{R}^{T \times F} \rightarrow \mathbb{R}^{Z}$ is the LSTM network that processes the longitudinal sequence of feature vectors and outputs a fixed-size feature vector of dimension $Z$.
     \item Classification Head: $\mathbb{R}^{Z} \rightarrow [0,1]$, is the classifier made of fully connected layers and the sigmoid activation. It returns the probability of progression based on the CNN-LSTM output.
 \end{itemize}

 Typically, unless otherwise stated, the DL output $\hat{y}_i$, 

\begin{equation}
\hat{y}_i = 
\begin{cases}
1 & \text{if } h(X_i) \geq 0.5 \\
0 & \text{otherwise}
\end{cases}
\end{equation}

Since $\mathcal{D}$ is imbalanced and noisy, the learning algorithm is derived by a Regularized Contrastive Learning Model which consists of three learning steps:

 \subsubsection{Binary Classification with Original Data}

 \textbf{Objective:} To differentiate between progression and non-progression using original labels. To do this, we train the model $h$ on $\mathcal{D}$ using the original labels, refered as $\mathcal{D}_{orig}$. We use $L_{BCE}$ as the loss objective which is the standard BCE loss (equation \ref{eq:bce}) defined earlier.

\subsubsection{Augmented Learning with Pseudo-Progression}

This step is inspired by the modified PU learning model described in Section \ref{sec: npu}. In this step, we introduce strong adversarial augmentations to the input features $X_i$ to ensure that our model captures the most discriminative and robust representations of $X_i$. This makes the DL model resilient to adversarial perturbations and label noise, which enhances the generalizability by focusing on consistent and invariant features of $X_i$.

We generate a new shuffled dataset from $\mathcal{D}$ where $X_{shuffle;i}$ represents shuffling of $X_i$ with permutation 1. We define probability $p$ as a parameter, where a sample from the original dataset is shuffled with $p$ if $y_i = 0$ (non-progressing), and with $(1-p)$ if $y_i = 1$ (progressing). The labels for these samples are set to $Y_{shuffle;i} = 0$ to represent characteristics of "true" non-progression. If $\mathcal{A}$ represent strong adversarial augmentations applied to image sequences, the new data pairs $(X_{shuffle}^{*}, y_{shuffle}) \sim \mathcal{D}_{augmented,shuffle}$ becomes:

\begin{equation}
\{X_{shuffle;i}^{*}, y_{shuffle;i}\} = 
\begin{cases} 
\{\text{shuffle}(\mathcal{A}(X_i)), 0\} & \text{if } y_i = 0 \text{ with probability } p, \\
\{\text{shuffle}(\mathcal{A}(X_i)), 0\} & \text{if } y_i = 1 \text{ with probability } 1-p, \\
\{\mathcal{A}(X_i), y_i\} & \text{otherwise},
\end{cases}
\end{equation}

We see that the label imbalance increases with $p$. The loss objective for the augmented dataset, $L_{smooth-BCE}$ is formulated as label smoothed BCE loss which uses BCE loss on $y_{smooth}$ by converting labels $y$ using formulae:
\begin{equation}
y_{\text{smooth}} = (1 - \epsilon) \cdot y + \frac{\epsilon}{2}
\end{equation}
Where $\epsilon$ is a smoothing constant.

The shuffling scheme, therefore, generates a criterion for the model to distinguish between progression and non-progression samples by shuffling a subset of sequences to generate eyes with "true" non-progressing characteristics.

\subsubsection{Contrastive Learning with SimCLR} 

We use contrastive learning in this project due to its effectiveness in handling label noise and imbalanced datasets (\cite{li2022selective, xue2022investigating}). This approach is essential because of its ability to extract subtle variations in the data. In our approach, this method dissects and contrasts features from the shuffled sequences (hard negatives) and the original dataset (potential glaucoma progression) to enhance the model's ability to differentiate between them. The adversarial augmentations in the second step allow for contrastive learning not only to regularize but also to teach invariances in images that are critical for structural progression determination. To this effect, we employ contrastive learning by implementing SimCLR, an unsupervised method for learning variations (\cite{chen2020simple}) because of its ability to inherently learn representations from datasets where explicit labeling might be noisy or inconsistent, like our dataset. 

Let $\phi$ and $\psi$ be the parameters of the two projection heads in the DL model. Using the model definitions explained above, we derive the latent features of the CNN-LSTM base model with parameters $\theta$ as:

\begin{align}
    Z_{orig} = h(X_{orig}; \theta) \\
    Z_{aug} = h(X_{augmented}; \theta)
\end{align}

Where $X_{augmented} = X_{shuffle}^{*}$. We obtain the projections of the original and augmented (with shuffling) images using the projection heads as:

\begin{align}
    Z_{orig\_proj} = h_{\phi}(Z_{orig}; \phi) \\
    Z_{aug\_proj} = h_{\psi}(Z_{aug}; \psi)
\end{align}

SimCLR is used to maximize agreement between positive pairs (similar sequences) while pushing negative pairs (dissimilar sequences) apart in both the projected spaces. The objective function for SimCLR is:

\begin{equation}
L_{SimCLR} = \frac{1}{2N} \sum\limits_{i=1}^{N} \left[ L_{con}(Z_{orig\_proj}, Z_{aug\_proj}) + L_{con}(Z_{aug\_proj}, Z_{orig\_proj}) \right]
\end{equation}

Where $L_{con}$ is the contrastive loss between representation pairs from the respective projection heads. The formula for $L_{con}$ with projections $z$ is:

\begin{equation}
L_{con}(i,j) = -\log \left( \frac{ \exp(sim(z_i, z_j) / \uptau)}{ \sum_{k=1}^{2N} \mathbbm{1}_{[k \neq i]} \exp(sim(z_i, z_k) / \uptau)} \right)
\end{equation}

where $sim(\cdot)$ is the cosine similarity function defined as $sim(u, v) = \frac{u \cdot v}{\left\Vert u \right\Vert_2 \cdot \left\Vert v \right\Vert_2}$, $\uptau$ is the temperature parameter which scales the similarity and $\mathbbm{1}_{[\cdot]}$ is an indicator function which is $1$ when the argument in $[\cdot]$ is true, $0$ otherwise.




\subsubsection{Joint Objective Function}

The final objective function, considering all three training steps, becomes:

\begin{equation}
\begin{split}
J_{joint} (\theta, \phi, \psi, \mathcal{D}_{orig}, \mathcal{D}_{augmented,shuffle}) = L_{BCE} (\theta, \mathcal{D}_{orig}) \\ + \alpha \cdot L_{smooth-BCE} (\theta, \mathcal{D}_{augmented,shuffle}) \\ + \beta \cdot L_{SimCLR} (\theta, \phi, \psi, \mathcal{D}_{orig}, \mathcal{D}_{augmented,shuffle})
\end{split}
\end{equation}

where $\alpha$ and $\beta$ are weighting factors for smooth-BCE and SimCLR loss objectives respectively



\subsection{Objective Rationale}

To show that the joint training offers an advantage over standard training on GPA labels, we need assumptions about the distributions of data and noise, the complexity of the DL models, and more. In general, adding adversarial augmentations with shuffling (emulating "true" non-progression) and leveraging contrastive learning aims to make the model more robust and generalizable, thus reducing generalization error. However, quantifying these improvements in terms of error bounds is complicated and depends on the complex characteristics of the data, model, and augmentations, which is out of scope. We provide empirical evidence to show that the joint model is better than the base model trained on original data.

\chapter{Weakly Supervised Time Series Learning to Detect Glaucoma Progression from Optical Coherence Tomography B-scans}
\label{hello}

The research discussed in this chapter was collaboratively carried out with Alessandro A. Jammal, MD, PhD and Felipe A. Medeiros, MD, PhD.

An abstract was presented in Investigative Ophthalmology \& Visual Science, ARVO, 2023. Refer to \cite{mandal2023noise} for details.

A preliminary version of this work is in review at the American Journal of Ophthalmology, AJO, 2024 as a Full Length Article.

\section{Introduction}

This chapter proposes a novel DL algorithm to to detect glaucoma progression using OCT images, in the absence of a reference standard. Recent years have witnessed a surge in research centered around the development of DL and AI algorithms aimed at improving glaucoma assessment (\cite{thompson2020review}). While most of these algorithms have been developed for cross-sectional assessment, only some have addressed the critical need for tracking longitudinal change - a fundamental element in monitoring the progression of glaucoma. 

A common thread among traditional DL methods that utilize supervised learning is the dependence on accurate and precisely labeled datasets. These are crucial for training the models based on universally accepted reference standards, ensuring trustworthy classifications (\cite{thompson2020review}). In the context of glaucoma progression, however, no such reference standard exists. While the detection of disease progression can be aided by clinical software based on parameters from OCT and SAP over time (\cite{heijl2003measuring}), the determination of progression is ultimately dependent on the clinician's subjective evaluation. Such evaluations include the complex task of discerning true glaucomatous changes from normal aging effects (\cite{bussel2014oct,vianna2015importance}). Even when performed by expert graders, this assessment suffers from the low agreement and reproducibility (\cite{ohnell2016structural}). For training AI algorithms, this reliance on human evaluation as the reference standard can be problematic, reducing the accuracy of the algorithm if an imperfect classification is used as the "gold standard." 

Alternative approaches have been suggested for dealing with the need for a perfect reference standard when assessing changes over time. For instance, one method used to assess progression on OCT randomly rearranges sequences of images from glaucomatous eyes, effectively eliminating any systematic changes over time (\cite{belghith2015learning}). The rearranged sequences are then classified as stable or "non-progressing" cases. The algorithm is trained to recognize these stable sequences, and any sequence not classified as stable is presumed to show progression. The "hit ratio", or the percentage of images not identified as stable, serves as an indirect measure of the algorithm's sensitivity in detecting progression.

However, this approach overlooks a key factor: the presence of age-related changes in OCT images (\cite{sung2009effects}). An algorithm trained solely on scrambled images could be confounded by these changes, which it might mistakenly identify as "progression." Prior research indicates that age-related changes in OCT B-scans are pretty common and can impact multiple layers (\cite{margolis2009pilot, ramrattan1994morphometric, shigueoka2021predicting}), which may lead to a high false-positive ratio for an algorithm trained exclusively on scrambled images. 

In the current study, we propose an innovative approach for training a DL algorithm for detecting glaucoma progression on OCT scans by combining training on scrambled images with a parallel training process for recognizing age-related changes. We demonstrate that such an approach performs superiorly to standard approaches based on summary parameters for detecting changes over time while maintaining high specificity. 

\section{Methods}

Data for this study was obtained from a database registry designed to investigate longitudinal structural and functional changes in glaucoma. The database involves retrospective data retrieved from Electronic Health Records of subjects seen at the Bascom Palmer Eye Institute, University of Miami, Florida, and Duke University, Durham, North Carolina. The Institutional Review Board from both institutions approved the study, and the methods conformed with the tenets of the Declaration of Helsinki and the regulations of the Health Insurance Portability and Accountability Act for research involving human subjects. 

The study included subjects with a diagnosis of open-angle glaucoma as well as healthy individuals followed over time. Glaucoma subjects had evidence of glaucomatous optic neuropathy and reproducible visual field defects on SAP, defined as Glaucoma Hemifield Test outside normal limits or pattern standard deviation with $P<5\%$ (\cite{keltner2005normal}). Normal controls were obtained from a subset of eyes with IOP below 22 mmHg and no history of elevated IOP, normal ophthalmologic examination, normal appearance of the optic disc on stereo photographs, and at least two reliable normal VFs in both eyes. Subjects with a history of other ocular or systemic diseases that could affect the optic nerve were excluded. 

All eyes were required to have at least 5 Spectralis SDOCT (Heidelberg Engineering GmbH, Dossenheim, Germany) images over time. Scans were acquired using a circular scanning pattern with a 3.5mm diameter around the ONH. The sequences of B-scan images for each eye were used as input features to the deep learning model, with each SD-OCT B-scan being a grayscale image of $768 \times 496$ points resized to $224 \times 224$ pixels. Each series of five consecutive SDOCT B-scans from each eye was treated as a separate observation in the model. For eyes with more than five reliable SDOCT tests, all possible sequences of five consecutive tests from each eye were included in the dataset (e.g., $n_1$ to $n_5$; $n_2$ to $n_6$, etc.). 

\subsection{Weakly Supervised Time Series Learning}

We developed a weakly supervised time-series learning model, called Noise Positive-Unlabeled (Noise-PU) deep learning, to classify whether sequences of OCT B-scans showed progression. Positive-Unlabeled (PU) learning is a machine learning scenario where the training data consists of a set of labeled instances (positive) and a set of unlabeled instances (\cite{bekker2020learning}). 'Noise' refers to irrelevant or meaningless data in the dataset that can negatively impact the performance of a model. Therefore, a "Noise-PU" model refers to a deep learning model that is specifically designed to handle datasets with high levels of noise or mislabeled instances. The Noise-PU model was built in two steps, which used a parallel learning scheme: (a) PU Learning and (b) Noise Learning models. Both models' bases were CNN and LSTM networks, which were combined to form a CNN-LSTM model, which was then used as a spatiotemporal encoder for time series learning. 

The first learning scheme (PU-learning) was used to discriminate healthy eyes from glaucoma eyes using a sequence of OCT B-scans from a highly imbalanced class dataset (i.e., many more glaucoma than healthy eyes). To do this, all healthy eyes from the dataset were identified and labeled as "normal," as any change over time would be considered related to normal aging only and not from glaucoma. All other observations obtained from glaucoma eyes were kept unlabeled for progression as they might or might not be actually progressing. A one-class classifier was used to discriminate healthy eyes from unlabeled eyes (\cite{bekker2020learning, wolf2022weakly, yang2012positive}).

A second learning scheme was used to distinguish between possible systematic time-related changes in glaucoma from test-retest variability (i.e., noise). For that, noise was defined as a sequence of images where any possible temporal changes were removed by randomly scrambling the order of the images. These sequences were treated as negative labels (i.e., pseudo-labeled for non-progressing), while the original sequence of SD-OCT B-scans for the same eye was marked as a positive label (pseudo-labeled for progression). Of course, not all original sequences of images from glaucoma eyes would, in fact, be truly progressing, as some might be stable or exhibit age-related changes. However, this training step ensures that the model learns to identify test-retest variability in the process of building the final model. The learned parameters from this model were used to tune the PU classifier directly without the need for estimating class priors (\cite{wang2021asymmetric}).

Finally, features from the PU and Noise learning models that were learned simultaneously were combined at the classification stage and jointly trained using two classification heads (one for PU learning and the other for noise learning) against the original PU labels to determine eyes with progression while accounting for normal age-related loss. The architecture for the weakly supervised time-series model is shown in Figure \ref{fig: cnnlstm}.

\begin{figure}[!ht]
\centering
	\includegraphics[width=0.7\textwidth]{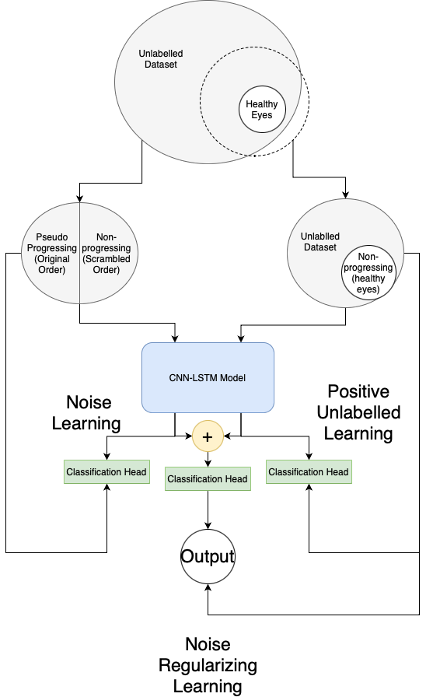}
\caption[]{An overview of the Noise-PU learning scheme, which simultaneously incorporates both positive unlabeled learning and noise learning.}
\label{fig: noisepu}
\end{figure}

The model used a Residual Deep Neural Network, ResNet50, pre-trained on the ImageNet1K dataset as the CNN encoder to learn spatial features from SD-OCT b-scans (\cite{he2016deep}). The use of pre-trained ResNet50 and transfer learning has been widely used in the development of new deep learning algorithms to save computational power while still being able to produce state-of-the-art results on smaller datasets. A 3D-CNN replaced the top layer of the ResNet50 to encode the short-term temporal dependencies. 3D-CNN is a variant of CNN used for processing 3D images, usually MRI or CT scans, but recently, it has been applied to a series of 2D images to encode time dependencies while preserving spatial features (\cite{singh20203d, parmar2020spatiotemporal}). LSTM network uses the sequence of spatial encoding obtained from the 3DCNN-ResNet50 network further to encode the long-term temporal dependencies in the data. A classification head made of FC layers was used to decode the spatiotemporal encodings into logits from which a softmax function computes the probability distribution of labels. The overall CNN-LSTM architecture is shown in Figure \ref{fig: noisepu}.

\subsection{Training and Validation}

Inputs to the DL model consisted of observations simultaneously from both the PU dataset and the artificially generated Noise dataset with 2-fold scrambling (i.e., for every observation in the training set pseudo-labeled for progression, two different randomly shuffled observations were created and pseudo-labeled for non-progression). Each observation in the dataset was made of a series of 5 consecutive SD-OCT B-scan images. A batch size of 16 was used for both datasets with a data split 70\% training set, 15\% validation set, and 15\% testing set for the PU learning. Importantly, data split was performed at the subject level to avoid data leakage from having eyes of the same subject in more than one partition. 

Image augmentations were introduced during the training phase to increase the variability in the dataset. Both one-shot and binary classification from PU learning and Noise learning, along with the predictions from the combined model used cross-entropy loss\footnote{Binary Cross-Entropy Loss (Equation \ref{eq:bce})} as the objective function. All the losses obtained from the DL model were added up\footnote{Losses were equally weighted i.e $\alpha = 1$ in equation \ref{eq: jointnpu}}, and the net loss was optimized. Stochastic gradient descent (SGD) with an initial learning rate of $8.9\mathrm{e}{-4}$, momentum 0.9 was used along with a scheduler of step size five and gamma 0.5 for 30 epochs. The evaluation was done after every epoch, and the model with the lowest validation loss was saved. Optimizing for the best cumulative sum of the sensitivity and specificity on the validation set was used to obtain the optimal weakly supervised time series model and used for further analysis on the test set. All the training and testing were done in Pytorch, Python 3.8.

\subsection{Model Evaluation and Statistical Analysis}

The performance of the DL model was compared to that of OLS regression of global retinal nerve fiber layer (RNFL) thickness, one of the clinical standards currently used for glaucoma progression detection for clinical validation (\cite{abe2016relative}). A sequence was declared as progressing with OLS if it got a statistically significant negative slope $(\mu m/year)$ with $p < 0.05$. Since there was no 'true label' or 'gold standard' test for glaucoma progression to provide a ground truth for comparison of OLS vs. the DL method, relative measures were used. Therefore, instead of sensitivity, the hit ratio served as a proxy for comparisons between the DL and OLS predictions. The DL method was said to be better at detecting glaucoma progression if its hit rate was higher than that of the OLS method when the specificities were equalized in the test set. The DL method's and OLS methods' specificities were matched by adjusting the DL method's probability threshold for predictions.

To assess the technical performance, the DL models AUC and hit-ratio (sensitivity) at matched specificities were compared to various machine learning (ML) and DL techniques used for identifying glaucoma progression. These techniques included classic ML methods like multi-layer perceptron (MLP) classifiers (\cite{bizios2010machine}), feature engineering methods such as fast Fourier transform or wavelet Fourier transforms with support vector machines (FFT-SVM and WFT-SVM; \cite{kim2013novel}, and unsupervised methods like Logistic Regression with principal component analysis (PCA; \cite{christopher2018retinal}). All these methods used RNFL thickness measures from OCT scans as inputs. DL methods for detecting progression included techniques like convolutional long short-term memory (ConvLSTM) networks (\cite{dixit2021assessing}) and transformers with image stitching inputs using the SWIN based classifier (\cite{liu2021swin}). These methods used a sequence of 2D OCT B-scan images as inputs. 

Additionally, two ablation studies were also done to assess individual components of our learning approach: the PU learning method (\cite{bekker2020learning}) and the Noise learning method (\cite{belghith2015learning}). The training was done on the same dataset for fair evaluations against a common baseline. Performance evaluations were carried out on the held-out test set comprising original image sequences from eyes not included in the training or validation datasets.

McNemar's test was used to compare the hit rates of the methods. A mixed effect model nested at patient and eye level was used to check for correlation between the DL model predictions and age to ensure that the DL method predicts true glaucoma progression rather than age-related changes. Demographics and clinical characteristics of eyes determined as progressing and not progressing by the DL model were compared using linear mixed models to account for inherent correlations between eyes of the same subject and sequences of the same eye. 

\section{Results}

\begin{table}[!ht]
\small
    \sffamily
    \setlength\tabcolsep{6pt} 
\centering
\begin{threeparttable}
\begin{tabularx}{\linewidth}{@{} l *{3}{c} @{}}
    \caption{Baseline Demographics and Clinical Characteristics for glaucoma and healthy eyes for all subjects included in the study.}
    \label{tab: tab1} \\
    
    \toprule
    
    & \thead[bl]{Glaucoma} 
    & \thead[bl]{Normal} 
    & \thead[bl]{Total} \\
    \midrule
    
\thead[bl]{No. Subjects}                            & 1802     & 57     & 1859  \\
\thead[bl]{No. eyes}                                & 3142     & 111    & 3253  \\
\thead[bl]{No. sequences}                           & 8165     & 620    & 8785  \\
\thead[bl]{Female Sex $\pmb{(\%)}$\tnote{1}}                        & 54\%     & 66\%   & 55\%  \\
\thead[bl]{Race, Black or AA $\pmb{(\%)}$\tnote{1,3}}                  & 24\%     & 60\%   & 25\%  \\
\thead[bl]{Age at baseline $\pmb{(years)}$\tnote{1}}                 & $65.8 \pm 10.4$  & $59.7 \pm 11.0$   & $65.6 \pm 10.5$  \\
\thead[bl]{Baseline RNFL thickness $\pmb{(\mu m)}$\tnote{2,3}}          & $79.4 \pm 15.5$    & $96.2 \pm 10.1$   & $80.6 \pm 15.7$  \\
\thead[bl]{Mean Follow-Up time $\pmb{(years)}$\tnote{2}}           & $3.6 \pm 1.5$     & $2.2 \pm 1.0$    & $3.5 \pm 1.5$   \\
\thead[bl]{Global RNFL thickness slope $\pmb{(\mu m/year)}$\tnote{2,3}} & $-0.73 \pm 1.36$    & $-0.51 \pm 1.53$  & $-0.72 \pm 1.56$ \\
    \bottomrule

\end{tabularx}
\begin{tablenotes}
\item[1] Reported on a patient level.
\item[2] Reported on a sequence level.
\item[3] AA = African American; RNFL = retinal nerve fiber layer.
\end{tablenotes}
\end{threeparttable}
\end{table}

\begin{figure}[!ht]
\centering     
\subfigure[Loss Plot]{\label{fig:anpu}\includegraphics[width=0.47\textwidth]{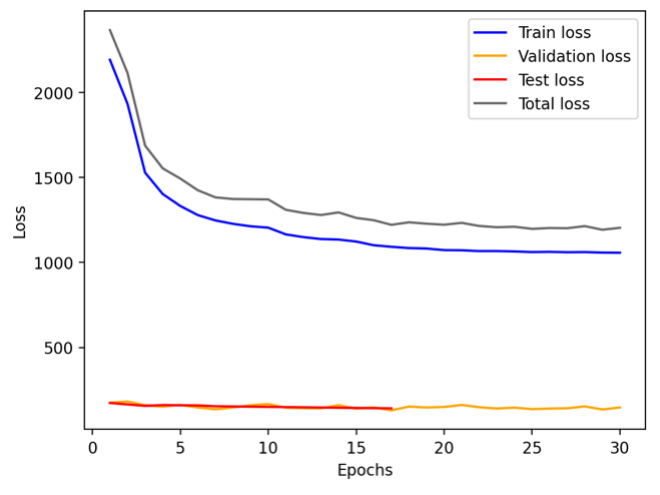}}
\subfigure[Accuracy Plot]{\label{fig:bnpu}\includegraphics[width=0.47\textwidth]{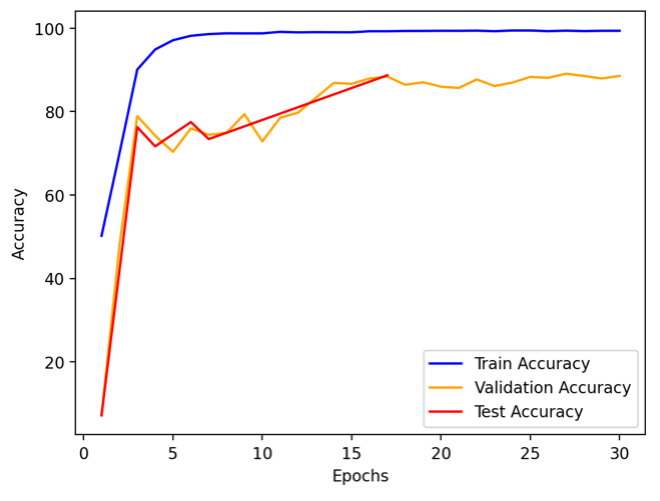}}
\caption[The Net Loss and Accuracy plots for predictions obtained from the Noise-PU Model.]{The Net Loss vs Epochs (left) and Accuracy vs Epochs (right) plots for predictions obtained from the deep learning model trained on the Noise-PU dataset.}
\label{fig: loss_acc}
\end{figure}

This study included 21,797 SDOCT B-scans from 3,253 eyes of 1,859 subjects with 8,785 sequences of 5 consecutive SD-OCT tests. Table \ref{tab: tab1} shows demographic and clinical characteristics of glaucoma and healthy eyes included in the study. The dataset was split at a patient level with a split ratio of 70:15:15 for training, validation, and testing. The model was fit to the dataset by training using the SGD algorithm. Model training was stopped when training loss and accuracy reached a stable and satisfactory state, also called convergence. The training process to convergence is given in Figure \ref{fig: loss_acc}.

From the 1225 sequences of OCT scans of the 446 glaucoma eyes in the test set, the fully trained Noise-PU DL model identified 462 sequences (38\%) as non-progressing and 763 (62\%) as progressing. Of note, since the predictions were made at the sequence level and each eye had multiple sequences, the same eye could have different sequences predicted as non-progressing and progressing. The hit-ratios of the DL and OLS methods were 0.623 (95\% CI, 0.595—0.649) and 0.069 (95\% CI, 0.056—0.084) respectively ($P<0.001$) when the specificities were equalized to 0.947 (95\% CI, 0.883—0.977). A comparison of performance metrics for the DL model against various conventional, ML and DL methods evaluated on the testing set is summarized in Table \ref{tab: tab2}.

Eyes deemed as progressing by the DL algorithm presented significantly faster rates of global RNFL loss compared with those not progressing ($-0.82 \pm 1.50 \mu m/year$ vs. $-0.63 \pm 1.54 \mu m/year$, respectively; $P=0.008$). A comparison of the demographic and baseline clinical characteristics of eyes classified as progressing versus non-progressing by the DL model is given in Table \ref{tab: tab3}. Glaucoma eyes that were classified as progressing had a significantly longer follow-up time ($P<0.001$) but presented similar disease severity as determined by the baseline age ($P=0.395$) and baseline RNFL thickness ($P=0.704$).

\begin{table}[htpb]
\small
    \sffamily
    \setlength\tabcolsep{4pt} 
\centering
\begin{threeparttable}
\begin{tabularx}{\linewidth}{@{} R *{4}{C} @{}}
    \caption[Comparison of Performance Metrics of various Machine Learning and Deep Learning methods with the Noise-PU model.]{Comparison of Performance Metrics obtained from predictions by various Machine Learning and Deep Learning methods with the Noise-PU model.}
    \label{tab: tab2} \\
    
    \toprule

    & \thead[bl]{Input}
    & \thead[bl]{AUROC\tnote{1}\\(95\% CI)} 
    & \thead[bl]{Specificity\tnote{1}\\(95\% CI)} 
    & \thead[bl]{Hit-ratio\tnote{2}\\(95\% CI)} \\
    \midrule
    
\thead[bl]{OLS\\Regression\\Method}
& Global RNFL Mean
& -     
& \makecell{0.947\\(0.883–0.977)} 
& \makecell{0.069\\(0.056–0.084)} \\

\thead[bl]{MLP\\Classifier}   
& RNFL Thickness
& \makecell{0.509\\(0.441 – 0.581)}     
& \makecell{1.000\\(1.000–1.000)} 
& \makecell{0.000\\(0.000–0.000)} \\

\thead[bl]{FFT-SVM\\Method}   
& RNFL Thickness
& \makecell{0.779\\(0.734 – 0.821)}     
& \makecell{0.947\\(0.899–0.989)} 
& \makecell{0.441\\(0.415–0.467)} \\

\thead[bl]{WFT-SVM\\Method}   
& RNFL Thickness
& \makecell{0.749\\(0.704 – 0.792)}     
& \makecell{0.947\\(0.896–0.989)} 
& \makecell{0.418\\(0.391–0.446)} \\

\thead[bl]{PCA-Logistic\\Method}   
& RNFL Thickness
& \makecell{0.821\\(0.779 – 0.859)}     
& \makecell{0.947\\(0.901–0.989)} 
& \makecell{0.439\\(0.415–0.467)} \\

\thead[bl]{ConvLSTM\\Network}
& 2D OCT B-scan
& \makecell{0.719\\(0.670 – 0.766)}  
& \makecell{0.947\\(0.899 – 0.989)} 
& \makecell{0.154\\(0.136 – 0.174)} \\

\thead[bl]{SWIN\\Base\\Transformer}
& 2D OCT B-scan
& \makecell{0.792\\(0.733 – 0.850)}  
& \makecell{0.915\\(0.859 – 0.971)} 
& \makecell{0.187\\(0.164 – 0.208)} \\

\thead[bl]{Noise\\Learning\\Model}
& 2D OCT B-scan
& \makecell{0.583\\(0.529 – 0.644)}  
& \makecell{1.000\\(1.000 – 1.000)} 
& \makecell{0.000\\(0.000 – 0.000)} \\

\thead[bl]{PU\\Learning\\Model}
& 2D OCT B-scan
& \makecell{0.711\\(0.662 – 0.760)}  
& \makecell{0.947\\(0.899 – 0.989)} 
& \makecell{0.159\\(0.139 – 0.180)} \\

\thead[bl]{Noise-PU\\Learning\\Model}
& 2D OCT B-scan
& \makecell{0.858\\(0.832 – 0.885)}  
& \makecell{0.947\\(0.883 – 0.977)} 
& \makecell{0.623\\(0.595 – 0.649)} \\

    \bottomrule
\end{tabularx}
\begin{tablenotes}
\item[1] 95\% Confidence Interval obtained from DeLong Method.
\item[2] Reported at matched Specificities at 95\%.
\end{tablenotes}
\end{threeparttable}
\end{table}

\begin{figure}[!htbp]
\centering
	\includegraphics[width=0.81\textwidth]{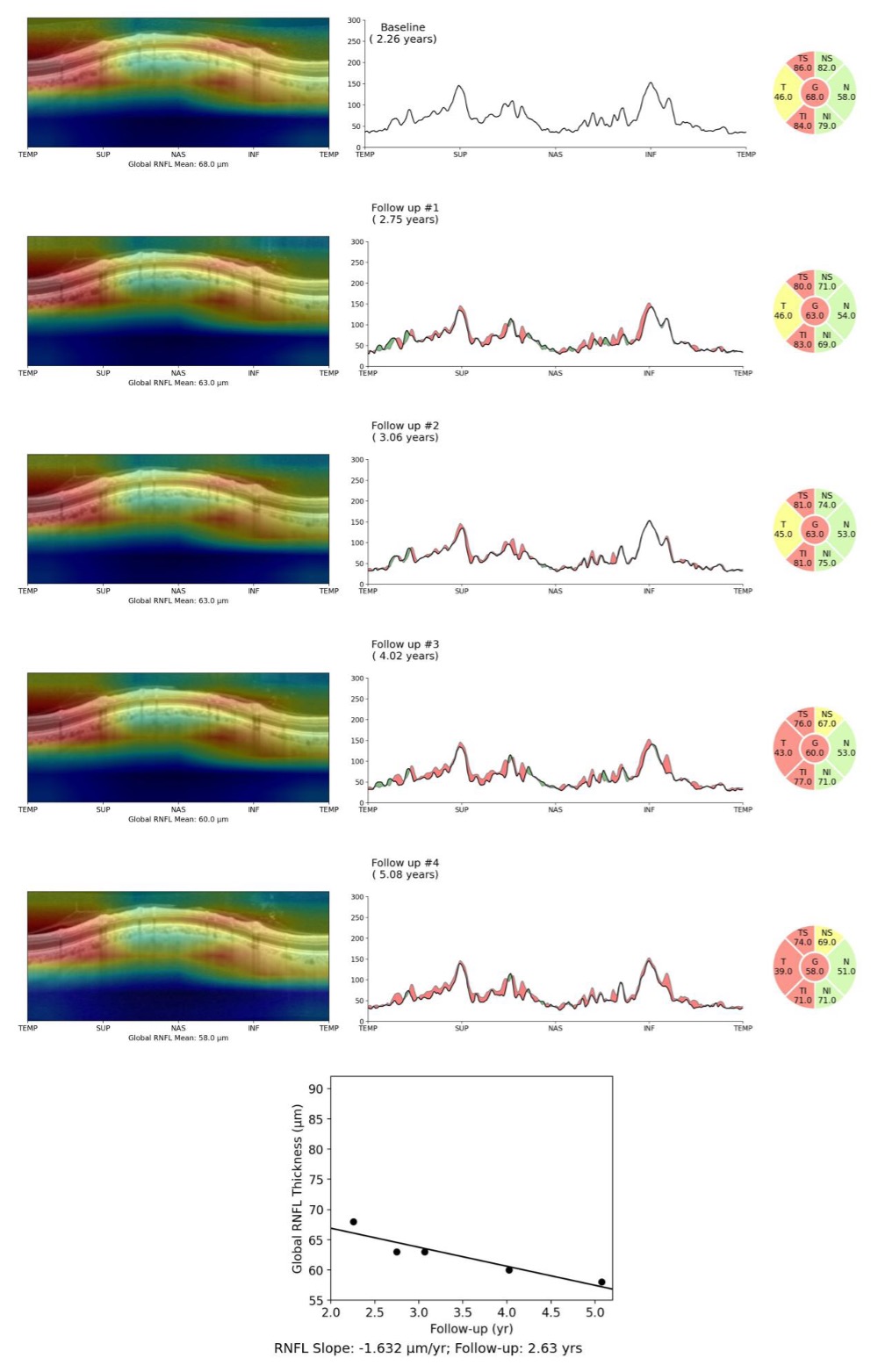}
\caption[Deep Learning Heatmaps of a glaucoma eye sequence predicted as progressing by the Noise-PU Model.]{Representative sequence of a glaucoma eye predicted as progressing by the Noise-PU Model: DL Heatmap (left), RNFL thickness profile (center), RNFL thickness sectors (right), global RNFL trend-line (bottom).}
\label{fig: prog_eg}
\end{figure}

\begin{figure}[!htbp]
\centering
	\includegraphics[width=0.81\textwidth]{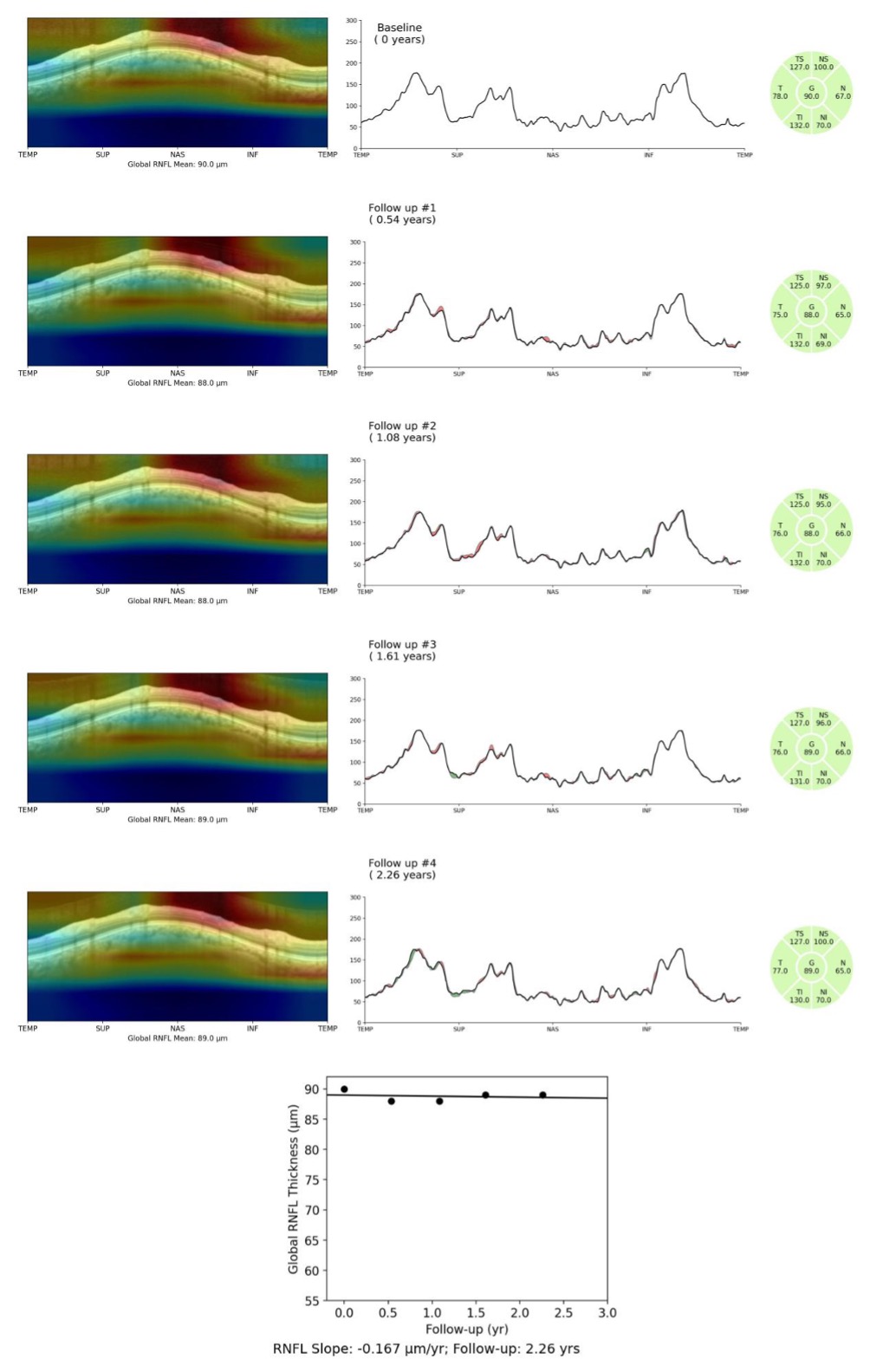}
\caption[Deep Learning Heatmaps of a glaucoma eye sequence predicted as non-progressing by the Noise-PU Model.]{Representative sequence of a glaucoma eye predicted as non-progressing by the Noise-PU Model: DL Heatmap (left), RNFL thickness profile (center), RNFL thickness sectors (right), global RNFL trend-line (bottom).}
\label{fig: noprog_eg}
\end{figure}

\pagebreak
Two representative examples of the sequence of tests predicted as progression and non-progression, along with the activation heatmaps by the DL model, the RNFL thickness profile change from baseline, sector averages, and global RNFL thickness trend-line, are given in Figures \ref{fig: prog_eg} and Figure \ref{fig: noprog_eg}. Figure \ref{fig: prog_eg}, an eye characterized as progressing by the DL model, shows a significant change in the global average RNFL thickness over time, with prominent loss in the inferior temporal and superior temporal sectors. Although little change would have been seen under manual inspection of the B-scans, the heatmaps emphasize the temporal superior, and temporal inferior regions as the most relevant areas for determining this eye as progressing by our model, in agreement with the expected pattern of glaucomatous damage. In contrast, when an eye was predicted as non-progressing, the heatmaps often highlighted non-retinal structures, like parts of the sclera and vitreous or the nasal sector (Figure \ref{fig: noprog_eg}).

\begin{table}[!htbp]
\small
    \sffamily
    \setlength\tabcolsep{4pt} 
\centering
\begin{threeparttable}
\begin{tabularx}{\linewidth}{@{} R *{3}{c} @{}}
    \caption[Baseline Demographics and Clinical Characteristics for the Deep Learning model Predictions of Glaucoma Eyes in the test set.]{Baseline Demographics and Clinical Characteristics for Glaucoma Eyes in the test set Predicted as Progressing versus Non-Progressing by the Deep Learning model.}
    \label{tab: tab3} \\
    
    \toprule
    
    & \thead[bl]{Progression} 
    & \thead[bl]{Non-progression} 
    & \thead[bl]{p-value} \\
    \midrule
    
\thead[bl]{No. Subjects}   & 210    & 145  & -   \\
\thead[bl]{No. eyes}    & 336    & 195   & -      \\
\thead[bl]{No. sequences}   & 763    & 462  & -      \\
\thead[bl]{Female Sex $\pmb{(\%)}$\tnote{1}} & 51\%  & 51\%    & 1.000\tnote{3} \\
\thead[bl]{Race, Black or AA $\pmb{(\%)}$\tnote{1}}  & 24\%  & 29\%   & 0.452\tnote{3} \\
\thead[bl]{Age at baseline $\pmb{(years)}$\tnote{2}}  & $68.2 \pm 9.9$   & $65.0 \pm 10.3$   & 0.395\tnote{4} \\
\thead[bl]{Baseline RNFL   Thickness $\pmb{(\mu m)}$\tnote{2}}   & $77.4 \pm 14.1$  & $84.3 \pm 13.9$   & 0.704\tnote{4} \\
\thead[bl]{Mean Follow-Up Time $\pmb{(years)}$\tnote{2}} & $3.8 \pm 1.6$  & $3.0 \pm 1.6$  & \textbf{0.001}\tnote{4} \\
\thead[bl]{Mean RNFL Slope $\pmb{(\mu m/year)}$\tnote{2}}   & $-0.82 \pm 1.50$  & $-0.63 \pm 1.54$    & \textbf{0.008}\tnote{4} \\
\thead[bl]{Median RNFL Slope $\pmb{(\mu m/year)}$\tnote{2}} & \makecell{-0.74\\(-1.52 – -0.098)} & \makecell{-0.57\\(-1.30 – 0.107)} & - \\

    \bottomrule
\end{tabularx}
\begin{tablenotes}
\item[1] Reported on a patient level.
\item[2] Reported on a sequence level.
\item[3] $\text{Chi}^2$ test.
\item[4] LMM nested at the patient and eye levels. 
\end{tablenotes}
\end{threeparttable}
\end{table}

\pagebreak

\pagebreak

\pagebreak

\section{Discussion}

In this study, we developed and validated a novel time-series DL algorithm to detect glaucoma progression in the absence of ground truth labels. The DL method consisted of a CNN-LSTM encoder to learn the spatiotemporal features of a series of SDOCT B-scans taken over a follow-up period. The algorithm was based on weak supervision on a severely imbalanced, partially labeled dataset (PU dataset) aimed to learn the true characteristics of structural progression for glaucoma while accounting for normal age-related loss. This was made possible by dividing the learning process into two steps: (a) PU learning and (b) Noise learning, where PU learning identifies age-related changes, and noise learning discriminates between progressing and non-progressing RNFL loss. Our methods showed statistically significant improvement over the conventional OLS linear regression for progression detection by obtaining a hit ratio of 62.3\% when compared to a hit ratio of 6.9\% by the OLS method when the specificities were equalized to 95\% ($P<0.001$). The DL method was also found to outperform other classical ML and DL methods, including advanced methods such as transformers. Ablations studies showed that the combined Noise-PU model performed better than the individual components of the model when tested on the same dataset. In contrast to other works in this area, our algorithm did not rely on any reference standard to detect glaucoma progression. 

Although SD-OCT of the retina and optic nerve head has become a widespread diagnostic tool for detecting structural damage, its application for identifying true glaucoma progression remains challenging. Inconsistent structure-function relationship, test-retest variability, age-related loss, and absence of clear reference standards are some of the reasons which have challenged clinicians and researchers in developing new approaches for glaucoma progression analysis (\cite{thompson2020review, jammal2020effect, abe2016relative, giangiacomo2006diagnosing, harwerth1999ganglion, heijl1989test, medeiros2012integrating}). So far, several traditional ML and DL methods have been proposed to diagnose glaucoma progression using human gradings or trend-based analysis as the reference standard (\cite{christopher2018retinal, yousefi2013glaucoma, murata2014new}); although they achieve high performance, these methods have considerable drawbacks. For example, human gradings are prone to subjectivity and bias, along with reproducibility issues. On the other hand, it is difficult to differentiate pathological progression from age-related losses, although some statistical approaches have been proposed to improve its specificity (\cite{leung2013impact, wu2017impact}). In the absence of a perfect reference standard, some studies have used unsupervised learning to identify patterns for glaucoma progression (\cite{christopher2018retinal,sample2005unsupervised}). Still, these techniques have produced either subpar results or present difficulties in real-world implementation (\cite{thompson2020review}), limiting their incorporation into clinical practice. Our approach overcomes these challenges by eliminating the need for a ground truth or a reference standard for progression. 


While most PU learning research has focused on utilizing simple features or simple classifiers to model noisy imbalanced data, the DL method we presented estimates the class priors internally by regularizing the PU learning with a surrogate noise learning in the classification stage. This gives the classifier sufficient information to reweight both for class imbalances and partial labeling, thereby preventing overfitting. Our model also bypasses the need for estimating the class priors and reweighting the label frequency to account for sampling bias ahead of time or through a separate density estimator (\cite{huang2006correcting, sugiyama2007direct}), which are often unavailable and inaccurate for highly imbalanced data, such as datasets in medicine (\cite{su2021positive}). Since the DL model uses state-of-the-art expressive neural networks, the algorithm can be easily replicated for complex datasets with unknown class priors while preserving model performance. Therefore, our algorithm has the potential to be translated to other disease progression tasks if similar data on a control group of stable subjects is available. In our case, our model was supplemented by learning features related to age-related change in the SDOCT B-scans from normal eyes in the PU dataset to detect glaucoma progression, but the same architecture could be used for datasets for other progressive eye diseases that affect the retina, such as age-related macular degeneration.

Studying the activation heatmaps from the DL model’s predictions can both provide insight into the most important features used by the model and may help clinicians identify crucial regions in the images that may be missed by the human eye. For instance, in Figure \ref{fig: prog_eg}, the DL model focused on the temporal, temporal superior, and temporal inferior regions of the B-scan images. This essentially means that the DL model identified that such regions behaved differently than what the model learned was expected from normal aging or noise, ultimately identifying that sequence as progressing. In sequences of images classified as non-progressing (Figure \ref{fig: noprog_eg}), the DL method frequently highlighted areas outside the retina, such as parts of the sclera or vitreous, or areas with less probability of progression, such as the nasal sector. 



The DL method in this study used information from the whole B-scan image to make determinations about glaucoma progression. While hit ratios indicated a superior ability to detect glaucomatous changes compared to simple OLS linear regression, a limitation of the DL method is that it cannot produce quantitative estimates of the rate of glaucoma progression (\cite{abe2016relative}). On the other hand, although rates can be estimated from the OLS model, it is a simplistic approach that relies on the global peripapillary RNFL thickness, making it susceptible to information loss and possibly affected by segmentation errors. Although trend analysis can also be applied to sectors, it is often unclear how to consider the many different slopes from all possible sectors for the assessment of change. It is interesting to note that the performance of OLS regression in this study was inferior to those of some previous investigations. The likely reason for that was because the evaluation was done on sequences of only 5 tests over time, which may have limited the precision of OLS estimates. However, in clinical practice clinicians are often faced with the challenge of making decisions based on a small number of tests available over time.

This study had limitations . Despite the superior hit-ratio at matched specificities of the deep learning algorithm compared to OLS, at this time, there is no other way to confirm cases of progression. Although subjective assessment could be used, this would negate the very motivation of using the proposed approach to train the models. Although the heatmaps contribute to indicate the clinical relevance of the findings,  these maps were obtained through a method called score-based class activation maps (CAM) technique. Findings from CAM methods are solely built to highlight CNN activations and caution should be exercised when extrapolating its results for clinical interpretation. Of note, we have also limited the present analysis to sequences of 5 images over time. However, the model could be expanded to consider longer sequences in the future. Finally, while our work shows promising results for the proposed approach, external validation should still be performed in datasets from different populations.

In conclusion, we demonstrated that a DL model can identify glaucomatous progression using a weakly supervised learning framework that learned features related to normal aging, and was able to differentiate change from test-retest noise. The proposed approach could potentially be expanded to other imaging modalities or diseases where a perfect reference standard for progression is lacking.

\chapter{Regularized - Contrastive Learning to Predict Functional Glaucoma Progression Using Longitudinal OCT Scans.}

The research discussed in this chapter was collaboratively carried out with Alessandro A. Jammal, MD, PhD and Felipe A. Medeiros, MD, PhD.

An abstract of the work is in review at the Investigative Ophthalmology \& Visual Science, ARVO, 2023.


\section{Introduction}

This chapter builds on the previous chapter to propose a novel DL algorithm to predict SAP progression from longitudinal OCT scans. The rate of glaucoma progression can be difficult to predict due to confounding risk factors, uncertainty in prognosis, and limitations in tests, which has allowed detecting and predicting glaucoma progression as an emerging field in glaucoma research (\cite{susanna2009unpredictability, omodaka2022clinical, termote2010challenges}). The above reasons along with the evolution and reliance on computer-aided algorithms for clinical diagnosis have motivated researchers to focus on the development of new algorithms to capture disease characteristics (\cite{giangiacomo2006diagnosing}). Recently, DL methods have been the forefront for computer-aided diagnosis and detection of glaucoma progression (\cite{thompson2020review, guergueb2023review}).

There is no gold standard test nor unified approach to evaluate glaucoma progression. Subjective clinical judgments for glaucoma progression require expertise and are prone to biases, uncertainty, and judgment errors (\cite{thompson2020review, mariottoni2023deep}). Thus, clinicians prefer assessment using objective criteria and quantitative methods to assess glaucoma progression. Amongst all forms, SAP testing and Optic Disc photography are the most common and emerging measures for diagnosis, screening, and assessment of the rate of change of glaucoma (\cite{alencar2011role, yaqoob2005spectral}). SAP test measures visual field loss and changes over time by mapping patients' responses to contrast stimuli (\cite{lucy2016structural}). Due to subjectivity and cognitive fatigue accrued during testing, SAP tests are subject to limitations such as test-retest variability and reproducibility (\cite{yohannan2017evidence}). On the other hand, SDOCT, an important test for glaucoma diagnosis, is a non-invasive test that measures structural loss by quantifying the ONH and RNFL Thickness (\cite{gracitelli2015spectral, strouthidis2010comparison}). In contrast to SAP tests, SD-OCT is objective and precise, producing high-resolution RNFL information with excellent reproducibility (\cite{abe2016relative}). However, monitoring glaucoma through SDOCT is sometimes slow, requires a high degree of expertise, and becomes unreliable in advanced stages (\cite{thenappan2021detecting}).

Previous studies have shown an association between RGC loss and visual field damage (\cite{garway2002relationship}). Research has shown that decreased RGC count usually precedes the vision loss observed through SAP tests (\cite{harwerth1999ganglion}). Clinicians and researchers have applied different statistical approaches such as joint survival, longitudinal and event-based, and mixed-effect models to predict glaucoma progression using structure and function relationship (\cite{medeiros2011combining, medeiros2009detection, nouri2021prediction}). However, no consensus has been found on the modeling criteria that can accurately capture glaucoma progression (\cite{abe2016relative}). Owing to the nuances and subtleties in glaucoma progression, the DL approach becomes a potential way to identify complex patterns in raw structural data. 

Artificial intelligence (AI) algorithms, especially DL methods, are becoming an emerging field for glaucoma progression prediction and detection. Recent studies have shown that DL methods can overcome some statistical methods' limitations in understanding the glaucoma progression criteria (\cite{mariottoni2023deep, HOU2023854, meira2013predicting}). These studies have emphasized using RNFL information from longitudinal SD-OCT scans to predict visual field worsening. But ubiquitously, these studies still need to address data imbalance, covariate shift, and noise. These also need universal gold standard criteria for evaluation. Recently, a gated transformer network (GTN) obtained state-of-the-art performance to predict visual field worsening with longitudinal OCT RNFL Thickness data. This study used an ensemble of objective and subjective criteria to obtain visual field worsening (\cite{HOU2023854}). This study showed the critical need for an accurate gold standard criterion to measure the DL method's performance. A proper gold standard represents an accurate structure-function relationship and reduces false positives in the DL model's predictions for glaucoma progression. In another study, a CNN-LSTM DL model trained on longitudinal SD-OCT images was able to distinguish between glaucoma progressing and non-progressing using only the knowledge of a healthy cohort (\cite{mandal2023noise}). This study used weak supervision to teach age-related structural deterioration and showed that the DL model generalizes well on progressing samples. 

In this study, we developed a DL method that improves the previous models to detect glaucoma progression. We use a combination of 1) standard binary classification to predict SAP progression using original OCT sequences; 2) training on a subset of adversarially augmented, selective shuffled sequences to improve model robustness; and 3) applying self-supervised contrastive learning between original and shuffled datasets to discern "true" representations of change over time. The selective shuffling process is an extension of the research in the previous chapter, which uses random shuffling to generate "hard negatives" for progression(\cite{mandal2023noise}). SimLR contrastive learning was added to learn contrastive artifacts between original and adversarially augmented images, learning underlying data distribution from potentially noisy data (\cite{chen2020simple, xue2022investigating}). Overall, the combined approach, in conjunction with the label-smoothed classifier and contrastive learning, was aimed to reduce classification noise, improve performance stability, and enhance the predictive accuracy of the base classifier. We use SAP GPA, an event-based method, to determine clinically relevant visual field loss outcomes as the reference criteria. It reports consistently high specificity amongst all other SAP progression techniques (\cite{nguyen2019detecting,rabiolo2019comparison}). We compare and evaluate the DL model's performance in identifying glaucomatous progression, specifically specificities and hit ratios, against clinically validated and recent state-of-the-art algorithms for glaucoma progression (\cite{nouri2021prediction, medeiros2009detection, HOU2023854,medeiros2023validation}).

\section{Methodology}

The data set used in this study is derived from a retrospective cohort study of patients from a database registry containing tests with longitudinal structural and functional changes in eyes. The database contains EHRs of subjects seen at the Bascom Palmer Eye Institute, University of Miami, Florida, and Duke University, Durham, North Carolina. This study was approved by the institutional review board from both institutes, along with a waiver of informed consent for being a retrospective study. Data collection methods adhered to the Declaration of Helsinki's tenets and the Health Insurance Portability and Accountability Act regulations for Human Research.

To qualify for the study, subjects needed to have a diagnosis of open-angle glaucoma based on the international classification of disease (ICD) codes, a minimum of 5 reliable VF SAP tests (Humphrey Field Analyzer II, Carl Zeiss Meditec, Inc.), and 5 reliable SDOCT (Spectralis, Heidelberg Engineering GmbH, Dossenheim, Germany) scans at an age over 18 years. Glaucoma was defined as having the Glaucoma Hemifield Test outside normal limits or Pattern Standard Deviation with $P<5\%$. Individuals who did not conform to the criteria mentioned above or had a history of other ocular or systemic diseases that could affect the optic nerve or visual field were excluded. Subjects were also excluded if the tests were performed after treatment with photocoagulation as per CPT codes.

The dataset consisted of longitudinal scans of the retina around the optic nerve head (ONH) obtained from the Spectralis SD-OCT (Heidelberg Engineering, Heidelberg, Germany) over routine clinical care. The acquisition protocol has been described in detail previously (\cite{leite2011comparison}). In summary, each SDOCT scan in the longitudinal series consisted of a cross-sectional image of the retina (B-scans; 768 x 496 points) with the peripapillary circular scanning pattern of 3.5 diameters around the ONH. This scan pattern has been established as the gold standard for the evaluation of structural glaucomatous damage by identifying the thinning of the RNFL thickness at a micrometer scale. The global average of the RNFL thickness for each scan was recorded and used as a comparison for the model (see Model Evaluation section below). Scans were excluded if they had segmentation or artifact errors or the quality score was lower than 15, according to the manufacturer's recommendation. Each observation for an eye consisted of five equally spaced sequences of successive good-quality SDOCT Bscan images.

\subsection{Definition of Glaucoma Progression}

We used the Guided progression analysis (GPA; Humphrey Field Analyzer II (HFA II), Carl Zeiss Meditec, Inc., Dublin, USA) as the definition for glaucoma progression criteria. GPA is a proprietary algorithm of HFA that uses pointwise event-based analysis on repeatable visual field loss observed in longitudinal 24-2 SAP tests (\cite{giraud2010analysis}). The algorithm flags a point on the SAP pattern deviation plot as progression if it exceeds the expected test-retest variability in the follow-up exams in comparison to two baseline exams. If three or more points over three consecutive tests present such deterioration, the algorithm flags the test as “likely progressing” (\cite{wu2017impact}). This method produces a relatively simple qualitative measure of glaucoma progression and improves on the other methods by accounting for pointwise test-retest variability, sensitivity, and age (\cite{nguyen2019detecting}). The results of the algorithm were summarized under a binary classification (progressing vs. non-progressing) and applied to each series of  SDOCT B-scan images as the label for the event of glaucoma progression. The date of the first event (out of the three consecutive events of progression) was considered as the date of progression. After a glaucoma progression event, the baseline was reset at the date of the event, and successive SAP tests were again considered to create a new event window. Such event windows were repeated until at most three glaucoma progression events were marked by GPA or no glaucoma progression was observed.

\subsection{DL Method}

A novel DL method was developed to predict whether a longitudinal sequence of SD-OCT B-scan images presented glaucoma progression or remained stable. A recent study showed that a hybrid CNN-LSTM model can learn spatiotemporal relationships exhibited by the time series image sequences (Figure \ref{fig: cnnlstm}) (\cite{mandal2023noise}). We employed a similar architecture for our research. A pretrained ResNet50 residual deep neural network was used as the CNN encoder due to its ability to produce state of the art results even on smaller datasets (\cite{he2016deep}). 3D-CNN, a 3D variant of CNN, was used as the top layer for its ability to encode short-term time dependencies and relevant spatial features (\cite{parmar2020spatiotemporal}). Due to its ability to learn sequential data, a bidirectional LSTM network was used as a time-series encoder, which encodes long-term temporal dependencies in the spatial encoding sequence obtained from the encoder. LSTM networks also have memory blocks believed to retain inter and intra-test variability in the data, thereby improving sequence predictions (\cite{mousavi2019inter}).

\subsubsection{Regularized – Contrastive (RegCon) Learning}

The DL model used a classification head of several fully connected (FC) layers to generate logits for the classification task. Softmax function was used to compute the probability distribution of labels from the logits to train with categorical cross-entropy (CCE) loss (Equation \ref{eq: cce}) \footnote{The loss objective CCE is used interchangeably with BCE from Section \ref{sec: conmodel} since the CCE is done over two categories.}. The data acquisition to generate features and outcomes for the model is expected to subsume label noise. As studies have shown that DL methods trained with CCE loss are sensitive to label noise (\cite{feng2021can}), finding a training protocol robust to noisy data is of utmost importance. Following the research outcomes of the previous chapter, which showed that identifying test variability due to normal aging in longitudinal SDOCT images could improve DL model performance (\cite{mandal2023noise}), we create a parallel learning step to  generate a subset of randomly shuffled sequence data to introduce "hard negatives." This step, called selective shuffling, generates negative (non-progressing) samples by shuffling image sequences with probability $p$ if they are progressing set and $(1-p)$ if they are non-progressing set in the training data. All the images in each image sequence are adversarially augmented (\cite{wang2022contrastive}) to introduce sufficient regularization and structural invariance. Using the novel DL algorithm discussed earlier, we create this learning step as a label-smoothed binary classifier due to increased imbalance in the modified dataset (Equation \ref{eq: smcce}).

A third learning step with self-supervised contrastive learning was introduced to mitigate the effects of label noise. Unlike other methods that improve robustness, contrastive learning produces generalizable and transferable results. Specifically, we use the SimCLR contrastive learning framework over the CNN-LSTM encoder on the original and modified dataset to identify reliable time-series representations of image sequences. Without going into details, SimCLR is an unsupervised learning method used to generate good representations of the data by comparing and contrasting with different perspectives of the original sequences through strong data augmentations (\cite{wang2022contrastive}). The loss function for SimCLR is given in Equation 3. The use of selective shuffling with adversarial (strong) data augmentation helps the model discern "true" representation of structural loss over time. 

\begin{figure}[!ht]
\centering
	\includegraphics[width=0.75\textwidth]{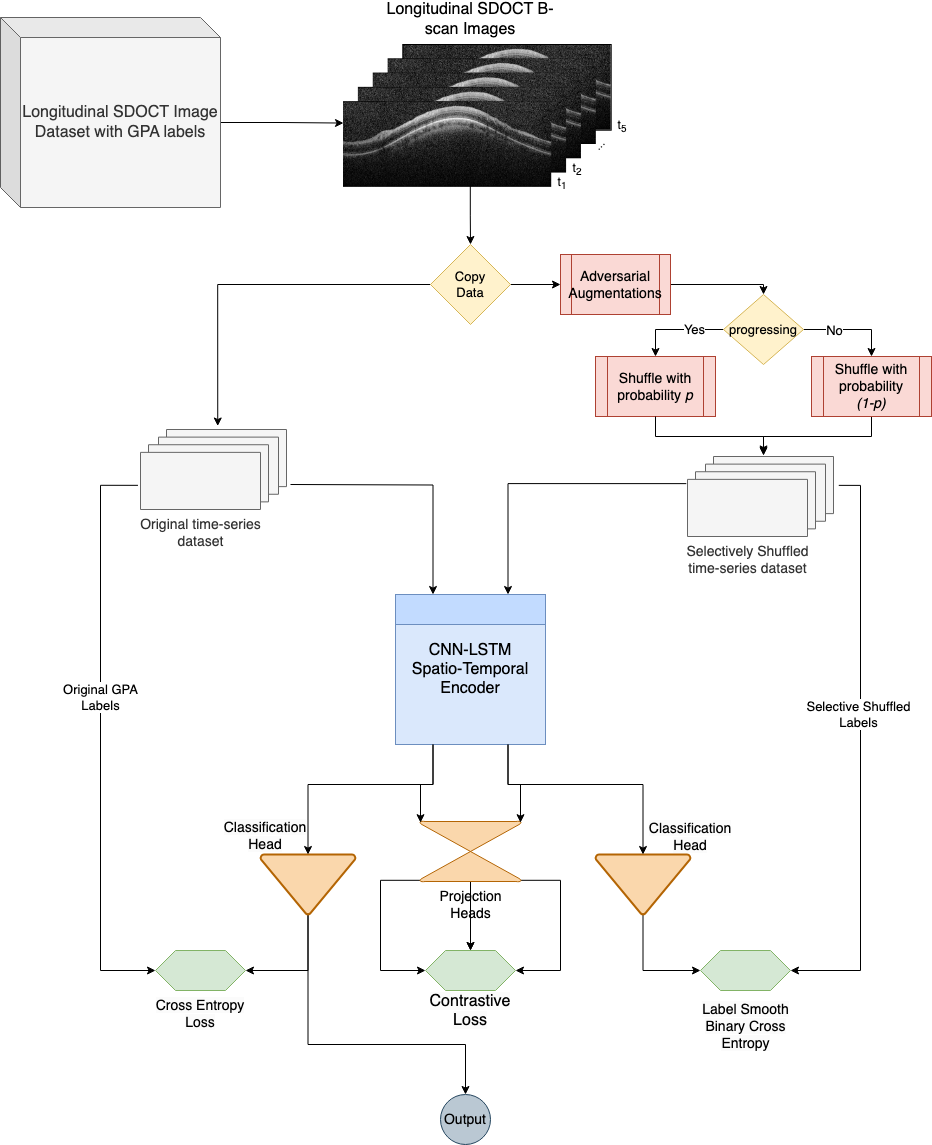}
\caption[An overview of the RegCon Semi-Supervised CNN-LSTM Network.]{An overview of the RegCon CNN-LSTM Network: made of three DL loops, main VF GPA classifier (left), selective shuffled label smoothed classifier (right) and self-supervised contrastive learning (center).}
\label{fig: regcon}
\end{figure}

Overall, the original and augmented sequences were passed through the CNN-LSTM encoder to obtain their respective time-series feature encodings. A Classification Head generated logits or probability measures for both the datasets simultaneously, which was used for Binary classification (Equation \ref{eq: cce}) and Label-Smoothed Binary Classification (Equation \ref{eq: smcce}). With the time-series encodings as input, two different projection heads made of FC layers were used to learn contrastive representations of the original and augmented features from the DL model. A contrastive loss given by Equation \ref{eq: clr} was used to maximize the agreement between representations by aligning the data distributions (negative and hard negatives) of the original and augmented sequences. This small addition improves the model's performance on real-world data exhibiting label noise and variability. Equation \ref{eq:joint} defines the overall objective function for the DL model. $\alpha$ and $\beta$ represent the contributions of the Label-Smoothed Classifier and SimCLR loss in the joint training approach. The RegCon Learning model architecture is shown in Figure \ref{fig: regcon}.

\begin{equation}
    \label{eq: cce}
    L_{CCE} = -\frac{1}{N} \sum_{i=1}^{N} Y_{t,i} \log(Y_{p,i})
\end{equation}

\begin{equation}
    \label{eq: smcce}
    L_{smooth-CCE} = -\frac{1}{N} \sum_{i=1}^{N} \bar{Y}_{t,i} \log(Y_{p,i})\text{; }\bar{Y}_{ti} = (1-\mu) Y_{ti} + \frac{\mu}{K}
\end{equation}

\begin{equation}
\label{eq: clr}
L_{simCLR} = \frac{1}{2N} \sum_{k=1}^{N} \left[ L_{\text{pos}(2k-1,2k)} + L_{\text{pos}(2k,2k-1)} \right]
\end{equation}
where, $L_{\text{pos}(i,j)} = - \log \left( \frac{\exp(\text{sim}(Z_i, Z_j) / \uptau)}{\sum_{k=1}^{2N} \mathbbm{1}_{[k \neq 1]} \exp(\text{sim}(Z_i, Z_j) / \uptau)} \right)$ and $\text{sim}(u_i, u_j) = \frac{u_i \cdot u_j}{\Vert u_i\Vert_2 \cdot \Vert u_j\Vert_2}$;

\begin{equation}
\label{eq:joint}
L_{\text{Joint}} = L_{\text{CCE}} + \alpha \cdot L_{\text{smooth-CCE}} + \beta \cdot L_{\text{simCLR}}
\end{equation}

\subsection{Training and Validation}

The DL model's inputs consist of a pair of original observation sequences for glaucoma progression prediction, a selectively shuffled, adversarially augmented view of the same observation of label-smoothed classification, and a subsequent contrastive learning loop. Each observation comprised a series of 5 successive SD-OCT B-scan images resized to $224 \times 224$ pixels. The model was trained with a data split of 70\% training set, 10\% validation set, and 20\% testing set with a batch size 48. 

The model was trained to optimize the objective function given in Equation \ref{eq:joint}. The values of $\alpha$ and $\beta$ were set to 1 each after empirical analysis. A likely reason for this hyperparameter setting might be that a lower value prioritizes CCE, which might overfit, and a higher value prioritizes label smoothing or contrastive learning, which over regularizes. The objective function was optimized with an $l_2$ regularized stochastic gradient descent (SGD) algorithm with a learning rate of 0.002, a momentum of 0.9, and a weight decay of 0.1 for 120 epochs. Cosine Annealing with Warm Restarts was used to search the learning rate space for optimal learning rate. Models at training epochs were saved if the validation loss of the model at that epoch was lower than in previous epochs. The cumulative sum of original validation specificity and sensitivity was compared for every such model to obtain the best DL model setting. All training and testing were done on the latest Pytorch snapshot in Python 3.8. Logit predictions produced by the classification head of the DL model represented the final outcome of glaucoma progression.

\subsection{Model Evaluation and Statistical Analysis}

We determined that the DL model achieved convergence when the validation loss stabilized and accuracy reached a satisfactory plateau. The receiver operator characteristic (ROC) curve, with the AUC, and the precision-recall (PR) curve, with the average precision (AP), was used to assess the discriminative power of the overall DL model. Accuracy, sensitivity (hit ratio; specificity matched at 95\%), precision, f1-score, and Matthew's correlation coefficient (MCC) were some metrics used to assess model performance. To appraise the model's performance regarding true prediction ability and relevance, these performance metrics were compared with the GTN-based DL algorithm (\cite{HOU2023854}) and a conventional non-DL approach, OLS Linear Regression. OLS is the currently widely accepted clinical standard that determines eyes as glaucoma progressing if the rates of change of the peripapillary RNFL thickness shows a significant ($P<0.05$) negative slope during analysis (\cite{wu2017impact}). 

All these models underwent training and evaluation on the same dataset to ensure fair comparison metrics. Two ablation studies, by removing (a) Contrastive Learning, and (b) both the Label-Smoothed Classifier and contrastive Learning, were performed. The ablation studies necessitated evidence that joint training was essential for the study. The DL model was considered superior if it outperformed other methods in most metrics. As a final indicator of model performance, hit ratios and specificities for the DL method were compared to others. McNemar's test between our DL model's predictions and other methods provided significance in the prediction ability. Additionally, the demographics and clinical characteristics of the eyes were compared for clinical validation. Cohen's kappa between the model's prediction and GPA progression criteria for the test set showed the level of agreement between the two. Finally, a real-life test example of the predicted progressing eye was presented to analyze the DL model's interpretation of input sequences to classify from complex data.

\section{Results}

\begin{table}[!ht]
\small
    \sffamily
    \setlength\tabcolsep{4pt} 
\centering
\begin{threeparttable}
\begin{tabularx}{\linewidth}{@{} l *{3}{c} @{}}
    \caption{Baseline Demographics and Clinical Characteristics for Progressing and Non-progressing Subjects based on the GPA criteria.}
    \label{tab: tab1_rc} \\
    
    \toprule
    
    & \thead[bl]{Non-progression\\by GPA} 
    & \thead[bl]{Progression\\by GPA} 
    & \thead[bl]{Total} \\
    \midrule
    
\thead[bl]{No. Subjects}                            & 415     & 42     & 424  \\
\thead[bl]{No. eyes}                                & 593     & 45    & 614  \\
\thead[bl]{No. sequences}                           & 593     & 56    & 649  \\
\thead[bl]{Female Sex $\pmb{(\%)}$\tnote{1}}   & 51.1\%     & 41.1\%   & 50.2\%  \\
\thead[bl]{Race, Black or AA $\pmb{(\%)}$\tnote{1,3}}  & 27.5\%     & 19.6\%   & 26.8\%  \\
\thead[bl]{Age at baseline $\pmb{(years)}$\tnote{1}}  & $65.1 \pm 10.1$  & $70.3 \pm 9.3$   & $65.5 \pm 10.2$  \\
\thead[bl]{Baseline RNFL thickness $\pmb{(\mu m)}$\tnote{2,3}}  & $77.5 \pm 16.0$    & $67.0 \pm 14.0$   & $76.6 \pm 16.1$  \\
\thead[bl]{Mean OCT Follow-Up time $\pmb{(years)}$\tnote{2}} & $5.6 \pm 1.3$     & $3.2 \pm 1.2$    & $5.4 \pm 1.5$   \\
\thead[bl]{Mean RNFL thickness slope $\pmb{(\mu m/year)}$\tnote{2,3}} & $-0.64 \pm 1.04$    & $-0.70 \pm 1.47$  & $-0.65 \pm 1.08$ \\
\thead[bl]{Mean SAP Follow-Up time $\pmb{(years)}$\tnote{2, 3}}  & $7.2 \pm 2.8$     & $4.6 \pm 3.3$    & $7.0 \pm 2.9$   \\
\thead[bl]{SAP MD at Baseline $\pmb{(dB)}$\tnote{2, 3}}  & $-3.4 \pm 4.5$     & $-10.8 \pm 7.0$    & $-4.0 \pm 5.2$   \\
\thead[bl]{Mean SAP MD slope $\pmb{(dB/year)}$\tnote{2,3}} & $-0.09 \pm 0.30$    & $-0.52 \pm 0.34$  & $-0.13 \pm 0.33$ \\
    \bottomrule

\end{tabularx}
\begin{tablenotes}
\item[1] Reported on a patient level.
\item[2] Reported on a sequence level.
\item[3] AA = African American; RNFL = retinal nerve fiber layer; SAP = standard automated perimetry.
\end{tablenotes}
\end{threeparttable}
\end{table}

The demographics and clinical characteristics of the dataset used in this study are given in Table \ref{tab: tab1_rc}. The study included 3178 SD-OCT B-scans and 4091 SAP tests from 614 eyes of 424 subjects. 9\% or 56 of the original sequences (45 eyes) were classified as progressors based on the GPA. The average age at baseline was $65.5 \pm 10.2 years$, with a mean follow-up of $5.4 \pm 1.5 years$ for the SD-OCT test and $7.0 \pm 2.9 years$ for the SAP tests. The dataset comprised 50.2\% females, and 26.8\% individuals self-identified as Black or African Americans (AA). Of the 649 SAP series, 56 (8.6\%) were identified as glaucoma progressing, and 593 (91.4\%) were non-progressing. The baseline SAP mean deviation was $-4.0 \pm 5.2 dB$. The average rate of SAP MD loss was $-0.52 \pm 0.34 dB/year$ for the glaucoma progressing group and $-0.09 \pm 0.30 dB/year$ for the stable group ($P < 0.001$).

\begin{table}[!ht]
\small
    \sffamily
    \setlength\tabcolsep{4pt} 
\centering
\begin{tabularx}{\linewidth}{@{} l *{9}{c} @{}}
    \caption[Dataset distribution used by the DL model at eyes and observation levels for training, validation, and testing set.]{Dataset distribution used by the DL model at eyes and observation levels for training, validation, and testing set (Sub = Subjects, Seq = Sequences).}
    \label{tab: tab2_rc} \\

\toprule
\multirow{2}{*}{} & \multicolumn{3}{l}{\thead[bl]{Non-progression by GPA}} & \multicolumn{3}{l}{\thead[bl]{Progression by GPA}} & \multicolumn{3}{l}{\thead[bl]{Total}}    \\
\midrule
                  & \thead[bc]{No.\\Sub}      & \thead[bc]{No.\\Eyes}      & \thead[bc]{No.\\Seq}     & \thead[bc]{No.\\Sub}     & \thead[bc]{No.\\Eyes}    & \thead[bc]{No.\\Seq}    & \thead[bc]{No.\\Sub} & \thead[bc]{No.\\Eyes} & \thead[bc]{No.\\Seq} \\
\midrule
\thead[bl]{Training}          & 252          & 355           & 355         & 25          & 27          & 34         & 258     & 367      & 389     \\
\thead[bl]{Validation}        & 65           & 89            & 89          & 7           & 7           & 8          & 66      & 93       & 97      \\
\thead[bl]{Testing}           & 98           & 149           & 149         & 10          & 11          & 14         & 100     & 154      & 163     \\
\thead[bl]{Total}             & 415          & 593           & 593         & 42          & 45          & 56         & 424     & 614      & 649  \\  
    \bottomrule

\end{tabularx}
\end{table}

\begin{figure}[!ht]
\centering     
\subfigure[Loss Plot]{\label{fig:arc}\includegraphics[width=0.47\textwidth]{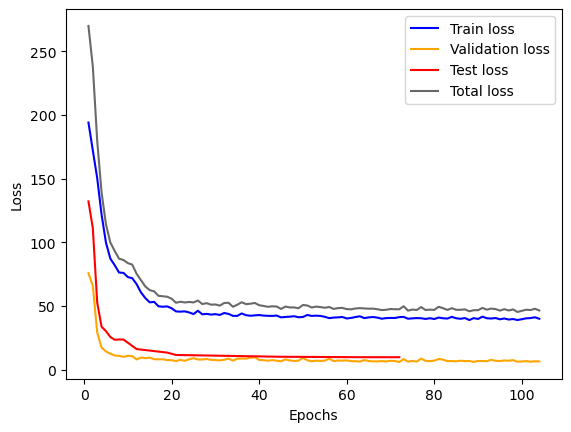}}
\subfigure[Accuracy Plot]{\label{fig:brc}\includegraphics[width=0.47\textwidth]{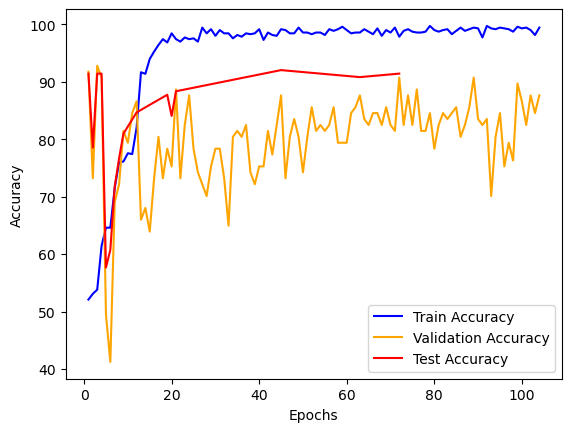}}
\caption[The Net Loss and Accuracy plots for predictions obtained from the RegCON Model training.]{The (a) Net Loss vs Epochs and (b) Accuracy vs Epochs plots during RegCON Model training for glaucoma progression detection.}
\label{fig: loss_acc_rc}
\end{figure}

\begin{figure}[!ht]
\centering     
\subfigure[ROC Curve Plot]{\label{fig:arocrc}\includegraphics[width=0.47\textwidth]{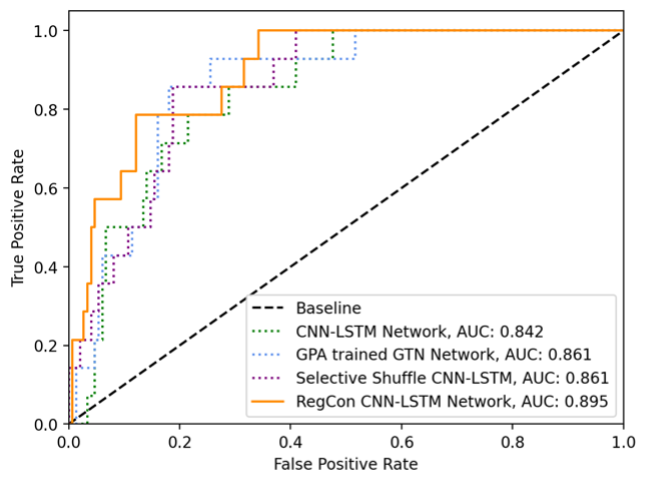}}
\subfigure[PR Curve Plot]{\label{fig:bprrc}\includegraphics[width=0.47\textwidth]{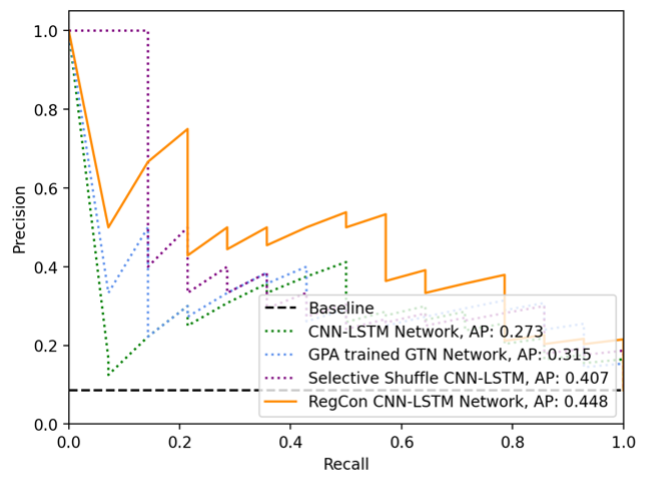}}
\caption[Receiver Operating Characteristic and Precision Recall Plots for various DL methods versus RegCon Model.]{(a) Receiver Operating Characteristic (ROC) Plot and (b) Precision Recall Plot for different DL methods with the RegCon Model evaluated on the test set for glaucoma progression.}
\label{fig: roc_pr_rc}
\end{figure}

The data was split at a patient level for training, validation, and testing, shown in Table \ref{tab: tab2_rc}. The demographics and clinical characteristics of progressing and non-progressing eyes in the test set are shown in Table \ref{tab: tab3_rc}. The training was done to fit the DL model to the dataset using the SGD algorithm till the loss and accuracy reached a stable and satisfactory state. Figure \ref{fig: loss_acc_rc} shows the loss and accuracy plots for training and testing the DL model. Evaluation of the testing set showed that the model identified 14 sequences (11 eyes; 8.6\%) as glaucoma progressing and 149 sequences (145 eyes; 91.4\%) as non-progressing. It is to be noted that each eye could have multiple GPA progressing or non-progressing events. Table \ref{tab: tab4_rc} compares performance metrics between different models used in other studies. Figure \ref{fig: roc_pr_rc} shows the predictive performance of glaucoma progression (ROC curve and PR curve plots) for all the methods. Both Table \ref{tab: tab3_rc} and Figure \ref{fig: roc_pr_rc} suggest that our DL model outperforms all other approaches. Our DL obtained an AUC score of 0.894 (95\% CI; 0.825 - 0.963). In comparison, AUC for other methods was 0.861 (95\% CI; 0.785 - 0.937) for CNN-LSTM + Selective Shuffling (ablation study 1), 0.861 (95\% CI; 0.783 – 0.940) for GPA trained GTN (\cite{HOU2023854}), 0.842 (95\% CI; 0.757 – 0.926) for CNN-LSTM model (ablation study 2). The average precision of our model (0.448) was higher than all other methods as well (0.273 – 0.407). A comparison of hit ratios at equalized specificities showed that our DL method produces better predictions when compared to other methods. Our DL Model correctly identified OCT test sequences as glaucoma progressing with a hit ratio of 0.500 (95\% CI; 0.492 – 0.508) versus 0.286 (95\% CI; 0.278 – 0.293; $P<0.001$) for CNN-LSTM + Selective Shuffling, 0.143 (95\% CI; 0.137 – 0.149; $P<0.001$) for GPA trained GTN, 0.071 (95\% CI; 0.067 – 0.076; $P<0.001$) for CNN-LSTM, 0.071 (95\% CI; 0.067 – 0.076; $P<0.001$) for OLS Regression method, current widely used to evaluate glaucomatous structural progression in routine care (all specificities matched at 95\%).

\begin{table}[!ht]
\small
    \sffamily
    \setlength\tabcolsep{4pt} 
\centering
\begin{threeparttable}
\begin{tabularx}{\linewidth}{@{} l *{3}{c} @{}}
    \caption{Baseline Demographics and Clinical Characteristics of the Test Set based on GPA reference standard.}
    \label{tab: tab3_rc} \\
    
    \toprule
    
    & \thead[bl]{Non-progression\\by GPA} 
    & \thead[bl]{Progression\\by GPA} 
    & \thead[bl]{p-value} \\
    \midrule
    
\thead[bl]{No. Subjects}                            & 98     & 10     & -  \\
\thead[bl]{No. eyes}                                & 149     & 11    & -  \\
\thead[bl]{No. sequences}                           & 149     & 14    & -  \\
\thead[bl]{Female Sex $\pmb{(\%)}$\tnote{1}}   & 42.9\%     & 30.0\%   & 0.653\tnote{3}  \\
\thead[bl]{Race, Black or AA $\pmb{(\%)}$\tnote{1}}  & 28.6\%     & 10.0\%   & 0.374\tnote{3}  \\
\thead[bl]{Age at baseline $\pmb{(years)}$\tnote{2}}  & $63.1 \pm 11.7$  & $73.4 \pm 6.2$   & 0.167\tnote{4}  \\
\thead[bl]{Baseline RNFL thickness $\pmb{(\mu m)}$\tnote{2}}  & $77.8 \pm 14.7$    & $65.7 \pm 12.5$   & 0.180\tnote{4}  \\
\thead[bl]{Mean OCT Follow-Up time $\pmb{(years)}$\tnote{2}} & $5.4 \pm 1.3$     & $3.0 \pm 1.1$    & \textbf{0.000}\tnote{4}   \\
\thead[bl]{Mean RNFL slope $\pmb{(\mu m/year)}$\tnote{2}} & $-0.35 \pm 1.40$    & $-0.58 \pm 1.28$  & 0.998\tnote{4} \\
\thead[bl]{Mean SAP Follow-Up time $\pmb{(years)}$\tnote{2}}  & $7.3 \pm 2.9$     & $4.5 \pm 3.0$    & \textbf{0.009}\tnote{4}   \\
\thead[bl]{Mean SAP MD slope $\pmb{(dB/year)}$\tnote{2}} & $-0.07 \pm 0.33$    & $-0.51 \pm 0.42$  & \textbf{0.007}\tnote{4} \\
    \bottomrule

\end{tabularx}
\begin{tablenotes}
\item[1] Reported on a patient level.
\item[2] Reported on a sequence level.
\item[3] $\text{Chi}^2$ test.
\item[4] LMM nested at the patient and eye levels.
\item[5] Boldface indicated statistical significance $(P<0.05)$.
\end{tablenotes}
\end{threeparttable}
\end{table}

\begin{figure}[!htbp]
\centering
	\includegraphics[width=0.81\textwidth]{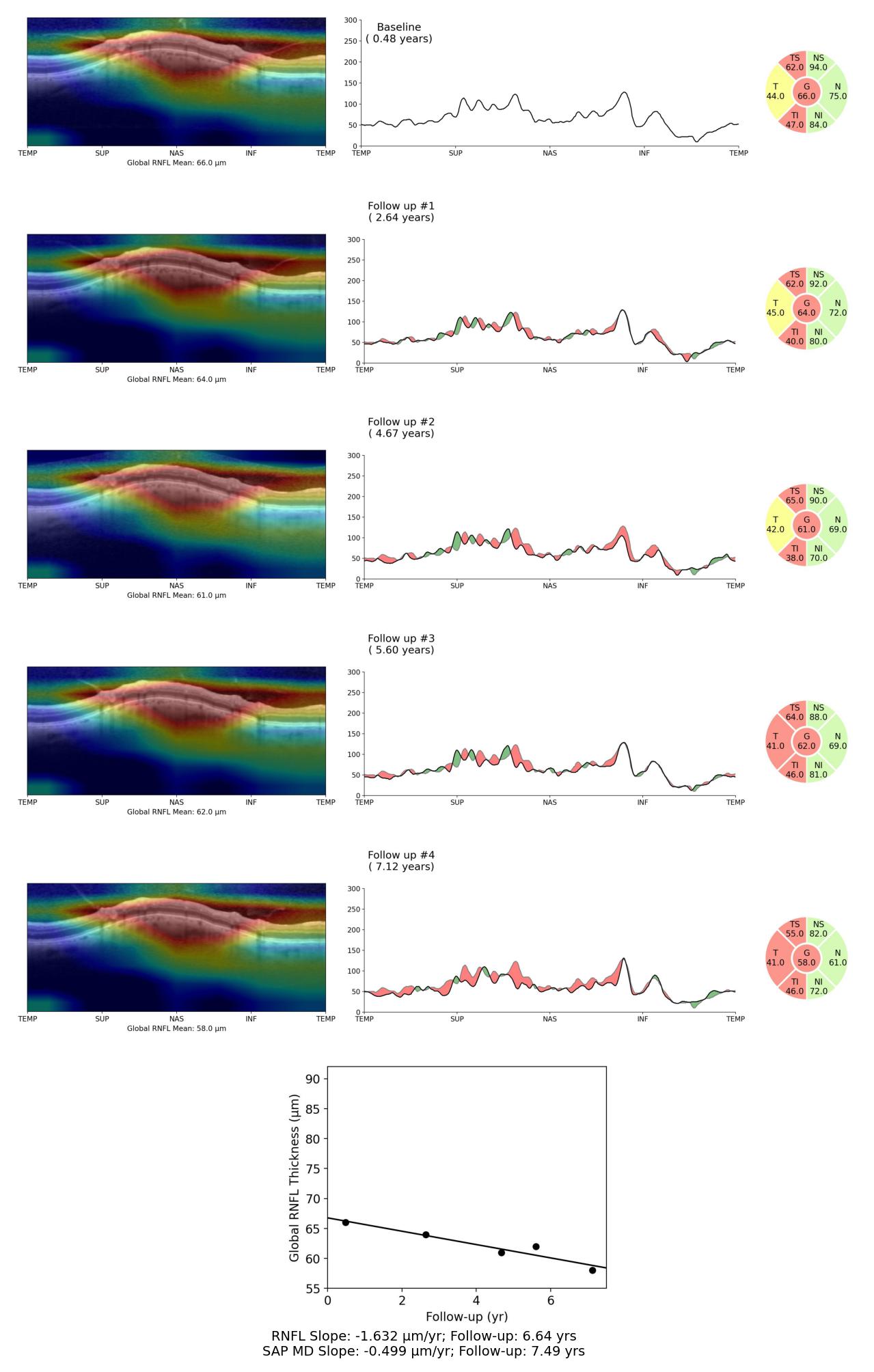}
\caption[Representative example sequence of an eye predicted as glaucoma progressing by the RegCon Model.]{Representative example sequence of an eye predicted as glaucoma progressing by the RegCon CNN-LSTM Model: DL Heatmap (left), RNFL thickness profile (center), RNFL thickness sectors (right), global RNFL trend-line (bottom).}
\label{fig: prog_eg_rc}
\end{figure}

Table \ref{tab: tab5_rc} compares the demographic and clinical characteristics of the classifications obtained from our DL model. Eyes that were predicted as glaucoma progressing had a faster rate of RNFL Thickness loss ($-0.59 \pm 1.30 \mu m/year$ vs. $-0.27 \pm 1.25 \mu m/year$; $P=0.695$) although significance was not obtained. Comparing the SAP MD slopes, the DL model obtained a significantly faster rate of SAP MD loss over time ($-0.39 \pm 0.30 dB/year$ vs $-0.09 \pm 0.35 dB/year$; $P=0.013$) when comparing progressing and non-progressing sequences. These clinical characteristics resembled the clinical and demographic characteristics obtained from the GPA progression criteria with moderate agreement (cohen’s kappa, $\kappa = 0.453$). Figure \ref{fig: prog_eg_rc} shows a representative example of the SDOCT B-scan sequence predicted as progressing by our DL model. The class activation heatmaps (CAM) from the DL method show that the model focuses on the Nasal Superior and Nasal Inferior regions as most progressing, which is reflected by the RNFL Thickness profile.

\begin{landscape}
\small
    \sffamily
    \setlength\tabcolsep{1pt} 
\centering
\begin{threeparttable}
\begin{tabularx}{\linewidth}{@{} l *{8}{c} @{}}
    \caption{Comparison of Performance Metrics across conventional and different DL model configurations trained and evaluated on our dataset.}
    \label{tab: tab4_rc} \\

\toprule

\thead[bl]{Model}  & \thead[bl]{Input} & \thead[bl]{AUROC\\(95\% CI)}  & \thead[bl]{Accuracy\\(95\% CI)}    & \thead[bl]{Sensitivity\\(95\% CI)} & \thead[bl]{Specificity\\(95\% CI)} & \thead[bl]{Precision\\(95\%   CI)} & \thead[bl]{F1 Score\\(95\% CI)}  & \thead[bl]{MCC Score\\ (95\% CI)} \\
\midrule


\thead[bl]{GPA trained\\GTN\\Classifier} & \makecell{RNFL\\Thickness\\Estimates} & \makecell{0.861\\(0.783 - 0.940)} & \makecell{0.890\\(0.888 – 0.891)} & \makecell{0.143\\(0.137 – 0.149)} & \makecell{\textbf{0.960}\\(0.959 – 0.961)} & \makecell{0.250\\(0.241 – 0.260)} & \makecell{0.182\\(0.177 – 0.187)} & \makecell{0.133\\(0.133 – 0.134)} \\

\thead[bl]{CNN-LSTM\\Classifier} & \makecell{SDOCT\\B-scans\\Images} & \makecell{0.842\\(0.757 – 0.926)}  & \makecell{0.883\\(0.882 – 0.885)} & \makecell{0.071\\(0.067 – 0.076)} & \makecell{\textbf{0.960}\\(0.959 – 0.961)} & \makecell{0.143\\(0.135 – 0.151)} & \makecell{0.095\\ (0.091 – 0.099)} & \makecell{0.043\\(0.043 – 0.043)} \\

\thead[bl]{Selective\\Shuffle\\CNN-LSTM} & \makecell{SDOCT\\B-scans\\Images}                & \makecell{0.861\\(0.785 – 0.937)}  & \makecell{0.896\\(0.894 – 0.897)} & \makecell{0.286\\(0.278 – 0.293)} & \makecell{0.953\\(0.952 – 0.954)} & \makecell{0.364\\(0.355 – 0.373)} & \makecell{0.320\\(0.314 – 0.326)}  & \makecell{0.267\\ (0.266   – 0.267)} \\

\thead[bl]{RegCon\\CNN-LSTM} & \makecell{SDOCT\\B-scans\\Images} & \makecell{\textbf{0.895}\\(0.825 – 0.963)}  & \makecell{\textbf{0.920}\\(0.919 – 0.922)} & \makecell{\textbf{0.500}\\ (0.492 – 0.508)} & \makecell{\textbf{0.960}\\(0.959 – 0.961)} & \makecell{\textbf{0.538}\\(0.530 – 0.547)} & \makecell{\textbf{0.519}\\(0.512 – 0.525)} & \makecell{0.475\\ (0.475   – 0.476)} \\

\bottomrule
\end{tabularx}
\begin{tablenotes}
\item[1] All metrics reported at a specificity equalized to 95\%.
\item[2] 95\% Confidence Interval for AUROC obtained from DeLong Method.
\end{tablenotes}
\end{threeparttable}
\end{landscape}

\begin{table}[!htbp]
\small
    \sffamily
    \setlength\tabcolsep{4pt} 
\centering
\begin{threeparttable}
\begin{tabularx}{\linewidth}{@{} l *{3}{c} @{}}
    \caption[Baseline Demographics and Clinical Characteristics for eyes predicted as Progressing versus Non-Progressing by RegCon model.]{Comparison of Baseline Demographics and Clinical Characteristics for eyes predicted as Progressing versus Non-Progressing by RegCon CNN-LSTM model.}
    \label{tab: tab5_rc} \\
    
    \toprule
    
    & \thead[bl]{Non-progression\\by DL Model} 
    & \thead[bl]{Progression\\by DL Model} 
    & \thead[bl]{p-value} \\
    \midrule
    
\thead[bl]{No. Subjects}                            & 97     & 12     & -  \\
\thead[bl]{No. eyes}                                & 146    & 12    & -  \\
\thead[bl]{No. sequences}                           & 149     & 14    & -  \\
\thead[bl]{Female Sex $\pmb{(\%)}$\tnote{1}}   & 41.2\%     & 25.0\%   & 0.440\tnote{3}  \\
\thead[bl]{Race, Black or AA $\pmb{(\%)}$\tnote{1}}  & 28.9\%     & 16.7\%   & 0.582\tnote{3}  \\
\thead[bl]{Age at baseline $\pmb{(years)}$\tnote{2}}  & $63.3 \pm 11.8$  & $71.4 \pm 7.5$   & 0.757\tnote{4}  \\
\thead[bl]{Baseline RNFL thickness $\pmb{(\mu m)}$\tnote{2}}  & $78.1 \pm 14.4$    & $62.3 \pm 11.7$   & 0.161\tnote{4}  \\
\thead[bl]{Mean OCT Follow-Up time $\pmb{(years)}$\tnote{2}} & $5.3 \pm 1.4$     & $4.1 \pm 1.8$    & 0.877\tnote{4}   \\
\thead[bl]{Mean RNFL slope $\pmb{(\mu m/year)}$\tnote{2}} & $-0.27 \pm 1.25$    & $-0.59 \pm 1.30$  & 0.695\tnote{4} \\
\thead[bl]{Mean SAP Follow-Up time $\pmb{(years)}$\tnote{2}}  & $7.2 \pm 2.9$     & $5.5 \pm 4.5$    & 0.449\tnote{4}   \\
\thead[bl]{Mean SAP MD slope $\pmb{(dB/year)}$\tnote{2}} & $-0.09 \pm 0.35$    & $-0.39 \pm 0.30$  & \textbf{0.013}\tnote{4} \\
    \bottomrule

\end{tabularx}
\begin{tablenotes}
\item[1] Reported on a patient level.
\item[2] Reported on a sequence level.
\item[3] $\text{Chi}^2$ test.
\item[4] LMM nested at the patient and eye levels.
\item[5] Boldface indicated statistical significance $(P<0.05)$.
\end{tablenotes}
\end{threeparttable}
\end{table}

\section{Discussion}

In this study, we developed and validated a novel DL algorithm that detects the presence of glaucoma progression from data and labels obtained from different sources. The DL model used longitudinal structural SD-OCT B-scan images as inputs and functional VF GPA as the reference standard for model training. The DL model used a CNN-LSTM spatiotemporal encoder to identify progression artifacts from longitudinal scans. In addition to standard classification, the algorithm consisted of joint learning with a label smoothing classification step and a self-supervised contrastive learning step to learn the underlying data distribution from noisy labeled - highly imbalanced data. The label-smoothed selective shuffling process increases the specificity by teaching the DL model to discern non-progressing samples using a subset of "hard negatives." The representations of the original and augmented data were further used to regularize the CNN-LSTM encoder with SimCLR-based contrastive learning to identify any time-series or structural image invariances and other sources of noise and discern "true" representations of change over time. The DL model we developed addressed several data challenges, such as small dataset size, label noise, and class imbalance issues observed in glaucoma progression detection. 

Our model improves on the previous methods by producing a good approximation of functional outcomes from longitudinal structural scans without reliance on clinical expertise or post-acquisition software algorithms. Unlike other methods, including the state-of-the-art GTN model, which uses clinical data preprocessing, tabular biomarkers, or clinicians' supervision while data acquisition (\cite{HOU2023854}), our approach uses a longitudinal sequence of SDOCT B-scan images, which are readily available at acquisition, without further processing or application of segmentation algorithms. This algorithm is one of the recent advances of its kind to detect functional glaucoma progression using direct longitudinal structural OCT image data requiring minimal expertise.

Since this study used events of progression as flagged by the GPA clinical algorithm as the reference standard for glaucoma progression, the longitudinal structural features learned by the DL model from the OCT images can be assumed to be a close approximation of the structure-function relationship that would lead to clinically relevant functional damage in glaucoma. The temporal features observed in longitudinal SDOCT B-scans by our model can reveal patterns of functional loss. Thus, our models overcome the difficulties observed in traditional approaches, which could not formulate true glaucoma progression criteria. Furthermore, ablation studies prove incorporating contrastive learning ensures the DL model learns functional deficits from structural changes accurately. This is particularly important because earlier methods could not identify at which stage test variability exceeds the limits for advancing glaucoma progression. Since glaucoma progression manifests as RNFL Thinning over time, our DL method can learn the variability and distinguish between "true" representations glaucomatous damage over time from other factors.

It is worth mentioning that the lower performance of the other methods on our dataset stems from the fact that other techniques are still needed to resolve data distribution and label noises. Our algorithm tries to mitigate this noise using an additional classification step with selective shuffled, adversarially augmented data. The selectively shuffled, adversarially augmented data generates a better decision boundary between progressing and non-progressing sequences by introducing hard negatives in the dataset. This classification step uses a label smoothing classifier, which teaches the model to accurately distinguish non-progressing eyes, thereby increasing specificity (\cite{mandal2023noise}). As longitudinal analysis of SD-OCT B-scans by design might be prone to test-retest variability and hinder the model's performance, a SimCLR-based contrastive learning step was added to identify and regularize time-series and structural image invariance (\cite{xue2022investigating}). It pushes the model to recognize only relevant features across different time points and reduce the influence of noise while learning artifacts in the SDOCT B-scans, improving prediction stability and adding generalization to the model. The joint learning makes our model more robust to variations during image acquisition conditions, increasing the DL model's performance, as seen in the ablation studies (Table \ref{tab: tab4_rc}).

Clinicians and researchers like to understand specific SD-OCT characteristics that contribute to glaucoma progression at various stages of the patient's life and their impact on the disease outcomes. Saliency heatmaps obtained by our DL model's activation layers using the CAM method highlight regions of interest for glaucoma (Figure \ref{fig: prog_eg_rc}). These regions of interest show evidence that the DL model looks at specific RNFL layers for discrimination. In the representative example, the DL method focuses more on Nasal Superior, Nasal Inferior, and parts of Temporal Inferior regions to determine progression. This was also reflected in the adjoining RNFL thickness profile and RNFL sector maps. We also saw that some Temporal regions showing signs of progression in the RNFL thickness profile were not highlighted by the DL model. A likely reason for this might be these regions showed uniform RNFL thickness loss, which might not have significant slopes and, therefore, not picked by our DL method. Since our study did not focus on pinpointing the exact conditions or features that influence glaucoma's progression, additional investigation of the DL model's interpretability of glaucoma features should be done with caution. However, our DL method improves over traditional interpretation techniques by adding a time-series component of glaucoma, facilitating the identification of glaucoma progression features. While it is not investigated in this study, a statistical analysis of the representation space holds significant potential to explore disease progression mechanisms, identify severity stages, and provide insights into localized deterioration for more targeted and effective treatment strategies. 

The DL model we developed achieved high performance for predicting progression (AUC 0.894 (95\% CI; 0.825 - 0.963), hit ratio 0.500 (95\% CI; 0.492 – 0.508), specificity equalized at 95\%; 0.960 (95\% CI; 0.959 – 0.961)). Compared to other methods, our model outperformed in almost all metrics with statistical significance ($P<0.001$). A likely explanation is that the GPA-trained GTN and trend-based analysis are trained with the RNFL Thickness profile, which is prone to outlier sensitivity, uncertainty, and test variability (\cite{thompson2020review, HOU2023854, hu2020functional}). GTN method might have to overfit to the high imbalance in the training data and label noise. Trend-based analysis, on the other hand, assumes changes are linear to derive glaucoma progression, which might not reflect actual progression (\cite{wu2017impact}). Since we use SD-OCT B-scan images directly, the DL model can capture complex features, interdependent RNFL profile characteristics, and progressive attributes in local receptive fields to increase the model's ability to capture high-dimensional time series image representations accurately. Another notable advantage of using our time series DL model is that it can characterize glaucoma attribute manifestations at different stages of the disease in different individuals. By leveraging the spatiotemporal and contextual information in longitudinal B-scans, our model can identify subtle changes and differential patterns across local sectors and time windows that other methods might miss.

Analysis of the clinical and demographic characteristics of the predictions from the DL method showed mixed results. Even though the rates of RNFL thickness loss in eyes predicted as progressing had faster rates of loss in average, it did not show statistical significance difference from non-progressing eyes as classified by the algorithm ($-0.59 \pm 1.30 \mu m/year$ vs. $-0.27 \pm 1.25 \mu m/year$; $P=0.695$) but showed significantly faster rates of SAP MD loss ($-0.39 \pm 0.30 dB/year$ vs $-0.09 \pm 0.35 dB/year$ respectively, $P=0.013$). A reason for this might be in the data modeling process. We used a longitudinal sequence of SDOCT B-scans from Spectralis as input features for the DL method. On the other hand, GPA labels were obtained as an end-point of event-based analysis of VF SAP data. This might cause inconsistencies in the data generation stage as visits in the OCT data might not coincide with visits in the VF SAP data.

It is important to emphasize that our model requires little clinical expertise to make predictions, as it only uses SDOCT B-scan images as inputs. This removes the dependency on manual data annotations or expertise in glaucoma diagnosis while providing inference. Relying on human gradings may make the model susceptible to biases. Using human-generated global or sectoral averages of the RNFL Thickness profile may not accurately capture pointwise variability and fail to learn the complexity of glaucoma progression. By learning glaucoma progression attributes directly from the SDOCT B-scan images and visual field SAP GPA criteria, our model learns subtle and intricate details that human graders can miss.

Despite its superior performance, this study had some limitations. One limitation of the model is its complexity, which requires significant computational resources and training times. We understand that there might be more straightforward techniques to solve label noise and data imbalance, but these methods only work for specific cases, are susceptible to data variations, and require study assumptions (\cite{natarajan2013learning}). As of now, only semi-supervised techniques provide us with the capabilities necessary for modeling such data variations. Unfortunately, in the current setting, we do not offer any analysis or study of model performance in different population settings. As with any DL model, since the performance varies across datasets, an exhaustive set of tests is required to validate for real-world use, mainly when evaluated with diverse demographic characteristics, which is affected more by the scarcity of benchmarking datasets. The comparisons provided in this study are done on our proprietary datasets constructed from different data sources and not tested for noise severity. Other models have assumed cleaner data for training and thus predictably underperformed on our dataset. Therefore, researchers should take caution when comparing statistics.

Although the regularized contrastive learning framework proposed in this study achieved comparative performance, if it not outperform other methods, it has inherent risks. The selective shuffling process might change the inherent characteristics of the data which is used to determine progression. While this process is shown to enhance specificity by introducing hard negatives, the training process still lacks information on true glaucoma-progressing samples and thus might generate a decision boundary that does not reflect the actual VF deterioration observed in glaucoma progression. Another limitation to note is using the visual field GPA as the gold standard itself. It is already understood amongst clinicians and researchers that glaucoma progression does not have a universal gold standard. However, the GPA algorithm demonstrates a clinically relevant functional progression, which has been associated with worse quality-of-life outcomes. A clinician's validation of the model performance and correlation with other clinical parameters might overcome this and strengthen the rationale behind such grading criteria. Further research on grading standards' applicability and relevance to structural progression can help optimize the model's efficacy and generalizability, addressing some of the above-mentioned limitations and ensuring its applicability in various clinical settings.

In conclusion, we developed a DL framework that combines classification with selective shuffling and contrastive learning over a CNN-LSTM network to detect glaucoma progression in noisy, imbalanced datasets. Our model demonstrates superior performance and surpasses current conventional and state of the art techniques by utilizing longitudinal SD-OCT B-scans and visual field GPA outcomes. Since the DL model is impervious to label noise, our model can be generalized and translated to other clinical practices that utilize real-world EHR data sources. By directly using SD-OCT B-scan images as inputs, our model was able to capture complex information and provide a more comprehensive evaluation of glaucoma progression. Further clinical validation for model performance and relevance with medical experts offers the potential for automated, accurate glaucoma progression detection, aiding clinicians to monitor glaucoma patients effectively.

\chapter{Conclusion and Future Work}

In this thesis, we developed various DL algorithms for detecting glaucoma progression using structural changes in the eye. Owing to the clinical need for precise glaucoma progression detection and the inherent challenges in medical image analysis, we explored the current landscape of conventional and DL methods for glaucoma progression in Chapter 2. The investigation was marked by understanding how structural, functional, or structure-function relationships improve glaucoma progression detection when integrated with complex algorithms. We observed three primary challenges in developing algorithms for progression detection: the lack of comprehensive longitudinal datasets, the absence of reliable reference standards, and complexities in structure-function relationship in glaucoma. The advent of powerful computer-aided algorithms, the accessibility to computational resources, and the availability of large datasets have helped develop innovative data-driven solutions for predicting glaucoma progression. 

In Chapter 3, we introduced the extensive Duke Ophthalmic Registry Dataset, which contains structural and functional assessments of the eye over routine clinical care of patients over two decades. Using the dataset and insights from Chapter 2, we established the input features for our DL algorithms as Longitudinal SDOCT scans. The subtleties and nuances in the structural sequences of SDOCT scans motivated us to design a time-series DL model utilizing CNN-LSTM networks to extract intricate spatiotemporal encodings. We also elaborated on the GPA reference standard and OLS linear regression-based trend analysis for glaucoma progression, which is currently widely used in clinical settings. We showed its relevance for comparison and post-hoc analysis in our studies. 

Addressing specifically the challenges of the absence of reliable reference standards and the intricacies in the structure-function relationship, in Chapter 4, we developed a novel DL algorithm. This DL model used a CNN-LSTM encoder-based approach to discern progressing eyes by training on longitudinal SDOCT scans without reliance on any reference standard. We showed that the DL model can generalize well on a time-series dataset by creating pseudo-progression criteria using age-related structural deterioration and knowledge of stable (healthy) eye characteristics. We provide a simple proof of concept for the DL algorithm in section 3.4 and show empirically with experiments in chapter 4 that the DL algorithm with modified Noise-PU learning scheme is, in fact, able to learn the intricate structural RNFL progression characteristics in glaucoma and differentiates well from natural age-related variability. 

In Chapter 5, we refined our model to a more realistic scenario to detect vision impairments observed as functional deterioration in glaucoma using structural assessments. We repurposed the DL model that uses a longitudinal sequence of SDOCT images to predict progression, as characterized by VF GPA endpoints. Building on the concepts of Chapter 4, we introduced a novel algorithm, dubbed RegCon network, that jointly addresses two critical problems in medical image analysis: data imbalance and label noise. We implemented selective shuffling to produce hard negatives. We paired these features with contrastive learning to discern patterns of age-related variability and structural invariances in longitudinal images, which indicate glaucoma progression. We demonstrated empirically that the joint learning approach outperforms other conventional methods, even exceeding the performance of the current state of the art GTN method. Our experiments showed that the DL algorithm is impervious to label noise and adeptly manages class imbalances by learning intricate details directly from the spatiotemporal representations of the DL model. Overall, the refined model showed superior performance in detecting functional deterioration in glaucoma from longitudinal structural assessments of the eye.

Despite the groundbreaking work, our research has limitations. The Noise-PU learning algorithm in Chapter 4 relied heavily on data modeling, especially the clinical characteristics of healthy subjects and the variability observed in age-related progression. The lack of positive samples (progressing eyes) possibly biased the results towards non-progressing sequences. This, coupled with the lack of benchmarking datasets, makes direct comparisons between the DL and conventional methods challenging. Our Noise-PU learning study primarily focused on detecting progression based on structural changes in the eyes. However, structural progression might not always reflect actual glaucoma progression. To ensure a comprehensive understanding, validating these findings against functional progression criteria is essential, an aspect we explored with the RegCon learning algorithm in Chapter 5. Although exhibiting competitive performance, the RegCon learning approach also had inherent risks. Specifically, the selective shuffling framework, a critical step in generating hard negatives, could change the intrinsic data distribution and manipulate the decision boundary. Even though this method showed improved sensitivity and a higher specificity, label noise (lack of knowledge of truly positive samples) might cause the DL model to generate suboptimal predictions. In both cases, validation against an external population is needed to draw accurate clinical insights into the performance of the DL models.

As the thesis concludes, several promising avenues of research emerge. Since we have shown that the DL algorithms can learn intrinsic disease characteristics from potentially noisy datasets, we can incorporate data from other sources with fewer assumptions. This allows researchers to implement a common DL algorithm across multiple datasets with minimal experitse and provide a comprehensive understanding of glaucoma progression across varied demographics, a notion wildly sought by the medical community. Integrating data from different imaging modalities, such as fundus photographs, gonioscopy, etc., can uncover deeper insights into the structure, function, and clinical relationship in glaucoma progression. Even though the performance of the CNN-LSTM network was satisfactory, the clinical need for a real-time glaucoma progression detection tool requires further optimization of our models and possibly the exploration of alternative architectures. DL networks such as the transformers have been shown to produce state of the art results for combined image and time-series analysis; incorporating them in our research might be beneficial and outperform the current setup. For clinicians and the medical community, trust and collaboration are paramount. Therefore, further research is needed to design DL methods that yield transparent and interpretable results seamlessly integrating with EHR data. This improved framework can facilitate a thorough analysis of progression for those without expert knowledge and can empower clinicians to offer individualized routine care.

In sum, our thesis has laid a solid foundation for glaucoma progression research and tackled some crucial problems observed in medical research involving real-world datasets. The developments in our study have the potential to provide purposeful and impactful solutions to various clinical problems. Validation and further research can address the limitations of our methods and refine the model's applicability across diverse clinical contexts.

\appendix
\bibliographystyle{./Bibliography/jasa} 
\cleardoublepage
\normalbaselines 
\addcontentsline{toc}{chapter}{Bibliography} 
\bibliography{./Bibliography/References} 

\biography
Sayan Mandal was born in West Bengal, India. His high school years were primarily spent in Jharkhand and Karnataka. Academically driven, Sayan secured the top position in his class at the Atomic Energy Central School in Jaduguda during the All India Senior School Certificate Examination (AISSCE). This foundation set the stage for his later accomplishments, including the achievement of the Kishore Vaigyanik Protsahan Yojana (KVPY) scholarship in 2015.

Subsequently, Sayan attended the Indian Institute of Technology (IIT) Kharagpur, completing a Bachelor in Technology (Honors) in Aerospace Engineering in April 2019. At IIT, he developed a deep interest in Control Systems and Machine Learning. This inclination led to an internship during his third year at Duke University's Humans and Autonomy Lab, resulting in published research (\cite{mandal2020acoustic,alaparthy2021comparison}).

Continuing his academic journey, Sayan enrolled in the Doctorate program at Duke University in Durham, focusing on Electrical and Computer Engineering with a specialization in Computer Engineering. His research interests revolved around the application of deep learning in medical image analysis. By December 2022, Sayan had completed his Master of Science in Electrical and Computer Engineering from Duke University. He is currently working towards his PhD degree.

\end{document}